\documentclass[12pt]{report}

\usepackage{titlesec}  
\usepackage{fancyhdr}  
\usepackage{graphicx}  
\usepackage{amsmath}   
\usepackage{amssymb}   
\usepackage{booktabs}
\usepackage{multirow}
\usepackage{amsmath}
\usepackage{bm}

\usepackage{graphicx}
\usepackage{array}
\usepackage{colortbl}
\usepackage{makecell}

\usepackage{xcolor}
\definecolor{mygray}{gray}{0.9}

\definecolor{Gray}{gray}{0.9}

\usepackage{caption}

\usepackage{hyperref}  
\hypersetup{
    colorlinks=true,    
    linkcolor=black,     
    citecolor=black,     
    filecolor=black,  
    urlcolor=black,      
    linktocpage=true    
}

\usepackage{lipsum}    

\titleformat{\chapter}[hang]{\huge\bfseries}{Chapter \thechapter:}{1em}{}
\titleformat{\section}[hang]{\Large\bfseries}{\thesection}{1em}{}
\titleformat{\subsection}[hang]{\large\bfseries}{\thesubsection}{1em}{}
\titleformat{\subsubsection}[hang]{\normalsize\bfseries}{\thesubsubsection}{1em}{}

\begin{document}

\begin{titlepage}
    \centering
    \vspace*{2cm}
    {\Huge\bfseries Personalized Video Summarization using Text-Based Queries and Conditional Modeling \\[1.5cm]}
    {\Large Ph.D. Candidate: \\ Jia-Hong Huang \\[1cm]}
    {\large Ph.D. Supervisor: \\ Prof. Dr. Marcel Worring \\[2cm]}
    {\large A thesis submitted in fulfillment of the requirements for the degree of \\[0.5cm]}
    {\large Doctor of Philosophy \\[0.5cm]}
    {\large Department of Computer Science \\[0.5cm]}
    {\large University of Amsterdam \\[2cm]}
\end{titlepage}

\pagenumbering{roman}
\tableofcontents
\newpage
\listoffigures
\newpage
\listoftables
\newpage

\pagenumbering{arabic}

\chapter{Introduction}
\begin{figure}[t!]
\begin{center}
\includegraphics[width=1.0\linewidth]{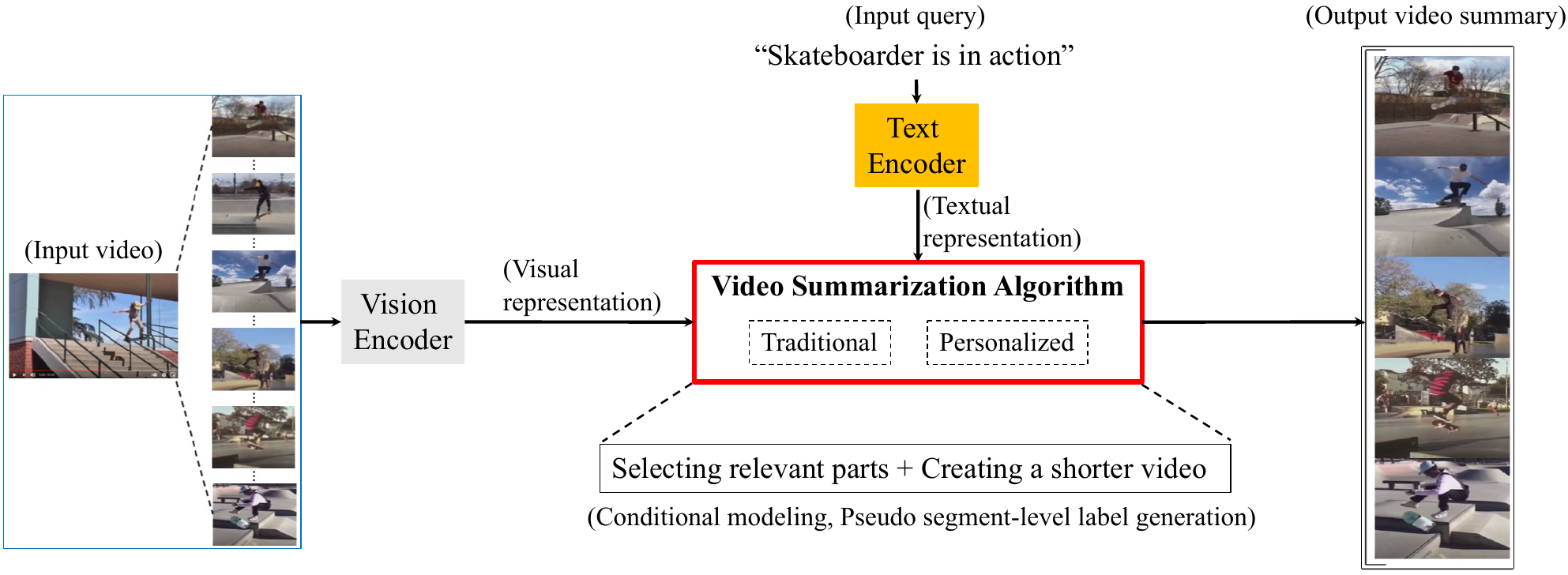}
\end{center}
\caption{\textbf{The process of video summarization.} Video summarization techniques strive to extract the most crucial and pertinent content from a video, condensing it into a concise representation. Traditional video summarization generates a summary independent of the context. To personalize the video summary, it can be created based on a given input query as context.
}
\label{fig_bias_human}
\end{figure}

Video is a crucial and natural medium for documenting real-world events due to its capacity to capture visual information, provide context, represent temporal dynamics, offer realism, facilitate analysis, and enable effective communication. Consequently, the number of users on video platforms like YouTube, Vimeo, or Brightcove is continuously increasing, leading to a massive amount of video content being generated every day \cite{yang2022science}. The large amount of video serves as a valuable source of information in many applications making it increasingly useful but also challenging for users to find and watch videos that match their preferences, interests, and investigative needs. This challenge to find the right information has spurred the development of automatic video summarization techniques which aim to extract the most significant and relevant content from a video and present it in a condensed form, as illustrated in Figure \ref{fig_bias_human}.

The quality of a summary can have a significant impact on its utility. For instance, in forensics, video evidence is often presented in court to assist in the investigation of a crime, and the quality of the generated summary can have a significant impact on an investigator's judgment. 
If the machine-generated video summary becomes unfocused, incorporating too many unrelated elements, investigators may miss crucial details essential to a case. Conversely, if the summary is rigid and lacks flexibility, it might influence an investigator's decision in a specific direction, potentially leading to inaccurate judgments.
Similar considerations apply in journalism when video summarization is used to create news reports.
If the video summary is not appropriately condensed it might not effectively encapsulate the essential aspects of an event leading to a lack of viewer interest. Additionally, it could present a distorted or incomplete perspective of the event, carrying unintended implications, especially if the video summary leans excessively toward a particular viewpoint.
Another example where the quality of the generated summary can have a significant impact on the outcome is in sports analysis. Sports videos, such as football matches, can be several hours long and analyzing them in their entirety is time-consuming and often impractical. In this context, video summarization can help analysts and coaches to quickly identify the key moments, such as goals, fouls, and other significant events, and extract insights that are relevant to their objectives. However, if the video summarization method fails to accurately identify and extract these key moments or presents an incomplete summary, the analysis and insights drawn from the video may be flawed or even misleading. This can lead to poor decision-making, ineffective strategies, and potential losses for the team. 
All these examples show that it is crucial to develop video summarization techniques that can generate effective video summaries while reducing unrelated elements, ensuring that the output is reliable and useful in various scenarios. 

\begin{figure}[t!]
\begin{center}
\includegraphics[width=1.0\linewidth]{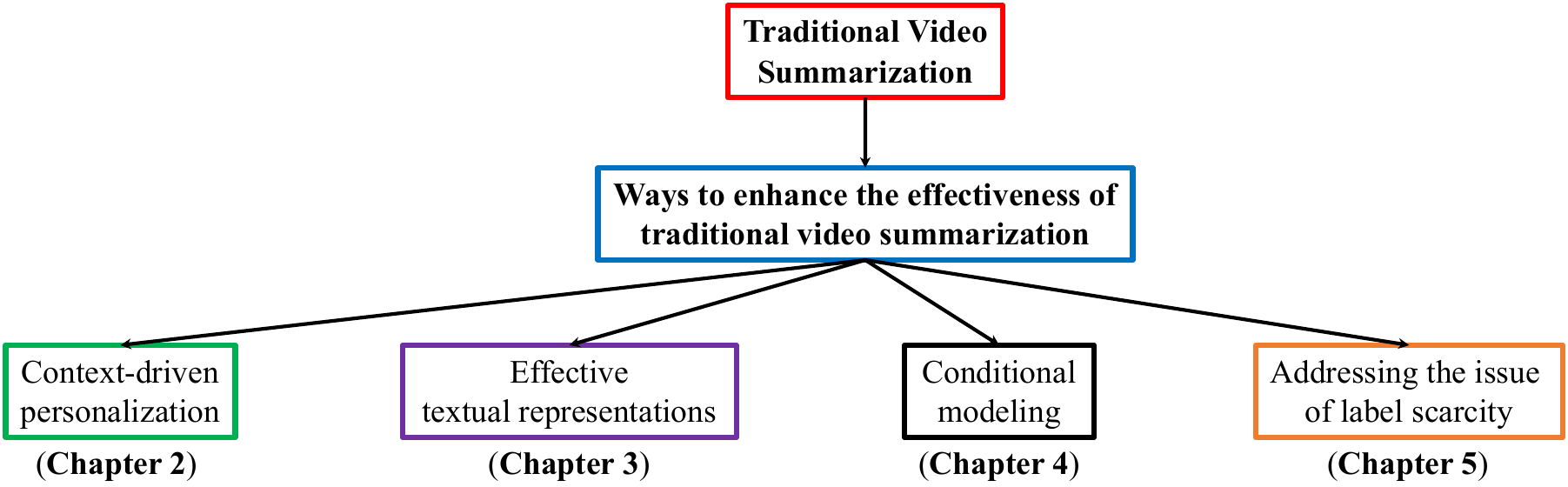}
\end{center}
\caption{\textbf{An overview of the thesis.}}
\label{fig_bias_dl}
\end{figure}

In this thesis, with reference to Figure \ref{fig_bias_dl} for an overview, we therefore pose the central research question:
\\\\
\centerline{\textit{\textbf{How can we enhance the effectiveness of automatic video summarization?}}}
\\\\
Traditional video summarization methods~\cite{gong2014diverse,gygli2014creating,zhang2016video,zhao2017hierarchical,zhao2018hsa,zhang2019dtr,ji2019video,ji2020deep} rely solely on visual information which has several limitations. Firstly, visual information alone may not provide sufficient context to fully comprehend the content of a video. For example, a cooking video may include visual instructions as well as textual descriptions, and relying solely on visual information may result in an incomplete summary. Secondly, visual information alone may not capture the emotional content of a video. For instance, a news video may contain significant emotional cues conveyed through textual language, descriptions, or grammar. This underscores the importance of text-based information related to a given video. Depending solely on visual information may overlook these critical cues, leading to an ineffective summary. To improve the effectiveness of video summarization, we propose to leverage text-based queries as a starting point to generate query-dependent video summaries that better align with the users' requirements.
Therefore, we pose the research question: 
\\\\
\centerline{\textit{\textbf{How to improve video summarization using text-based queries as context?}}}
\\\\
Addressing the challenge of efficiently navigating extensive video collections both within and across videos, traditional video summarization methods such as \cite{gong2014diverse,gygli2014creating,zhang2016video,zhao2017hierarchical,zhao2018hsa,zhang2019dtr,ji2019video,ji2020deep} have played a crucial role. A limitation of these approaches is their tendency to produce fixed summaries that may not consistently cater to the specific information requirements of users. In Chapter 2, we delve into this issue and investigate how video summarization techniques can be adapted to address it. 
Our approach involves developing a method that can generate a summary of a video tailored to a specific user's information needs by incorporating both textual queries and visual information extracted from the video. 
The primary challenge is to generate a summary that aligns with the user's needs.
We propose a multi-modal deep learning-based method that can handle both video and query information and generate an optimized summary based on the user's query. To integrate visual information with textual queries, we fuse visual and textual information at different levels of the model architecture. Evaluating the performance of the model is crucial, and we use metrics such as accuracy or F1-score to measure the quality of the generated summary. 
By modeling video summarization as a supervised learning problem, we develop an end-to-end deep learning-based approach consisting of a video summary controller, generator, and output module. To support our experiments, we introduce a new dataset with frame-based relevance/importance score labels. 

In the context of multi-modal video summarization, the text-based query plays a vital role as a controller that greatly influences the quality of the generated video summary. Therefore, it is essential to devise effective methods for encoding the given text-based query. This leads to the next research question:
\\\\
\centerline{\textit{\textbf{How to enhance the effectiveness of text-based query representations?}}}
\\\\
Multi-modal video summarization, involving the simultaneous utilization of visual and textual information, is a promising field of research that aims to enhance the usefulness and effectiveness of video summaries. This approach leverages diverse information sources to craft more comprehensive and insightful summaries that encapsulate both the content and emotional nuances of the original video. 
One pertinent application of multi-modal video summarization is in documentary films, where the combination of visual storytelling and narrated textual descriptions plays a crucial role in enhancing the viewer's comprehension and emotional engagement.
Similarly, it can enhance the summarization of educational videos by incorporating text annotations to provide crucial context and explanations. In both scenarios, amalgamating information from multiple sources contributes to the development of more comprehensive and informative summaries. The effectiveness of a multi-modal video summarization method heavily relies on a proficient encoding approach for text-based queries in generating the video summary.
To that end, in Chapter 3, a novel approach is introduced, leveraging a specialized attention network and contextualized word representations to proficiently encode both the text-based query and the video content. The proposed model integrates a contextualized video summary controller, multi-modal attention mechanisms, a specially designed attention network, and a video summary generator. In contrast to conventional text-based query encoding methods, such as Bag of Words (BoW), the utilization of contextualized word representations enables the effective capture of semantic meaning in the text-based query. Consequently, this approach enhances the overall effectiveness of text-based query representations.
Experimental results show that the proposed model outperforms the state-of-the-art method.

When creating a video summary, humans take into account both concrete factors, such as visual coherence, as well as abstract factors, such as the interestingness or the smoothness of the storyline. Therefore, a video summary created by humans is based on various factors.
Video summarization aims to shorten videos automatically while retaining the essential information to convey the overall story. However, current video summarization techniques rely mainly on visual factors such as visual consecutiveness and diversity, which may not be sufficient to fully comprehend the content of the video. To generate high-quality video summaries, non-visual factors such as interestingness, representativeness, and storyline consistency should also be considered. This raises the next research question:
\\\\
\centerline{\textit{\textbf{How to design a video summarization model that performs more like a human?}}}
\\\\
When crafting video summaries, humans take into account diverse elements including the storyline, contextualization, interestingness, and representativeness. In Chapter 4, a new approach to video summarization is proposed that integrates insights gained from human-created high-quality video summaries. This approach employs a conditional modeling perspective, introducing multiple meaningful random variables and joint distributions to capture the key components of video summarization. Helper distributions are utilized to improve model training, and a conditional semantics extractor is designed to mitigate potential performance degradation caused by extra input. The proposed method shows superior performance compared to existing methods
on commonly used video summarization datasets. But there is more than performance. The lack of explainability in automated machine learning-based decision-making systems, including video summarization, often hinders trust and acceptance. The proposed method is associated with a conditional/causal graph for video summarization, which makes it more explainable than existing methods. The experimental results demonstrate that the proposed approach achieves state-of-the-art performance in terms of the F1-score.

In this thesis, the video summarization task is treated as a fully supervised learning problem which requires a significant amount of human-labeled data for training. However, obtaining a sufficient amount of labeled data can be challenging and expensive, leading to the issue of data scarcity. Therefore, the final research question arises:
\\\\
\centerline{\textit{\textbf{How to tackle data scarcity in human-labeled data for video summarization?}}}
\\\\
The performance of supervised deep video summarization models is limited due to the expensive and small manually labeled query-based video summarization datasets. To overcome the challenge of data sparsity, self-supervision uses a pretext task to obtain additional data with pseudo labels to pre-train a supervised deep model. In Chapter 5, we propose a segment-level pseudo-labeling approach to model the relationship between the pretext task and the target task and the implicit relationship between the pseudo-label and the human-defined label. A pretext task is a surrogate or auxiliary task that is used to generate labels for training a model. The purpose of a pretext task is to design a task that requires the model to learn useful representations of the data without using external labels. Instead of relying on human-annotated labels, the pretext task leverages inherent patterns or structures within the data. 
The pseudo-labels are generated from existing frame-level labels provided by humans. To generate more accurate query-dependent video summaries, we introduce a semantics booster that produces context-aware query representations. Additionally, we propose a mutual attention mechanism to capture the mutual information between visual and textual modalities. We thoroughly validate our approach on three commonly used video summarization benchmarks, and our experimental results show that our algorithm achieves state-of-the-art performance.

\vspace{+6pt}\noindent In conclusion, this thesis concentrates on incorporating text-based queries as an additional input modality to improve the effectiveness of automatic video summarization and thus create personalized video summaries.

\chapter{Query-dependent Video Summarization}
\label{ch:spectral}

\section{Abstract}
Navigating through extensive video collections presents a daunting challenge in efficiently exploring content within individual videos and across them. Video summarization serves as a remedy to this issue by distilling video content to its core elements. However, conventional video summarization methods fall short of facilitating effective video exploration as they produce a single fixed video summary for each input video, disregarding the specific information requirements of the user. In Chapter 2, we formulate video summarization as a supervised learning problem and propose an end-to-end deep learning-based approach aimed at addressing this limitation. Our method leverages text-based queries to generate query-dependent video summaries, thus providing users with personalized summaries tailored to their information needs. The proposed model encompasses a video summary controller, video summary generator, and video summary output module. We introduce a dataset with frame-based relevance score labels to facilitate research in this area and validate the proposed method. Experimental results demonstrate the efficacy of our method in generating query-dependent video summaries. Notably, the integration of text-based queries enhances summary control and boosts overall model performance by $5.83\%$ in accuracy.

\section{Introduction} 
Video data has become ubiquitous in our daily lives, yet the majority of raw videos tend to be excessively long and contain redundant content. Consequently, the sheer volume of video data that individuals must sift through can be overwhelming, presenting new challenges in efficiently navigating both within and across videos. Video summarization techniques \cite{gong2014diverse,zhang2016summary,zhang2019dtr,zhou2018deep,vasudevan2017query} offer a solution by distilling the essence of a video, aiding in efficient exploration. 

Numerous approaches have been proposed to model the video summarization problem, including supervised methods \cite{gong2014diverse,gygli2014creating,li2018local,rochan2018video,zhang2016summary,zhang2016video,zhang2018retrospective,zhang2019dtr,zhou2018deep} and unsupervised methods \cite{de2011vsumm,chu2015video,kang2006space,lee2012discovering,liu2002optimization,ma2002user,panda2017collaborative,ngo2003automatic,potapov2014category,rochan2019video,zhao2014quasi,liu2020interpretable,gygli2015video}.
However, traditional video summarization methods, exemplified by \cite{ngo2003automatic,song2015tvsum,de2011vsumm,chu2015video,kang2006space,lee2012discovering,gygli2014creating,yang2019causal}, typically generate a single fixed video summary for a given input video. Specifically, they either create a summary attempting to cover all possible information needs and therefore yield limited reduction in time, or they lose essential information for specific needs. Consequently, the inflexibility of fixed summaries impedes the enhancement of video exploration efficacy.

\begin{figure}[t!]
\begin{center}
\includegraphics[width=1.0\linewidth]{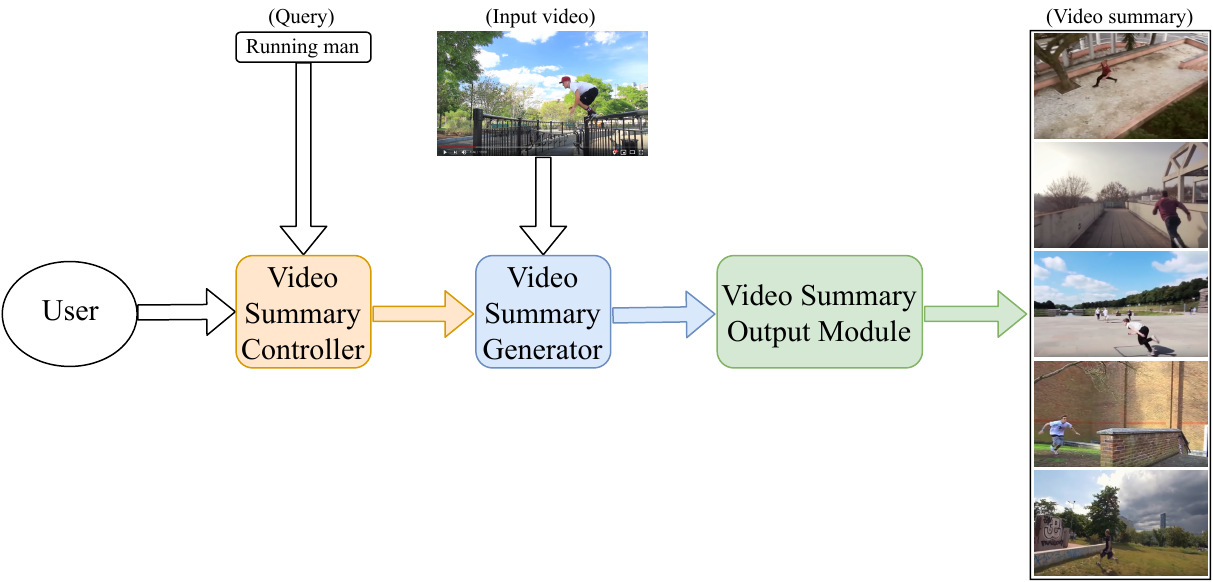}
\end{center}
   \caption{The concept of query-controllable video summarization. In query-controllable video summarization, a user provides a text-based query representing the desired video summary, directing the video summary generator to produce a summary aligned with the query. Notably, this method takes both the query and video as input. Consequently, when the same video input is paired with different queries, distinct query-dependent video summaries are generated. This differs from traditional video summarization, which relies solely on video input. For detailed information about the video summary controller and video summary generator, please refer to Figures \ref{fig:controller_text_only} and Figure \ref{fig:generator_new}.}
\label{fig:flowchart_text_only}
\end{figure}

To optimize the effectiveness and efficiency of video exploration, it is crucial to tailor the determination of essential elements in a video summary to the specific information needs of the user. Consequently, in this chapter, we introduce an end-to-end deep learning-based model, illustrated in Figure \ref{fig:flowchart_text_only}, designed to generate video summaries based on text-based input queries. This method, steered by the user's information needs, is capable of producing diverse video summaries for a given video. It is worth noting that non-end-to-end methods often entail numerous preprocessing steps, which can significantly diminish the efficiency of video exploration in practice \cite{dementhon1998video,gong2000video,ngo2003automatic,ngo2005video,lienhart1999dynamic,yu2003video,nam1999dynamic}.
The proposed model comprises a video summary controller, a video summary generator, and a video summary output module. The controller utilizes textual descriptions, such as words, phrases, or sentences, to articulate the desired video summary. Subsequently, the generator creates a video summary based on the implicit relationship between the text-based description and the input video. Finally, the video summary output module produces a video summary using a relevance score prediction vector.

The approach proposed in this chapter, termed ``query-controllable video summarization'', introduces distinct challenges compared to the well-explored domain of conventional video summarization. A key divergence lies in the incorporation of a text-based query input alongside the video input, contrasting with conventional video summarization that relies solely on the video input. Consequently, in query-controllable video summarization, modeling the implicit relations or interactions between the input query and video becomes imperative.
Our proposed model takes both a text-based query and video as input, making effective fusion of multi-modal features crucial while minimizing information loss. Previous studies have demonstrated that in similar multi-modal contexts, model performance can decline if feature fusion methods are not appropriately chosen. Addressing this issue remains an open question \cite{ben2017mutan,fukui2016multimodal}. Thus, in our proposed method, we explore three widely used feature fusion methods to construct our query-controllable system for videos and identify the one that yields the optimal performance.

Traditionally, the assessment of video summarization models often relies on human expert evaluation, involving predefined rules or a direct comparison of two different video summaries to select the superior one \cite{over2008trecvid,lu2013story,lee2012discovering}. Human-expert-based evaluation methods for this task pose challenges due to their inherent drawbacks, being both costly and time-consuming. These methods depend on human judgments for each evaluation, contributing to their inefficiency \cite{gygli2014creating}.
To address this challenge, we model video summarization as a supervised learning task, enabling the automatic evaluation of model performance. Additionally, we establish a new video dataset to validate the effectiveness of our model. In this thesis, preference is given to automatic evaluation methods for their efficiency.

In our experiments, we explore not only how the input query influences our model's performance but also how commonly used feature fusion methods impact our model's effectiveness. We present the experimental results of the proposed end-to-end deep model, illustrating its capability to generate query-dependent and controllable video summaries for given videos. Our experimental findings indicate that the inclusion of a text-based input query significantly influences the control and performance of the video summary. Specifically, the text-based input query enhances the performance of our summarization model, demonstrating an accuracy improvement of $5.83\%$.

\vspace{+3pt}
\noindent\textbf{Contributions.}

\begin{itemize}
    \item \textbf{End-to-End Deep Learning-based Model for Query-Controllable Video Summarization}: We propose an end-to-end deep learning-based approach to video summarization, specifically designed to address the limitation of traditional methods that generate fixed video summaries. By formulating video summarization as a supervised learning problem, our model leverages text-based queries to generate query-dependent video summaries, thus providing users with personalized summaries tailored to their specific information needs.
    
    \item \textbf{Dataset Introduction and Method Validation}: To facilitate research in the area of query-controllable video summarization and validate our proposed method, we introduce a new dataset with frame-based relevance score labels. This dataset serves as a valuable resource for evaluating video summarization models and comparing their performance.
    
    \item \textbf{Enhanced Video Exploration Efficacy}: Our proposed approach significantly enhances the efficacy of video exploration by allowing users to navigate extensive video collections more efficiently. By generating query-dependent video summaries, our model provides users with summaries that are more relevant to their specific information needs, thus improving the overall user experience.
    
    \item \textbf{Improved Model Performance}: Experimental results demonstrate the effectiveness of our proposed end-to-end deep learning-based model in generating query-dependent video summaries. Notably, the integration of text-based queries into the summarization process leads to a substantial improvement in model performance, with an accuracy enhancement of $5.83\%$. This improvement underscores the importance of considering user-specific information needs in video summarization tasks.
\end{itemize}

The subsequent sections of this chapter are organized as follows: In Section 2.2, we conduct a comprehensive review of related works. Section 2.3 outlines the specifics of our proposed dataset. Moving on to Section 2.4, we introduce our proposed method, delving into its design choices and implementation details. Lastly, in Section 2.5, we assess the performance of our approach.

\section{Related Work}
This section delves into the relevant literature, examining various methodologies and datasets employed in the field. Initially, we explore video summarization methods, categorizing them into two primary types: unsupervised and supervised approaches. Subsequently, we conduct a comprehensive review of prevalent video summarization datasets.

\subsection{Unsupervised Video Summarization}
Unsupervised methods in video summarization, as exemplified by \cite{de2011vsumm,chu2015video,kang2006space,lee2012discovering,lu2013story,liu2002optimization,ma2002user,panda2017collaborative,ngo2003automatic,potapov2014category,rochan2019video,zhao2014quasi,mahasseni2017unsupervised,plummer2017enhancing,gygli2015video}, typically rely on handcrafted heuristics to fulfill specific properties such as interestingness, representativeness, and diversity.

In \cite{de2011vsumm}, the authors propose a technique based on extracting color features from video frames and employing k-means clustering. In the work by \cite{chu2015video}, the authors note the recurrence of key visual concepts across videos with similar topics. They suggest generating a summarized video by identifying segments that frequently co-occur across videos, utilizing a Maximal Biclique Finding algorithm to identify sparsely co-occurring patterns. The approach presented in \cite{kang2006space} introduces a space-time video summarization method, extracting visually informative space-time segments from input videos and concurrently analyzing the distribution of spatial and temporal information. Additionally, \cite{lee2012discovering} introduces a video summarization technique tailored for egocentric camera data. The authors create region cues, such as proximity to hands, gaze, and occurrence frequency, in egocentric video and develop a model to predict the relative importance of a new region based on these cues. 

In \cite{lu2013story}, the authors introduce a method for video summarization aimed at uncovering the narrative of an egocentric video. Their approach identifies a concise sequence of sub-segments within the video that depict crucial events. In \cite{ma2002user}, the authors propose a generic framework for video summarization based on modeling viewer attention. This framework, devoid of full semantic content understanding, circumvents the need for complex heuristic rules in video summarization tasks by leveraging computational attention models. \cite{ngo2003automatic} presents a unified video summarization method centered on structural analysis and video highlights. Emphasizing content balance and perceptual quality, this approach incorporates a normalized cut algorithm for video partitioning and a motion attention model based on human perception for assessing perceptual quality. The authors of \cite{panda2017collaborative} devise a novel method to extract a video summary that simultaneously captures distinctive features unique to a given video and general characteristics observed across a set of videos. Finally, in \cite{gygli2015video}, the authors propose a new model for learning the importance score of global video summary characteristics. By jointly optimizing for multiple objectives, this model demonstrates the capability to produce high-quality video summaries.

In \cite{zhao2014quasi}, the authors present a method for learning a dictionary from a given video using group sparse coding. Subsequently, a video summary is generated by amalgamating segments that cannot be sparsely reconstructed using this dictionary. \cite{rochan2019video} introduces a novel formulation for video summarization from unpaired data. Their model aims to learn a mapping from a set of raw videos to a set of video summaries, ensuring that the distribution of the generated video summary aligns with the distribution of the set of video summaries by employing an adversarial objective. Additionally, they impose a diversity constraint on the mapping to ensure visually diverse generated video summaries. In \cite{zhou2018deep}, the authors propose an end-to-end reinforcement learning-based framework for training their video summarization model. This framework incorporates a new reward function that jointly considers the diversity and representativeness of the generated video summaries. Importantly, the design of this reward function eliminates the need for user interactions or labels.

While numerous existing works frame the video summarization task as an unsupervised problem and propose various methods to address it, the overall performance of unsupervised approaches tends to lag behind supervised ones. This discrepancy arises from the superior ability of supervised methods to capture nuanced cues from ground truth video summaries, a feat that is often challenging with handcrafted heuristics. In this chapter, we depart from the unsupervised paradigm and instead model video summarization as a supervised learning problem.

\subsection{Supervised Video Summarization}
Supervised methods represent another approach to video summarization, relying on human expert-labeled data to train the models \cite{gong2014diverse,gygli2014creating,li2018local,rochan2018video,sharghi2018improving,zhang2016summary,zhang2016video,zhang2018retrospective,zhang2018query,zhang2019dtr,zhou2018deep,sharghi2017query}. These methods learn from ground truth video summaries provided by human experts, enabling them to capture more nuanced information compared to unsupervised approaches. 

The authors of \cite{gong2014diverse} approach video summarization as a supervised subset selection task, introducing the sequential determinantal point process (SeqDPP) as a probabilistic model for this purpose. Unlike the standard DPP, which treats video frames as randomly permutable elements, SeqDPP acknowledges the inherent sequential structures in video data. Consequently, it not only addresses the limitations of the standard DPP but also retains the ability to model diverse subsets, crucial for effective video summarization. In \cite{gygli2014creating}, the focus shifts to user videos containing interesting events, where the method begins by segmenting the video based on a tailored superframe segmentation. Utilizing a range of low-level, mid-level, and high-level features, the method then selects an optimal subset of superframes to generate a video summary based on the estimation score of visual interestingness per superframe. In \cite{li2018local}, a new probabilistic model, built upon SeqDPP, is introduced to address the video summarization problem. This model allows for dynamic control throughout a video segment, where local diversity is imposed, and is trained using a reinforcement learning algorithm to ensure optimal performance. Formulating video summarization as a sequence labeling problem, the authors of \cite{rochan2018video} propose fully convolutional sequence models. They establish a novel connection between video summarization and semantic segmentation and utilize adapted semantic segmentation networks to generate video summaries. Finally, in \cite{sharghi2018improving}, an enhanced SeqDPP model is proposed, featuring a new probabilistic distribution integrated into SeqDPP, which allows the resulting model to accept user input regarding the intended length of the video summary, thus improving user control over the summarization process. 

In addressing exposure bias within SeqDPP, a large-margin algorithm is proposed, as detailed in \cite{zhang2016summary}, where the authors introduce a subset selection method leveraging human-created video summaries for keyframe-based video summarization. The core concept involves transferring summary structures nonparametrically from annotated videos to unseen ones, with an extension to sub-segment-based summarization also explored. Meanwhile, in \cite{zhang2016video}, the video summarization task is reformulated as a structured prediction problem on sequential data. A novel supervised learning technique incorporating Long Short-Term Memory is proposed to capture variable-range dependencies inherent in the task, with domain adaptation techniques based on auxiliary annotated video datasets used to enhance summary quality. In a sequence-to-sequence learning model proposed in \cite{zhang2018retrospective}, a retrospective encoder is introduced to augment standard sequence learning models, measuring whether the generated summary preserves the original video's information. This model adds a new loss to complement discriminative losses, comparing embeddings of the predicted summary and the original video in an abstract semantic space. Finally, in \cite{zhang2019dtr}, a dilated temporal relational generative adversarial network (DTR-GAN) is proposed for frame-based video summarization. Utilizing an adversarial approach with a three-player loss, DTR-GAN learns a dilated temporal relational generator and discriminator. A new dilated temporal relational unit is introduced to enhance temporal representation capture, facilitating the generation of keyframes by the generator. 

Modeling the video summarization task as a supervised learning problem offers the advantage of enabling efficient automatic evaluation. In this chapter, the proposed query-controllable video summarization approach is based on supervised learning, facilitating efficient automatic evaluation.

\subsection{Video Summarization Dataset Comparison}
We provide a brief overview of widely used video summarization datasets, namely the TVSum and SumMe datasets \cite{song2015tvsum,gygli2014creating}, and draw comparisons with the dataset employed in this chapter. 

The TVSum dataset, proposed by the authors of \cite{song2015tvsum}, consists of $50$ videos spanning $10$ categories, each associated with segment-level importance scores obtained through crowdsourcing. These categories are selected from the TRECVid Multimedia Event Detection task \cite{smeaton2006evaluation}, and the videos ($5$ per category) are sourced from YouTube using category names as search queries. Video selection adheres to several criteria: (i) the video should comprise more than a single segment; (ii) its title should be descriptive of the visual content; (iii) it should be under the Creative Commons license; and (iv) the duration should range from $2$ to $10$ minutes. The authors leverage Amazon Mechanical Turk (AMT) to gather $1,000$ responses ($20$ per video), treating them as gold standard labels. AMT participants follow a protocol that involves reading the video title, simulating typical online video browsing, watching the entire video in one sitting, and assigning importance scores to uniform-length segments (ranging from $1$ for not important to $5$ for very important). The absence of audio ensures that importance scores are based solely on visual stimuli, and the chosen $2$-second segment length aligns with the authors' experience in capturing local context with good visual coherence. 

In \cite{gygli2014creating}, the authors present another benchmark for video summarization known as SumMe. This benchmark comprises $25$ videos encompassing holidays, events, and sports, with varying lengths ranging from $1$ to $6$ minutes. Each video is summarized by $15$ to $18$ different individuals, with participation from $19$ males and $22$ females. Participants were tasked with creating video summaries that encapsulate the most important content from the original videos. They were provided with a simple interface to watch, cut, and edit the videos. The length of the video summaries was required to be between $5\%$ and $15\%$ of the original video length to ensure effective summarization rather than mere shortening. To maintain consistency, the videos were presented randomly, and the audio was muted during the summarization process to focus solely on visual stimuli. 
Previous evaluations of video summarization methods have traditionally relied on human expert assessment, typically involving predefined criteria such as the degree of redundancy and the inclusion of crucial content, or by presenting human experts with two distinct video summaries and asking them to choose the superior one \cite{over2008trecvid, lu2013story, lee2012discovering}. The conventional human expert evaluation methods, though widely used, pose significant challenges due to their inherent drawbacks \cite{gygli2014creating}. Specifically, these methods rely on human judges for each evaluation, rendering them expensive and time-consuming. For instance, in \cite{lu2013story}, the evaluation process demanded an entire week of human labor. While these methods assist in determining which video summary is superior, they do not inherently demonstrate what constitutes an ideal video summary. Consequently, \cite{gygli2014creating} refrains from employing these approaches. Instead, they enlist a group of participants to create their video summaries and gather multiple summaries for each video. This approach is adopted because there is no definitive answer for video summarization but multiple potential approaches. By utilizing human expert-generated video summaries, they can systematically compare any summarization method that generates automatic video summaries in a repeatable and efficient manner. In prior works such as \cite{khosla2013large,de2011vsumm}, such automatic versus human comparisons have been successfully employed for keyframes. Additionally, \cite{khosla2013large} demonstrates that comparing automatic keyframe-based summaries to human keyframe-based selections yields ratings comparable to those obtained when humans directly assess the automatic video summaries.

Both the TVSum and SumMe datasets facilitate the automatic evaluation of video summarization approaches. In this chapter, we introduce a new dataset, derived from an existing dataset proposed by \cite{vasudevan2017query}, designed for the automatic evaluation of our query-controllable video summarization task. The newly proposed dataset comprises $190$ videos with frame-level relevance score annotations. For clarity and comparison with existing datasets, we provide a summary in Table \ref{table:table1_chapter2}.

\begin{table}[ht!]
    \caption{Comparison of commonly used video summarization datasets. It becomes evident that the proposed dataset significantly surpasses others in terms of size and complexity, featuring both video and text input modalities. In contrast, the other datasets are exclusively comprised of video data and are relatively smaller in size. Therefore, the proposed dataset stands out as unique in its composition and scale.}
\begin{center}
\rotatebox{270}{
\scalebox{0.75}{
    \begin{tabular}{c|c|c|c|c}
    \toprule
    \textbf{Name of Dataset} & \textbf{Annotation Type} & \textbf{Content} & \textbf{Number of Videos} & \textbf{Input Modality} \\ 
    \midrule
    SumMe \cite{gygli2014creating} & Interval-based segment and frame-level scores & User videos & 25 & Video \\ 
    \midrule
    TVSum \cite{song2015tvsum} & Frame-based important scores & YouTube videos & 50 & Video \\ 
    \midrule
    \rowcolor{mygray} QueryVS (Ours) & Frame-based relevance scores & YouTube videos & \textbf{190} & Video, Text \\ 
    \bottomrule 
    
    \end{tabular}}
}

    \label{table:table1_chapter2}
\end{center}
\end{table}

\section{Dataset Introduction and Analysis}
In this section, we delve into the exploration and analysis of our proposed dataset for query-controllable video summarization. Our examination encompasses an in-depth investigation into the types of videos, the associated video labels, and the presentation of various statistical insights into the dataset. It's worth noting that, while the dataset introduced by \cite{vasudevan2017query} partially aligns with our research objectives and is publicly accessible, we have identified disparities in specifications, such as annotations and the number of videos, when comparing the published dataset with the details outlined in \cite{vasudevan2017query}. Furthermore, certain segments of the initially published dataset are no longer accessible. Consequently, we will first elucidate the process used by \cite{vasudevan2017query} to create the dataset and subsequently highlight the necessary modifications required to tailor it to our specific objectives.

\subsection{Setup}
As our dataset is derived from \cite{vasudevan2017query}, the data collection rules closely align with those outlined in \cite{vasudevan2017query}. The proposed dataset comprises $190$ videos, each retrieved based on a given text-based query. Following the methodology of \cite{vasudevan2017query}, video frames are annotated with labels indicating the relevance scores to the text-based queries, a technique reminiscent of the labeling approach in the MediaEval diverse social images challenge \cite{ionescu2014retrieving}. The primary objective of these human-expert annotated labels in the proposed dataset is to facilitate the automatic evaluation of methods for generating relevant and diverse video summaries.

In constructing the proposed dataset based on \cite{vasudevan2017query}, representative samples of queries and videos are gathered through the following procedure: seed queries, spanning $22$ different categories, are selected from the top YouTube queries between $2008$ and $2016$. Given the generic and concise nature of these queries, the YouTube auto-complete function is utilized to obtain more realistic and extended queries, such as ``ariana grande focus instrumental'' and ``ark survival evolved dragon''. For each query, the top video result with a duration of $2$ to $3$ minutes is collected.

The annotation task on AMT, as formulated by \cite{vasudevan2017query}, involves sampling all $190$ videos at one frame per second (fps). Each frame is annotated by an AMT worker with respect to its relevance to the given text-based query. The potential responses include ``Very Good'', ``Good'', ``Not good'', and ``Bad'', where ``Bad'' indicates that the frame is irrelevant and of low quality, marked by factors like poor contrast or blurriness. To mitigate subjectivity in labeling, every video in the proposed dataset is annotated by at least $5$ different AMT workers. A qualification task is defined to ensure high-quality annotations, with results manually reviewed to guarantee the workers provide annotations of good quality. Once workers pass this task, they are eligible for further assignments.

Given that the maximum number of frames for a video is $199$, an adjustment is made to ensure uniformity across all videos. Frames are repeated, starting from the first frame, until reaching a total of $199$ frames for each video, akin to the approach in \cite{sigurdsson2017asynchronous}. Figure \ref{fig:original_number_of_frames_test} illustrates the original number of frames for each video.

\subsection{Crowd-sourced Annotation}

In this subsection, we analyze the annotations of frame-based relevance scores acquired through the aforementioned procedure. Additionally, we elucidate the process by which we consolidate these relevance score annotations for each video into a unified set of ground truth labels.

\noindent\textbf{Label Distributions of Relevance Scores}:
The distribution of relevance score annotations is as follows: ``Very Good'': $18.65\%$, ``Good'': $55.33\%$, ``Not good'': $13.03\%$, and ``Bad'': $12.99\%$.

\noindent\textbf{Ground Truth}:
As discussed in the \textit{``Video Summarization Dataset Comparison''} subsection, human-based evaluation poses challenges in terms of both reliability and time consumption \cite{gygli2014creating}. Therefore, in this chapter, we adopt the evaluation approach outlined in \cite{vasudevan2017query}. For the assessment of testing videos, one method involves soliciting human experts to watch the entire video (rather than just the summaries) and evaluate the relevance of each part of the video. The responses of these experts serve as gold standard annotations \cite{gygli2014creating,khosla2013large,potapov2014category}. 
This approach offers the advantage of obtaining annotations that can be utilized for endless experiments, a crucial aspect in the development and testing of computer vision systems with multiple iterations. In our proposed dataset, we consolidate the relevance score annotations from AMT workers into a single ground truth relevance score label for each query-video pair. We employ the majority vote rule \cite{antol2015vqa,huang2019novel} for evaluating model performance in relevance score prediction. In this context, a predicted relevance score is considered correct if it aligns with the majority of human annotators who provided that specific score. In our dataset, we assign numerical values to annotations as follows: ``Very Good'' to 3, ``Good'' to 2, ``Not Good'' to 1, and ``Bad'' to 0.

It's important to highlight that, as shown in \ref{table:table1_chapter2}, the relevance score in this chapter differs from the importance score in TVSum \cite{song2015tvsum}. The relevance score in this chapter specifically aims to capture the relationship between a given text-based query and a video frame, while the importance score in TVSum is designed to reflect the importance of a video frame in the context of the final video summary \cite{song2015tvsum}.

\section{Method}

In this section, we introduce the proposed method for query-controllable video summarization. The method consists of three main components: the video summary controller, video summary generator, and video summary output module. The summary controller receives a text-based query as input and produces the vector representation of the query. Subsequently, the summary generator takes both the embedded query and a video as inputs, generating frame-based relevance score predictions. Finally, the video summary output module utilizes these score predictions to generate the ultimate video summary. Refer to Figure \ref{fig:flowchart_text_only} for a visual representation of the described procedure.

\subsection{Video Summary Controller}
\begin{figure}[t!]
\begin{center}
\includegraphics[width=1.0\linewidth]{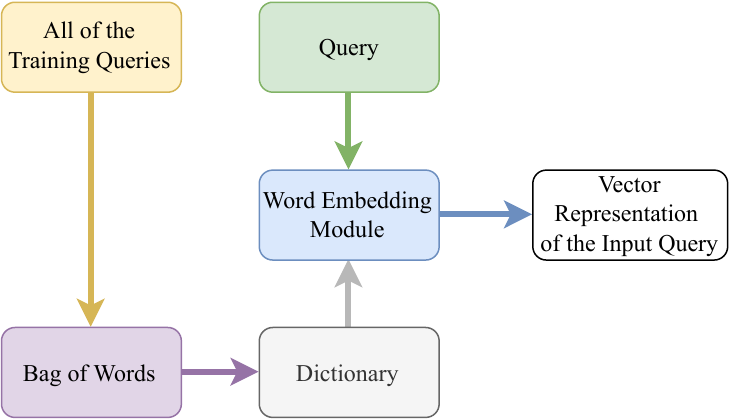}
\end{center}
   \caption{Illustrating the concept of our video summary controller. Within the controller, we gather all queries from our training set to construct a bag of words, subsequently forming a dictionary. This dictionary serves as the basis for embedding an input query.}
\label{fig:controller_text_only}
\end{figure}

Text-based queries serve to depict the desired content of the video summary while subtly conveying their semantic relationship. Therefore, we adopt a specific approach to encode input queries and incorporate their influence into our proposed method. In our approach, we leverage the vector representation of text-based input queries to govern the generated video summary. The essence of our video summary controller lies in generating a vector representation of an input query based on a pre-defined dictionary. Initially, we construct a dictionary using a bag of words extracted from all the unique words found in the training queries. Subsequently, we encode an input query using this dictionary, resulting in a vector representation that encapsulates the expected content of the video summary. To provide clarity on this procedure, we have created a flowchart illustrating the aforementioned steps, as depicted in Figure \ref{fig:controller_text_only}.

\begin{figure}
  \includegraphics[width=\textwidth]{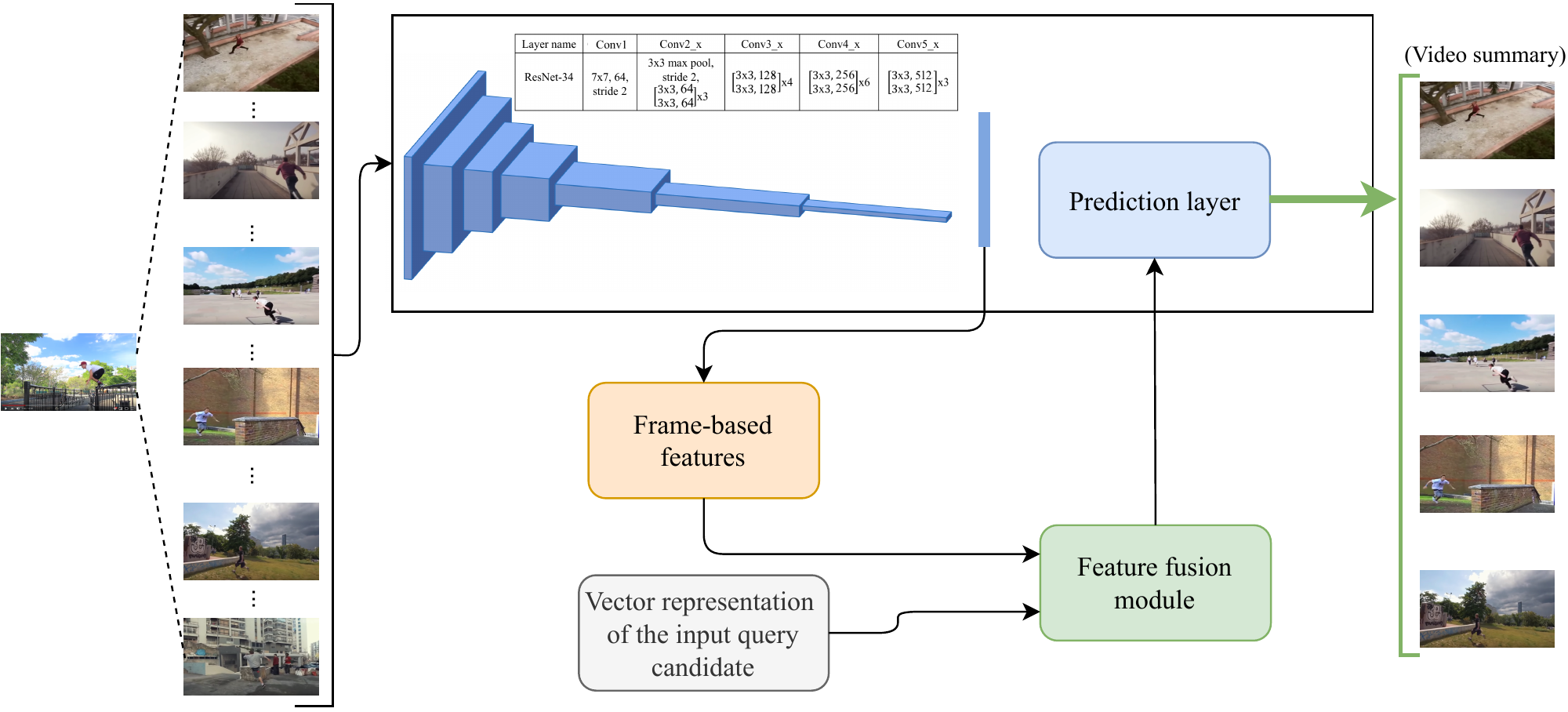}
  \caption{
    Illustration of the proposed video summary generator. The process begins by taking an input video and sampling it at a rate of $1$ fps. Subsequently, the sampled frames are fed into a CNN-based structure designed for extracting frame-based features. These features are then fused with the features derived from a text-based input query. Finally, the fused feature set is passed through a prediction layer to generate frame-based relevance scores. The predicted scores serve as the foundation for generating a query-dependent video summary.}
  \label{fig:generator_new}
\end{figure}

\subsection{Video Summary Generator}
The primary aim of the video summary generator is to leverage a vector representation of a text-based input query and a video input to generate a frame-based relevance score vector. This generator comprises a convolutional neural network (CNN) structure alongside a multi-modality feature fusion module. Notably, the CNN structure is trained using our designated training set. Initially, the input video is sampled at a rate of $1$ fps. We then utilize ResNet-34 \cite{he2016deep} to extract $199$ frame-based features for each input video, specifically from the visual layer immediately preceding the classification layer. After feature extraction, the features are fed through a feature fusion module designed to merge the frame-based features with the feature derived from the input text-based query. The resulting fused feature vector is subsequently passed to a fully connected layer to predict the frame-based relevance scores. Further details on the feature fusion module are provided in the subsequent subsection. For a visual representation of this process, please refer to Figure \ref{fig:generator_new}. It's worth mentioning that we adopt ``Cross-Entropy Loss'' as our loss function (Equation \ref{eq:loss_chapter_2}) and employ the Adam optimizer \cite{kingma2014adam}. The coefficients used for computing the moving averages of gradients and their squares are $\beta_{1}=0.9$ and $\beta_{1}=0.999$, respectively. Additionally, we introduce a term ($\epsilon=1e-8$) into the denominator to enhance numerical stability, while setting the learning rate to $\alpha=1e-4$.

\begin{equation}
    \label{eq:loss_chapter_2}
    \textup{Loss}(x, \textup{class}) = -x \left[\textup{class}\right] + \textup{ln}(\sum_{j}\textup{exp}(x \left[j \right])),
\end{equation}
where ``$\textup{class}$'' represents the ground truth class, while $x$ denotes the prediction.

\noindent\textbf{Multi-modality Features Fusion Module}:
A key technical challenge in our proposed method is effectively fusing query and frame-based features while minimizing information loss. In analogous multi-modal scenarios, it has been demonstrated that poorly designed models can lead to decreased performance, leaving the resolution of this issue a general open question \cite{ben2017mutan,fukui2016multimodal}. In this chapter, we explore three prevalent approaches—summation, concatenation, and element-wise multiplication—to integrate query and frame-based features.

\subsection{Video Summary Output Module}
Once we obtain the frame-based relevance score prediction vector from the video summary generator, it is forwarded to our video summary output module. The core concept behind this module is to generate a video summary based on the relevance score prediction vector. In our approach, we associate labels; specifically, ``Very Good'' is mapped to $3$, ``Good'' to $2$, ``Not Good'' to $1$, and ``Bad'' to $0$. If a predicted relevance score is $\ge 2$, we consider the corresponding frame as relevant. Conversely, if the predicted relevance score is $<2$, we deem the corresponding frame irrelevant. Subsequently, we compile $k$ relevant frames in chronological order to form our video summary. It's important to note that $k$ serves as a user-defined parameter, determining the length of the video summary.

\section{Experiments and Analysis}
In this section, we will assess the efficacy of our proposed end-to-end approach for the query-controllable video summarization task, utilizing the setup outlined in our proposed dataset. Additionally, we will conduct an in-depth analysis of the effectiveness of both the query and the methods employed for multi-modal feature fusion.

\begin{figure}[t!]
\begin{center}
\includegraphics[width=1.0\linewidth]{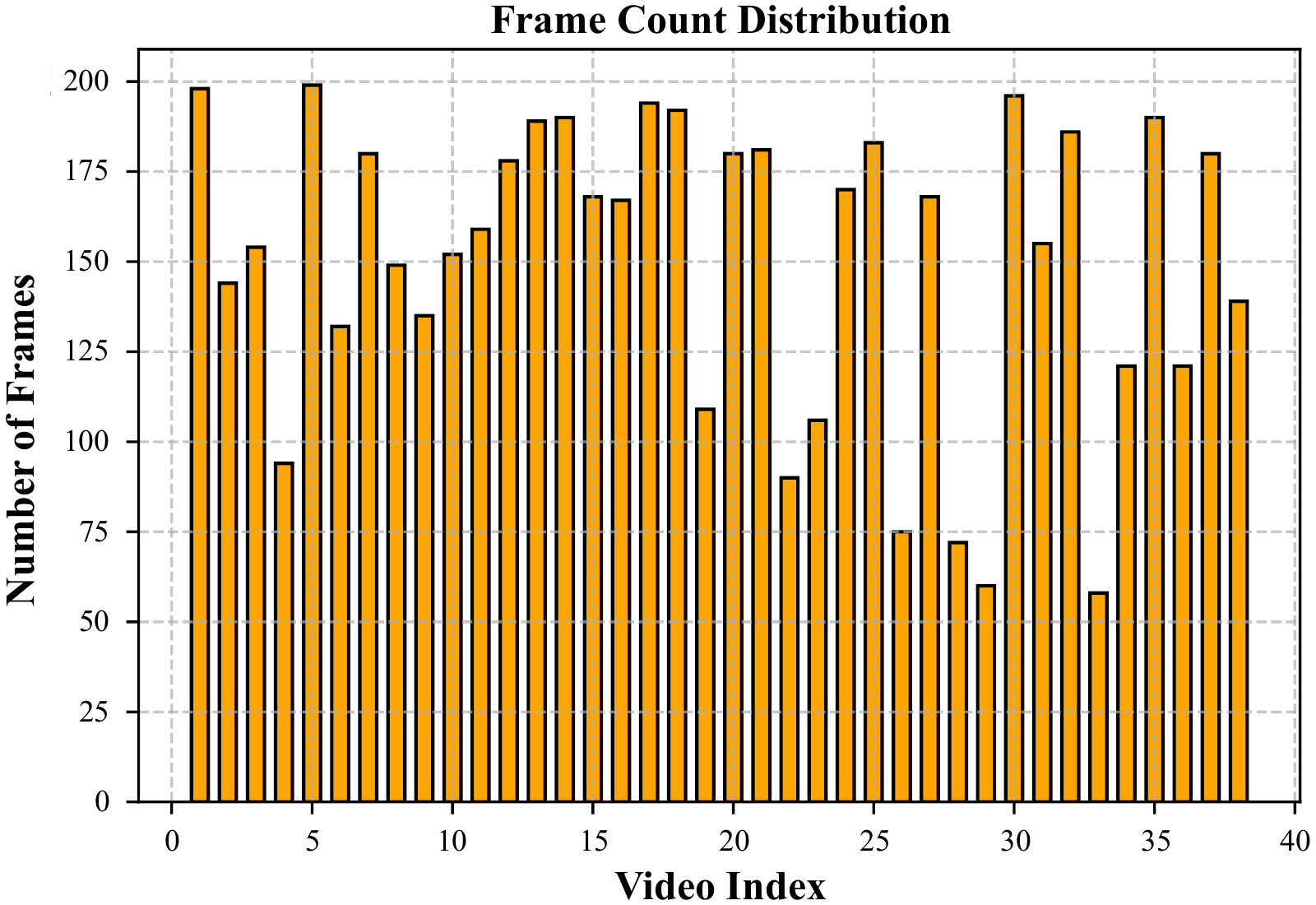}
\end{center}
   \caption{Initial frame count for each video. The x-axis represents the video index, while the y-axis indicates the original number of frames. It's important to note that, for the sake of convenience, we standardized all videos to have the same frame count during the development of our proposed method.
   }
\label{fig:original_number_of_frames_test}
\end{figure}

\subsection{Dataset preparation}
To validate our proposed query-controllable video summarization approach, we adhere to the specified dataset configuration for our experiments. The dataset is divided into training, validation, and testing sets, constituting $60\%$/$20\%$/$20\%$ (or $114$/$38$/$38$ videos, respectively). Each video is associated with a corresponding query, with a maximum query length of $8$ words. The input frame size for the CNN is set at $128 \times 128$ pixels with $3$ channels representing red, green, and blue. Image channels are normalized using $\textup{mean} = (0.4280, 0.4106, 0.3589)$ and $\textup{std} = (0.2737, 0.2631, 0.2601)$. The maximum number of frames per video is restricted to $199$. Following the video preprocessing method outlined in \cite{sigurdsson2017asynchronous}, we ensure uniformity in the number of frames for all videos, set to $199$. The original number of frames for each video is depicted in Figure \ref{fig:original_number_of_frames_test}.

\begin{figure}[t!]
\begin{center}
\includegraphics[width=1.0\linewidth]{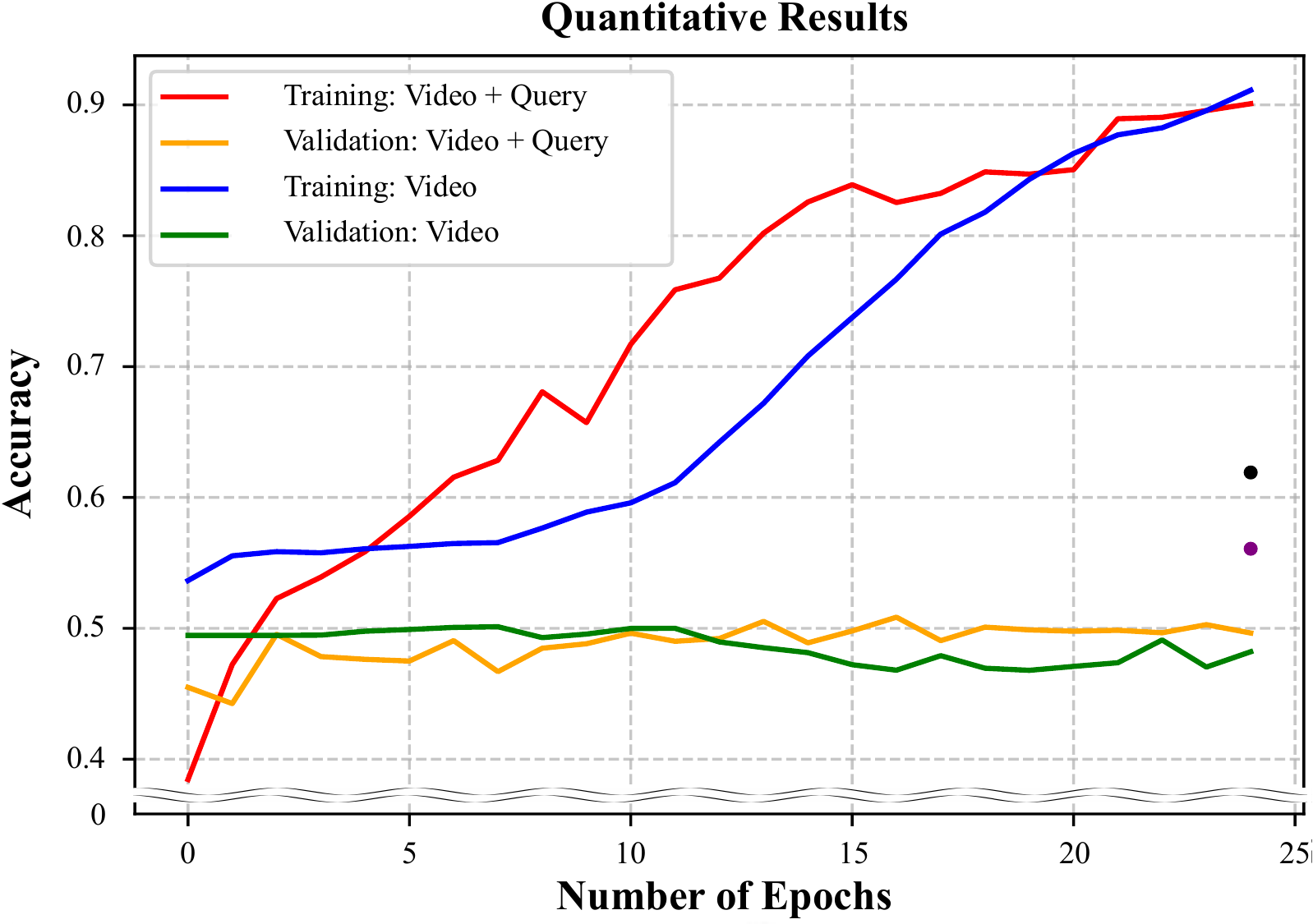}
\end{center}
   \caption{Quantitative analysis in two scenarios: ``video only'' versus ``video + query''. The x-axis represents the number of epochs, while the y-axis indicates the model accuracy. Testing is conducted after $25$ epochs of training. The ``purple point'' signifies the testing accuracy for the ``video only'' case, recorded at $0.5608$, while the ``black point'' represents the testing accuracy for the ``video + query'' case, measured at $0.6191$.}
\label{fig:acc_epochs_query_testing}
\end{figure}

\subsection{Effectiveness Analysis of Query}
\begin{figure}
\begin{center}
\includegraphics[width=1.0\linewidth]{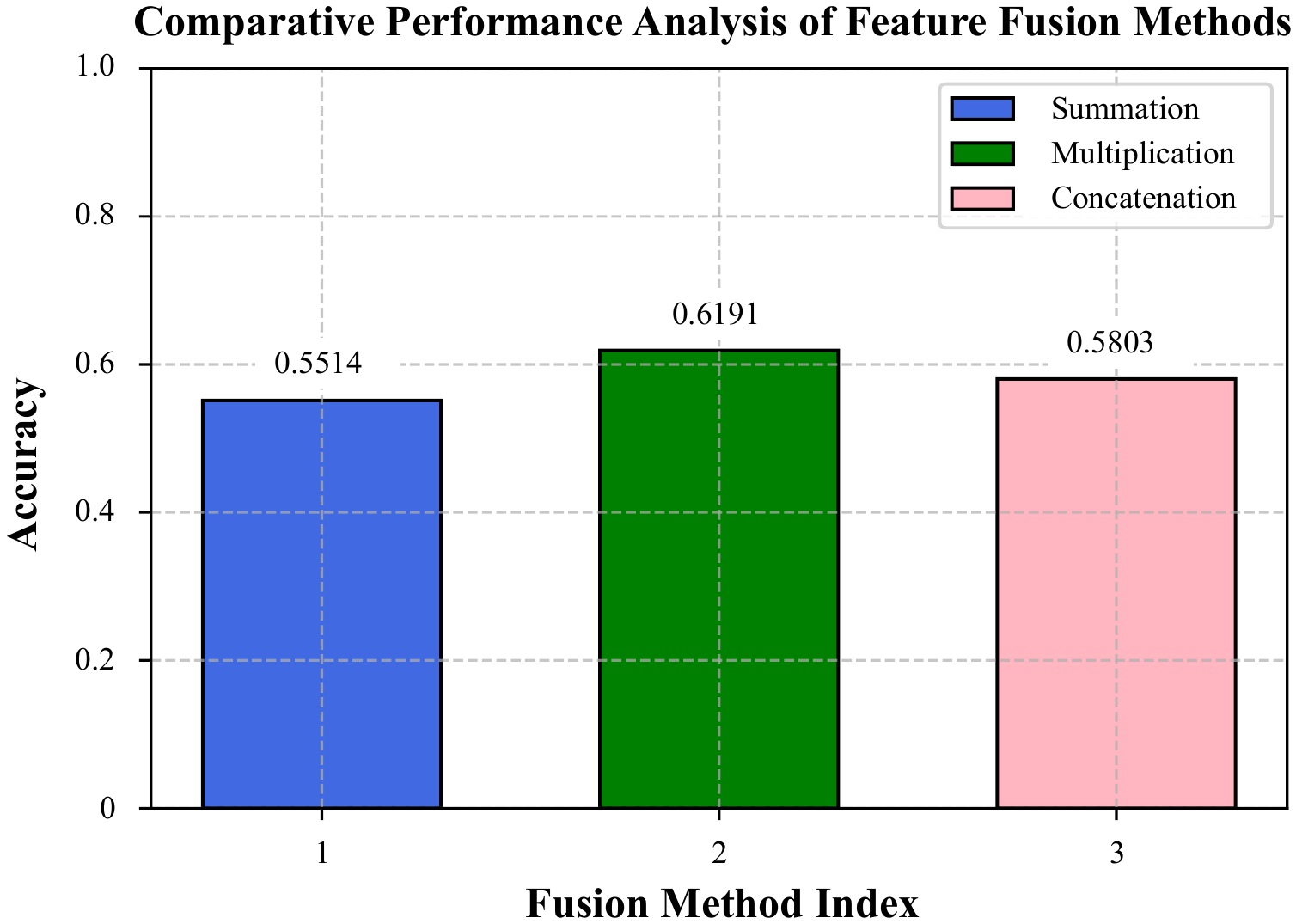}
\end{center}
   \caption{Quantitative analysis across three distinct scenarios: multi-modal feature fusion by ``summation'', ``concatenation'', and ``element-wise multiplication''. Notably, the model employing the element-wise multiplication fusion method demonstrates superior performance. The x-axis represents the case index, while the y-axis denotes the model accuracy. It's important to note that each model undergoes training for a varying number of epochs.}
\label{fig:fusion_method}
\end{figure}

In this experiment, we aim to assess the impact of text-based queries on the quality of generated video summaries. We compare two types of models: query-driven and non-query-driven. As illustrated in Figure \ref{fig:acc_epochs_query_testing}, the query-driven model exhibits superior testing accuracy ($0.6191$) compared to the non-query-driven model (testing accuracy: $0.5608$). Analyzing the validation accuracy across epochs reveals that the inclusion of a textual query guides the query-driven model to outperform its non-query-driven counterpart. However, when considering the worst-performing fusion model (summing feature vectors) in Figure \ref{fig:fusion_method} against the non-query-driven model, we observe that the non-query-driven model outperforms it. This observation motivates us to explore further in the subsequent experiment, comparing different multi-modal feature fusion methods, as detailed in the next subsection.

\subsection{Effectiveness Analysis of Different Fusion Methods}
In the realm of multi-modal feature fusion, the optimal approach remains an ongoing inquiry \cite{ben2017mutan,fukui2016multimodal}. To address this, we explore three commonly employed methods: summation, concatenation, and element-wise multiplication. As depicted in Figure \ref{fig:fusion_method}, our experimentation reveals that the model employing element-wise multiplication fusion exhibits the most favorable performance. Furthermore, compared to the non-query-driven model, the model utilizing the concatenation fusion method outperforms it. Conversely, the model employing the summation fusion method demonstrates inferior performance compared to the non-query-driven model.

\noindent\textbf{Interaction between Query and Video}: 
The findings from Figure \ref{fig:acc_epochs_query_testing} and Figure \ref{fig:fusion_method} underscore the significance of employing an appropriate multi-modal feature fusion method. This importance stems from the fact that the fused feature encapsulates the implicit interaction between the video frames and the query. In instances where an improper fusion method, such as summation, is employed, the query may inadvertently confuse the network. Similar observations have been noted in prior studies, such as \cite{antol2015vqa}.

\begin{figure}[t!]
\begin{center}
\includegraphics[width=1.0\linewidth]{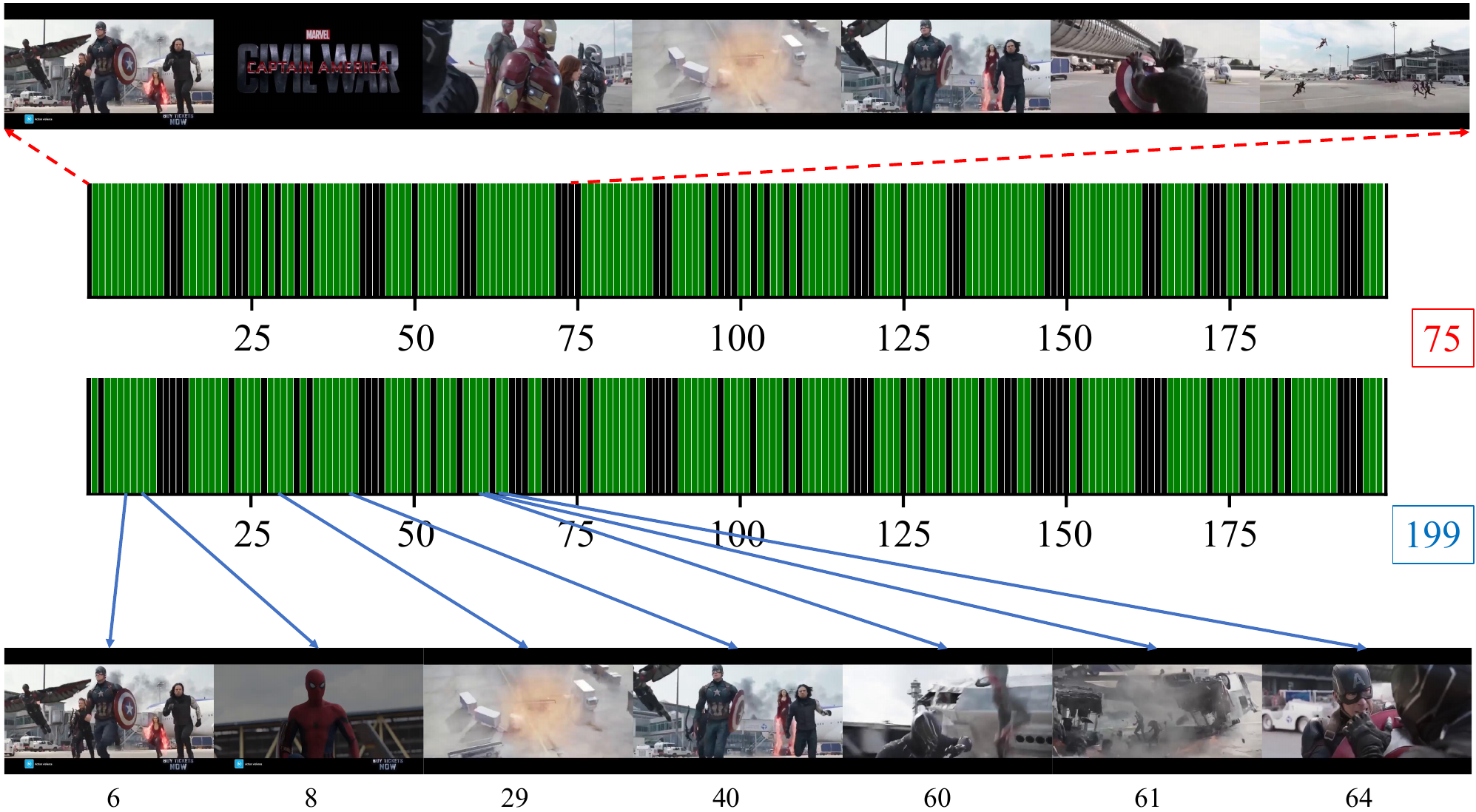}
\end{center}
   \caption{Query: ``civil war spiderman''. The correct number of predicted relevant frames out of the total number of frames is $94/199$. The total original number of frames in the input video is denoted by ``$75$'' and color-coded in red. The first row showcases selected frames from the original video, representing its content. The second row displays the entire original input video. The third row illustrates the model's predictions. Finally, the fourth row presents $k$ frames (e.g., $k=7$) representing our generated video summary. The numbers at the bottom denote the frame index in the original video, starting from $0$.}
\label{fig:testing_result_3d_movies_00_a}
\end{figure}

\subsection{Qualitative Results and Analysis}
\begin{figure}[t!]
\begin{center}
\includegraphics[width=1.0\linewidth]{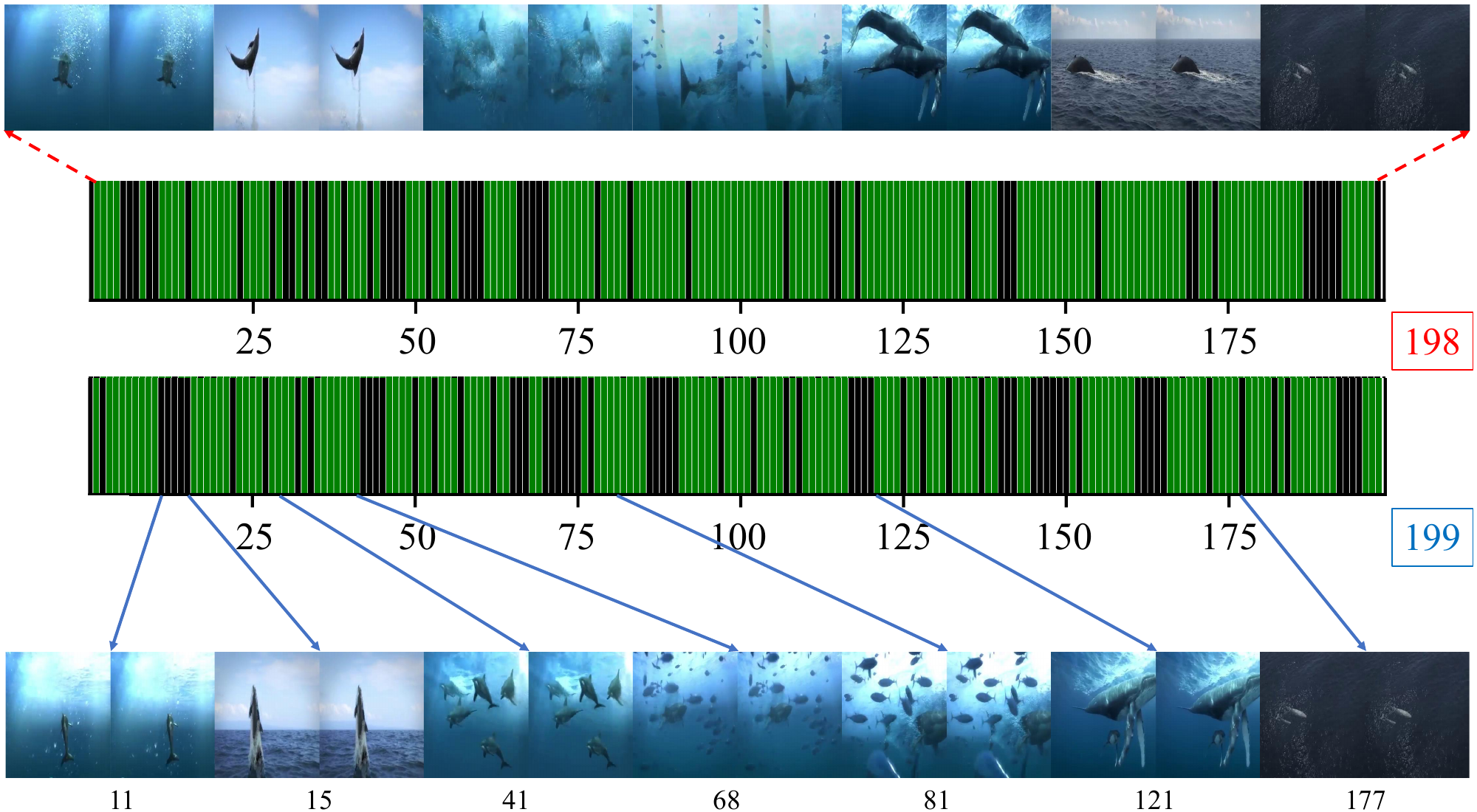}
\end{center}
   \caption{Query: ``3d movies''. The accurate number of predicted relevant frames out of the total number of frames is $120/199$. We provide the outcomes of the second generated video summary, with annotations following a format similar to that of Figure \ref{fig:testing_result_3d_movies_00_a}.}
\label{fig:testing_result_3d_movies_00_b}
\end{figure}

In this subsection, we present qualitative results showcased in Figure \ref{fig:testing_result_3d_movies_00_a} and Figure \ref{fig:testing_result_3d_movies_00_b}. Due to space limitations, we provide glimpses of both the original video and the corresponding generated video summary by displaying selected frames in chronological order.

In Figure \ref{fig:testing_result_3d_movies_00_a}, the input video is associated with the query ``civil war spiderman''. Following the explanation in the \textit{``Video Summary Output Module''} subsection, we use the color green to signify relevant frames and black for irrelevant ones. The second row in (a) illustrates video frames with ground truth labels, while the third row presents frames with predicted labels. The model correctly predicts $94$ relevant frames out of the total $199$. Given that the original video comprises $75$ frames, we display the video summary frames selected from $0$ to $74$.

Moving to Figure \ref{fig:testing_result_3d_movies_00_b}, the input corresponds to another video with the query ``3d movies'', utilizing the same color scheme to denote relevance. Similar to (a), the second row in (b) depicts video frames with ground truth labels, while the third row displays frames with predicted labels. The model accurately predicts $120$ relevant frames out of $199$. Considering the original input video with $198$ frames, we showcase the video summary frames selected from $0$ to $197$. 

The comparison between Figure \ref{fig:testing_result_3d_movies_00_a} and Figure \ref{fig:testing_result_3d_movies_00_b} underscores the capability of our proposed method to generate video summaries aligned with the content specified in the input query.

\section{Conclusion}
In conclusion, this chapter introduces a novel query-controllable video summarization approach. The conventional video summarization methods, generating fixed video summaries, often fall short in meeting the specific information needs of users, thereby limiting the efficacy of video exploration. Our proposed end-to-end deep learning-based method addresses this limitation by leveraging text-based queries to generate personalized, query-dependent video summaries. The introduced dataset with frame-based relevance score labels serves as a valuable resource for method validation and research in the domain of query-controllable video summarization. Through extensive experiments, our approach demonstrates a significant enhancement in model performance, improving accuracy by $5.83\%$. The proposed method not only contributes to the advancement of video summarization techniques but also offers a more efficient and tailored solution for navigating extensive video collections based on individual user preferences.
It's worth noting that our query-controllable video summarization technology introduced in Chapter 2, initially published at the International Conference on Multimedia Retrieval in 2020 \cite{huang2020query}, has been integrated into Google's large language model Gemini Pro-1.5 version in 2024 \cite{team2023gemini}.

\chapter{GPT-2 for Multi-modal Video Summarization}
\label{ch:filtering}

\section{Abstract}
Video summarization is crucial for condensing lengthy videos into concise clips, facilitating efficient content exploration and retrieval. Traditional methods often produce fixed summaries regardless of user preferences or content specifics, limiting their utility in video exploration. To overcome this, multi-modal video summarization has emerged, integrating text-based queries to guide the summarization process, allowing for more tailored and contextually relevant summaries. However, effectively encoding text-based queries and capturing nuanced interactions between queries and videos remain challenging. This chapter introduces a novel approach to enhance multi-modal video summarization by leveraging a specialized attention network and contextualized word representations, specifically Generative Pretrained Transformer-2 (GPT-2), to encode input text-based queries effectively. Additionally, a visual attention mechanism integrated with a pre-trained convolutional neural network (CNN) and a CNN-based interactive attention module is proposed to better model interactions between textual and visual features, improving the generation of video summaries. Extensive experiments highlight significant performance enhancements compared to state-of-the-art methods, achieving a $5.88\%$ increase in accuracy and a $4.06\%$ increase in $F_{1}$-score. These results underscore the superior performance of the proposed approach in generating effective video summaries.


\section{Introduction} 
Video summarization automatically generates concise video clips that encapsulate the essential content of longer videos, facilitating efficient content browsing and retrieval \cite{gong2014diverse,zhang2016summary,zhang2019dtr,zhou2018deep}. However, traditional video summarization methods, exemplified by \cite{ngo2003automatic,song2015tvsum,de2011vsumm,chu2015video,kang2006space,lee2012discovering,gygli2014creating}, typically produce fixed summaries regardless of user preferences or specific content requirements, limiting their utility in video exploration.

\begin{figure}[t!]
\begin{center}
\includegraphics[width=1.0\linewidth]{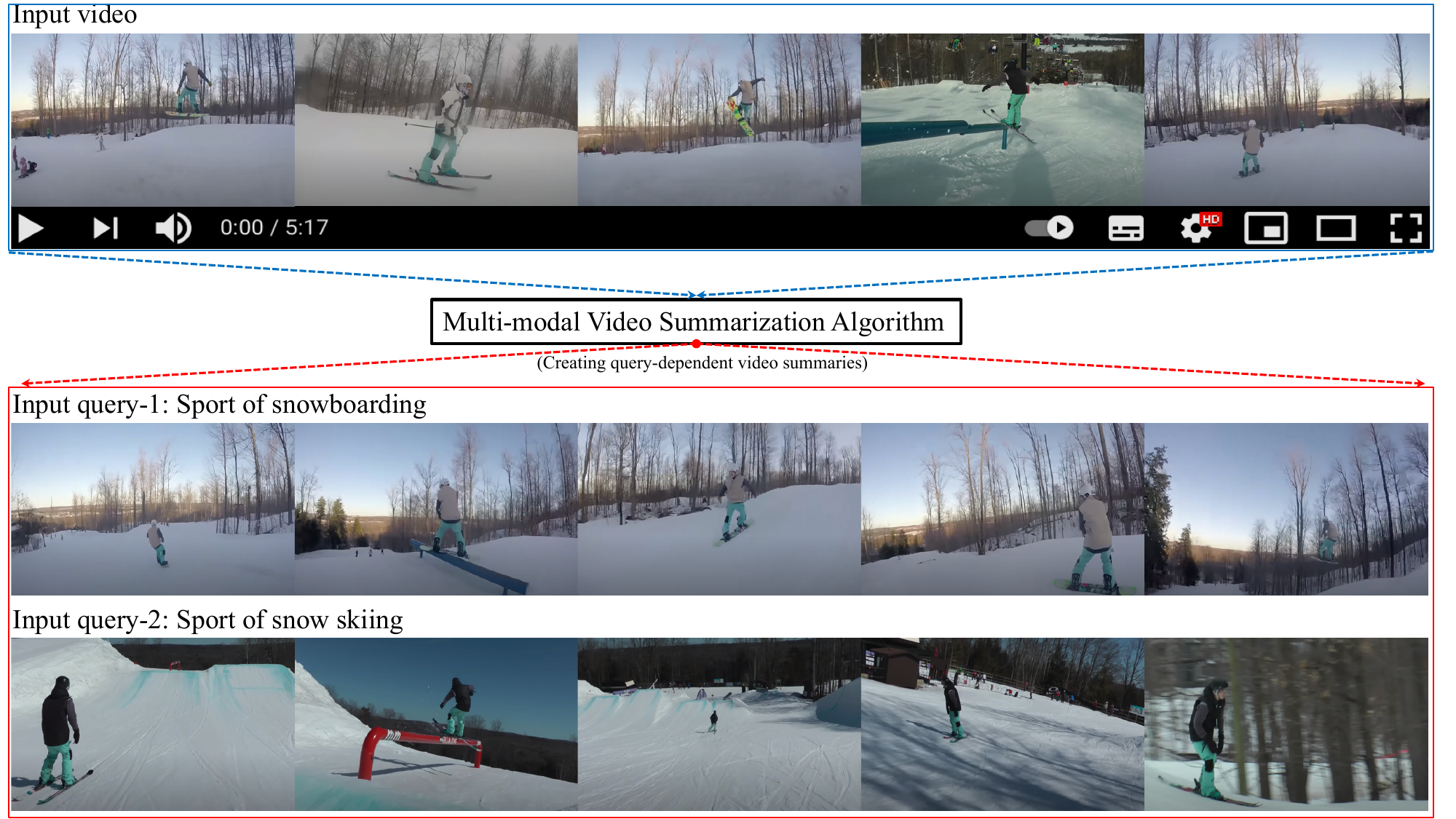}
\end{center}
   \caption{Multi-modal input integration in video summarization. The multi-modal video summarization process considers text-based queries to generate relevant summaries. For instance, with the input query ``Sport of snowboarding'', the algorithm generates a summary focused on snowboarding-related content. Similarly, when provided with the input query ``Sport of snow skiing'', the algorithm produces a summary that emphasizes skiing-related content. Each query independently guides the algorithm to tailor the summary according to the specific topic.}
\label{fig:multi-modal_vs_final_new}
\end{figure}

To address the limitations of conventional approaches, multi-modal video summarization has emerged as a promising strategy to enhance the efficiency and effectiveness of video exploration \cite{vasudevan2017query,huang2020query}. This approach leverages additional information, such as text-based queries, to guide the summarization process, as illustrated in Figure \ref{fig:multi-modal_vs_final_new}. Unlike traditional methods that rely solely on video input, multi-modal video summarization integrates user-provided text queries, allowing for more tailored and contextually relevant summaries \cite{vasudevan2017query,huang2020query}.

Effectively encoding the text-based query and capturing the nuanced interactions between the query and the video are crucial aspects of multi-modal video summarization, where the text query acts as a controller of the summary \cite{huang2020query}. In previous work \cite{huang2020query}, the Bag of Words (BoW) method was employed to encode the query input. However, while BoW has demonstrated success in various tasks such as document classification and language modeling, it has been acknowledged to have inherent limitations \cite{scott1998text,soumya2014text}. 
Firstly, the BoW method's effectiveness is heavily reliant on the careful design of its vocabulary, as the size of the vocabulary directly impacts the sparsity of the text representation. Sparse representations are inherently more challenging to model due to their increased space and time complexity. Moreover, the sparsity of information within such a large representation space often hinders the BoW method's overall effectiveness. Secondly, BoW fails to capture the sequential order of words in a text, limiting its ability to effectively convey contextual and semantic meaning.

In this chapter, a novel approach is introduced to address the aforementioned challenges and enhance the performance of multi-modal video summarization models. As highlighted in previous research \cite{ethayarajh2019contextual}, traditional methods like skip-gram with negative sampling (SGNS) \cite{mikolov2013distributed} or global vectors for word representation (GloVe) \cite{pennington2014glove}, which rely on static word embeddings, have been commonly used for text-based query encoding. However, a significant drawback of static word embeddings is their inability to differentiate between different senses of polysemous words, as they generate a single representation for each word \cite{ethayarajh2019contextual}. 
In contrast, contextualized word representations, such as those generated by Generative Pretrained Transformer-2 (GPT-2), are more effective according to recent studies \cite{ethayarajh2019contextual}. To address this, the proposed method outlined in Section 3.3 leverages a specialized attention network and utilizes contextualized word representations, specifically GPT-2, to encode input text-based queries more effectively.

In multi-modal video summarization, traditional methods like summation, element-wise multiplication, and concatenation \cite{antol2015vqa,huang2020query,huang2018robustness} have been commonly employed to encode the interactive information between textual and visual features. However, while these methods are capable of capturing some level of interaction between textual and visual information, they often suffer from information loss and may not effectively leverage the rich contextual cues present in both modalities \cite{huang2019novel,huang2017robustness,huang2017vqabq}. To address this limitation, we propose a visual attention mechanism integrated with a pre-trained convolutional neural network (CNN) to encode input video features more effectively. Additionally, to better capture the intricate interactions between textual and visual inputs, we introduce a CNN-based interactive attention module designed to enhance the model's ability to extract and fuse relevant information from both modalities.

In the realm of video summarization, the resulting summary typically comprises a selection of representative video frames or fragments \cite{apostolidis2021video}. Frame-based summaries, as emphasized in prior research \cite{calic2007efficient,wang2007video,apostolidis2021video}, offer enhanced flexibility in data organization for video exploration, as they are not constrained by synchronization or timing issues. Consequently, they furnish a more versatile framework for video exploration endeavors. Therefore, in this chapter, we validate our proposed model using a frame-based multi-modal video summarization dataset \cite{huang2020query}. Through extensive experiments conducted herein, we evaluate the effectiveness of our method. The experimental outcomes, assessed against the aforementioned benchmark, underscore the efficacy of our model. In comparison to the state-of-the-art approach, our method exhibits notable improvement, achieving a $5.88\%$ increase in accuracy and a $4.06\%$ increase in $F_{1}$-score.

\vspace{+3pt}
\noindent\textbf{Contributions.}

\begin{itemize}
    \item \textbf{Enhanced Text-Based Query Encoding in Multi-modal Video Summarization}: The proposed approach introduces a novel method for encoding text-based queries in multi-modal video summarization. Unlike the traditional BoW method, which suffers from inherent limitations such as vocabulary design challenges and the inability to capture contextual and semantic meaning, the proposed method leverages a textual attention mechanism and contextualized word representations, specifically GPT-2, to encode input text-based queries more effectively. This enhancement significantly improves the model's ability to understand and process textual inputs, resulting in more accurate and contextually relevant video summaries.

    \item \textbf{Improved Interaction Modeling between Textual and Visual Modalities}: To address the limitations of traditional methods for encoding interactive information between textual and visual features in multi-modal video summarization, the proposed approach introduces a visual attention mechanism integrated with a pre-trained 2D ResNet. Additionally, a CNN-based interactive attention module is introduced to better capture the intricate interactions between textual and visual inputs. These enhancements enable the model to extract and fuse relevant information from both modalities more effectively, leading to improved performance in generating effective video summaries.

    \item \textbf{Comprehensive Experiments and Analysis}: Extensive experiments assess the model's performance against an existing frame-based multi-modal video summarization dataset, revealing significant enhancements over state-of-the-art methods. Notably, the proposed method achieves a remarkable $5.88\%$ increase in accuracy and a $4.06\%$ boost in the $F_{1}$-score. These results underscore its superior efficacy in crafting high-quality video summaries that cater to both user preferences and specific content requirements.
\end{itemize}

\noindent The remainder of the chapter unfolds as follows: Section 3.2 provides a comprehensive review of the related work. Section 3.3 presents the proposed method in detail. Following this, Section 3.4 conducts an evaluation of the effectiveness of the proposed method, followed by a discussion of the experimental results.

\section{Related Work}
In this section, we explore related work concerning various video summarization techniques and word embedding methodologies. We begin by discussing two primary methods of video summarization: single-modality video summarization and multi-modal video summarization. Subsequently, we delve into an examination of word embedding methods.

\subsection{Video Summarization with Single Modality}
Several methods approach the problem of video summarization with a single modality, including fully supervised, weakly supervised, and unsupervised approaches.

\noindent\textbf{Fully Supervised}:
Fully supervised learning methods in video summarization, exemplified by works such as \cite{gygli2014creating,gong2014diverse,zhang2016video,zhao2017hierarchical,zhao2018hsa,zhang2019dtr,ji2019video,ji2020deep}, leverage ground truth video summaries—data expertly labeled by humans—to guide their models during training. 

In \cite{gygli2014creating}, a method tailored for user-generated videos showcasing interesting events is proposed. This approach utilizes superframe segmentation and diverse features to estimate visual interestingness scores per superframe. The final video summary is generated by optimally selecting a set of superframes. In \cite{gong2014diverse}, video summarization is approached as a supervised subset selection problem, introducing a probabilistic model named the Sequential Determinantal Point Process (SeqDPP). SeqDPP excels in selecting a diverse sequential subset, acknowledging the inherent sequential structures in video data. This addresses a limitation of the standard DPP, which treats video frames as randomly permutable elements. 

An early deep learning-based method \cite{zhang2016video} treats video summarization as a structured prediction challenge, estimating the importance of video frames by capturing their temporal dependencies. The Long Short-Term Memory (LSTM) unit \cite{hochreiter1997long} is employed to model variable-range temporal dependencies among frames while leveraging the Determinantal Point Process (DPP) \cite{kulesza2012determinantal} enhances the diversity of visual content in the generated video summary. A multilayer perceptron is utilized to estimate frame importance. Recurrent Neural Network architectures, hierarchically structured to model temporal relationships, have been applied \cite{zhao2017hierarchical,zhao2018hsa}. This temporal knowledge is then used to select video fragments for the summary.

Addressing the frame-based video summarization problem, \cite{zhang2019dtr} introduces a Dilated Temporal Relational Generative Adversarial Network (DTR-GAN). This model combines DTR and LSTM units to estimate temporal dependencies among video frames at different temporal windows. In distinguishing the machine-generated video summary from ground truth and a randomly created one, the model learns the summarization task by deceiving a trainable discriminator. In another approach, \cite{ji2019video} views video summarization as a sequence-to-sequence learning problem, introducing an LSTM-based encoder-decoder architecture with an intermediate attention layer. This model is subsequently expanded by integrating a semantic-preserving embedding network \cite{ji2020deep}.

\noindent\textbf{Weakly Supervised}:
Video summarization has also been explored within the realm of weakly supervised learning \cite{panda2017weakly,ho2018summarizing,cai2018weakly,chen2019weakly}. In a manner akin to unsupervised learning approaches, weakly supervised methods aim to alleviate the reliance on extensive human-generated ground-truth data. Instead of employing no ground-truth data, these methods harness less resource-intensive weak labels, such as video-level metadata or annotations for a small subset of frames, for model training. The underlying premise of weakly supervised learning posits that despite their imperfections compared to a complete set of human annotations, weak labels can effectively train video summarization models.

The authors of \cite{panda2017weakly} pioneered a method that bridges the gap between unsupervised and supervised learning approaches for video summarization, termed weakly supervised learning. They leverage video-level metadata, such as video titles, to categorize videos. Utilizing multiple videos within each category, they extract 3D-CNN features and train a parametric model to categorize new videos. Subsequently, the trained model is employed to select video segments that optimize the relevance between the video category and the summary. In \cite{ho2018summarizing}, the authors argue that amassing a large dataset of fully annotated first-person videos is more challenging compared to annotated third-person videos. Therefore, they propose a weakly supervised model trained on a combination of fully annotated third-person videos and a subset of first-person videos, where only a fraction includes ground-truth annotations.

The authors of \cite{cai2018weakly} propose a weakly supervised video summarization model that merges the architectures of Variational AutoEncoder (VAE) \cite{kingma2013auto} and the encoder-decoder with a soft attention mechanism. Within this architecture, the VAE component is designed to learn latent semantics from web videos. The model is trained using a weakly supervised semantic matching loss to effectively learn video summaries.
In \cite{chen2019weakly}, the authors leverage reinforcement learning principles to train a video summarization model utilizing a combination of handcrafted rewards and a limited set of human annotations. Their proposed method involves a hierarchical key-fragment selection process, which is partitioned into multiple sub-tasks. Each sub-task is learned via sparse reinforcement learning, and the final video summary is generated based on rewards reflecting its representativeness and diversity.

\noindent\textbf{Unsupervised}:
Given the absence of ground-truth data for training video summarization models, the majority of existing unsupervised methods, such as \cite{zhao2014quasi, chu2015video, panda2017collaborative, mahasseni2017unsupervised, rochan2019video, herranz2012scalable, apostolidis2019stepwise, jung2019discriminative, yuan2019cycle, apostolidis2020unsupervised}, operate on the principle that an effective video summary should provide viewers with a comprehensive understanding of the original video content.

In \cite{zhao2014quasi}, the authors introduce a method that learns a dictionary from input videos using group sparse coding. Subsequently, a video summary is crafted by combining segments that cannot be adequately reconstructed based on the learned dictionary. In \cite{chu2015video}, the authors argue that crucial visual concepts often recur across videos sharing the same theme. To leverage this insight, they propose a maximal biclique finding algorithm to identify sparsely co-occurring patterns. The video summary is then generated by identifying shots that frequently co-occur across multiple videos. In \cite{panda2017collaborative}, the authors present a video summarization approach capable of capturing both general trends observed across a set of videos and specific characteristics unique to individual videos. 
The authors of \cite{mahasseni2017unsupervised} propose a video summarization method that integrates a trainable discriminator, an LSTM-based keyframe selector, and a VAE. This model learns to generate video summaries through an adversarial learning process, aiming to minimize the discrepancy between the original video and its summary-based reconstructed version. 

Building upon the network introduced by \cite{mahasseni2017unsupervised}, \cite{apostolidis2019stepwise} proposes a stepwise label-based method to enhance the training of the adversarial part of the network, thereby improving overall model performance. In a similar vein, \cite{jung2019discriminative} introduces a method based on a VAE-GAN architecture, extending the model with a chunk and stride network.
The authors of \cite{rochan2019video} present a novel formulation for video summarization from unpaired data. The method aims to learn a mapping that aligns the distribution of the generated video summary with the distribution of a set of video summaries through an adversarial objective. To ensure visual diversity, a diversity constraint is imposed on the mapping. In \cite{yuan2019cycle}, the authors propose a method to maximize mutual information between the video and video summary using a trainable pair of discriminators and a cycle-consistent adversarial learning objective.
A variation of \cite{apostolidis2019stepwise} is suggested in \cite{apostolidis2020unsupervised}, where the VAE is substituted with a deterministic attention autoencoder. This modification aims to improve the key-fragment selection process by enabling an attention-driven reconstruction of the original video.

In general, supervised methods excel at capturing nuanced cues that might be challenging to discern using hand-crafted heuristics from ground truth summaries. Consequently, supervised approaches often outperform weakly supervised and unsupervised models, despite the latter being more label-efficient. In this chapter, we frame video summarization as a supervised learning task.

\subsection{Multi-modal Video Summarization}

Rather than solely relying on visual input, several studies have explored the potential of incorporating an additional modality, such as video captions, viewers' comments, or any other available contextual data, in the process of learning video summarization \cite{li2017extracting,vasudevan2017query,sanabria2019deep,song2016category,zhou2018video,lei2018action,otani2016video,yuan2017video,wei2018video,huang2020query}.

The authors of \cite{li2017extracting} propose a multi-modal video summarization method designed for key-frame extraction from first-person videos. Similarly, in \cite{sanabria2019deep}, a multi-modal deep learning-based approach is introduced for summarizing videos of soccer games. Building upon the context of \cite{song2016category}, \cite{zhou2018video} presents a method that learns category-driven video summaries by incentivizing the preservation of core parts found in summaries from the same category.
In a related approach, \cite{lei2018action} suggests training action classifiers with video-level annotations to facilitate action-driven video fragmentation and labeling. Subsequently, a fixed number of keyframes is extracted, and reinforcement learning is employed to select frames with the highest categorization accuracy, enabling category-driven video summarization.

In \cite{otani2016video} and \cite{yuan2017video}, video summaries are defined by maximizing their relevance with available video metadata, achieved by projecting textual and visual information into a common latent space. Meanwhile, \cite{wei2018video} implements a semantic-based video fragment selection and visual-to-text mapping, considering the relevance between original and automatically-generated video descriptions with the aid of semantic attended networks.
Although existing multi-modal video summarization methods typically employ static word embeddings \cite{mikolov2013distributed,pennington2014glove} to encode textual input, recent findings by \cite{ethayarajh2019contextual} indicate that static word embeddings may not be as effective as contextualized word representations.

This chapter introduces a novel multi-modal video summarization method leveraging contextualized word representations. This approach aims to enhance the efficiency and effectiveness of video exploration.

\subsection{Word Embeddings}

As per \cite{ethayarajh2019contextual}, word embedding methods are broadly classified into static word embeddings and contextualized word representations.

\noindent\textbf{Static Word Embedding Method}: 
GloVe \cite{pennington2014glove} and SGNS \cite{mikolov2013distributed} stand out as prominent models for generating static word embeddings. Despite their iterative nature in learning word embeddings, research has shown that both models implicitly perform factorization of a word-context matrix that contains co-occurrence statistics \cite{levy2014linguistic,levy2014neural}. A drawback of the static word embedding method is its limitation in assigning a single vector to capture all meanings of a polysemous word. This constraint arises from the creation of a unified representation for each word.

\noindent\textbf{Contextualized Word Representation Approach}: 
To overcome the limitations associated with static word embeddings, recent research has introduced methodologies for constructing context-sensitive word representations \cite{peters2018deep,devlin2018bert,radford2019language}. These approaches leverage deep neural language models that are fine-tuned to establish deep learning-based models capable of handling diverse downstream natural language processing tasks. Termed as contextualized word representations, these models derive internal word representations based on the entirety of the input sentence.
Studies, such as the work by \cite{liu2019linguistic}, indicate that contextualized word representations effectively capture task-agnostic and highly transferable language properties. The methodology proposed by \cite{peters2018deep} involves generating contextualized representations for each token by concatenating the internal states of a two-layer bidirectional LSTM, trained on a bidirectional language modeling task. Meanwhile, the approaches outlined by \cite{radford2019language} and \cite{devlin2018bert} incorporate unidirectional and bidirectional transformer-based language models, respectively. In a comparative analysis conducted by \cite{ethayarajh2019contextual}, it is demonstrated that static word embeddings are less effective when contrasted with contextualized word representations.

This chapter introduces a novel approach that leverages a specialized attention network alongside the contextualized word representation method for query encoding.

\begin{figure*}[t!]
  \includegraphics[width=\textwidth]{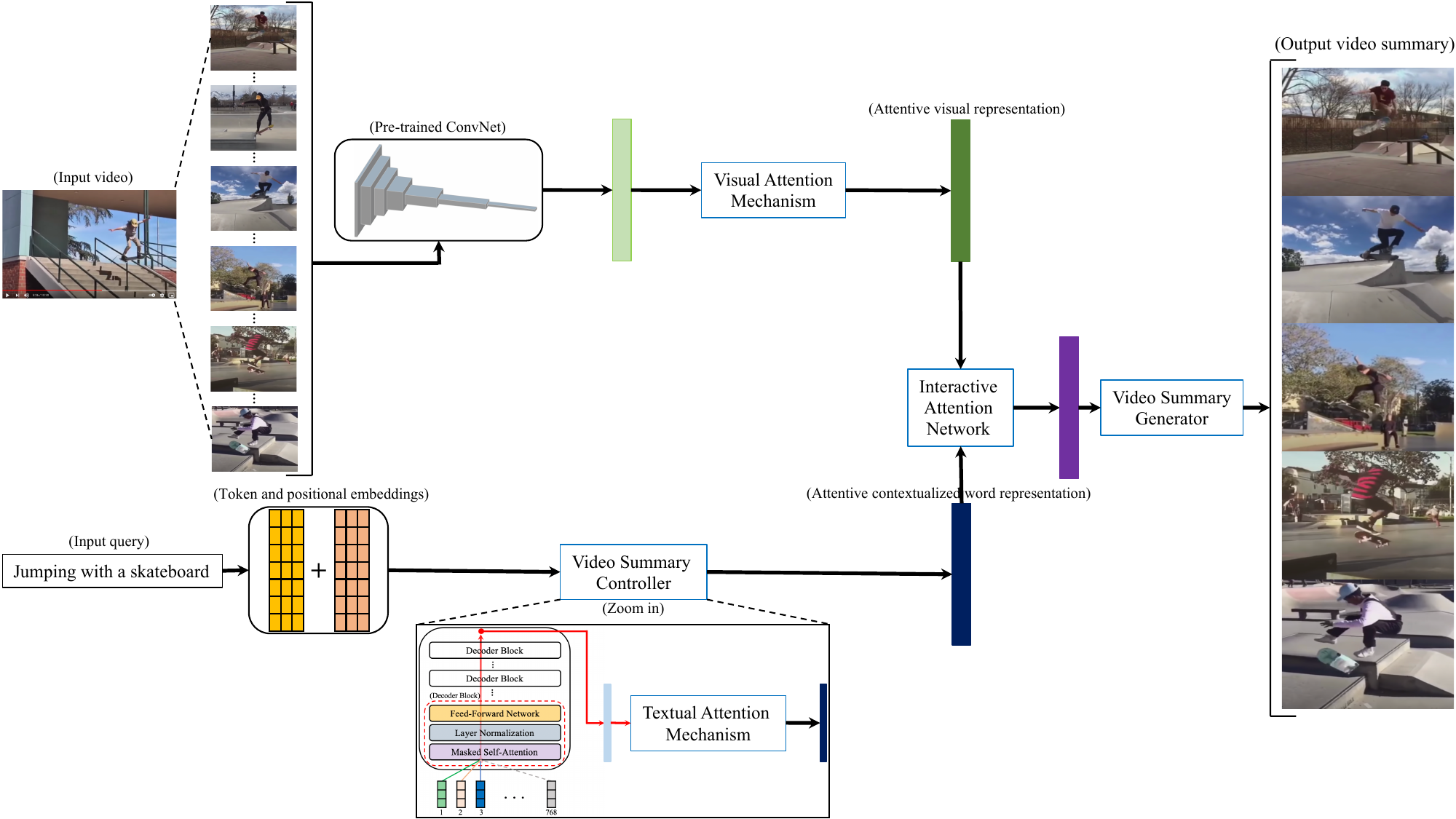}
  \caption{
  Flowchart illustrating the proposed multi-modal video summarization method. A pre-trained CNN extracts features from the visual input, enabling the ``Visual Attention Mechanism'' to produce the attentive visual representation (depicted in dark green). Concurrently, the ``Token and positional embedding'' processes the input text-based query, generating input for the ``Video Summary Controller''. The ``Textual Attention Mechanism'' generates the attentive contextualized word representation (depicted in dark blue). Subsequently, the ``Interactive Attention Network'' integrates the attentive visual and contextualized word representations to produce an informative feature vector (depicted in purple). This informative feature vector serves as input for the ``Video Summary Generator'', which generates the query-dependent video summary. Additional details are provided in the \textit{Methodology} section.}
  \label{fig:flowchart_final}
\end{figure*}

\section{Methodology}
In this section, we present a novel and detailed description of our multi-modal video summarization method. The proposed method integrates several key components: a contextualized video summary controller, a textual attention mechanism, a visual attention mechanism, an interactive attention network, and a video summary generator. The comprehensive flowchart of our proposed method is illustrated in Figure \ref{fig:flowchart_final}.

To extract features from the visual input, a pre-trained CNN, such as ResNet \cite{he2016deep}, is employed. The ``Visual Attention Mechanism'' is utilized to generate an attentive visual representation, highlighted in dark green. Simultaneously, an input text-based query (e.g., ``Jumping with a skateboard'') is processed by the ``Token and positional embedding'' to create the input for the ``Video Summary Controller''. This controller comprises a stack of decoder blocks and a ``Textual Attention Mechanism''. Each decoder block encompasses masked self-attention, layer normalization, and a feed-forward network, represented by the red dashed line box. The $768$ color-coded brick-stacked vectors serve as the input for the summary controller. Notably, the masked self-attention operates as a function of $Q$, $K$, and $V$ (i.e., $\textup{MaskAtten}(Q,K,V)$ in Equation (\ref{eq:attention})).

The ``Textual Attention Mechanism'' utilizes the output from the last decoder block to generate the attentive contextualized word representation, depicted in dark blue—the output of the summary controller. The proposed ``Interactive Attention Network'' then takes both the attentive visual representation and the attentive contextualized word representation as inputs, producing an informative feature vector shown in purple. This informative feature vector becomes the input for the ``Video Summary Generator'', whose role is to produce the query-dependent video summary.

\subsection{Contextualized Video Summary Controller}
The transformer architecture, as introduced by \cite{vaswani2017attention}, has emerged as a cornerstone in state-of-the-art methods for machine translation and language modeling. This architecture consists of a transformer-encoder and a transformer-decoder, both comprised of multiple basic transformer blocks. Drawing inspiration from the transformer-decoder structure, particularly exemplified by models like GPT-2, which leverages masked self-attention and parallelization, we incorporate these features in developing our contextualized video summary controller for text-based query embedding.

The detailed description of the summary controller is as follows: for an input token $k_n$, its embedding $x_n$ is defined as:

\begin{equation} 
    x_n = \mathbf{W}_e*k_n+P_{k_n}, n \in \{0,...,N-1\},
	\label{eq:text_embed1}
\end{equation} 
where $\mathbf{W}_e \in \mathbb{R}^{E_s \times V_s}$ represents the token embedding matrix, where $E_s$ is the word embedding size and $V_s$ is the vocabulary size. $P_{k_n}$ stands for the positional encoding of $k_n$, as introduced by \cite{vaswani2017attention}, and $N$ is the total number of input tokens. The variable $n$ takes non-negative integer values from $0$ to $N-1$. The subscripts $e$ and $s$ specifically denote ``embedding'' and ``size'', respectively. The symbol ``$*$'' indicates matrix multiplication.
 


The representation of the current word $Q$ is generated by a single linear layer, defined as:

\begin{equation}
    Q = \mathbf{W}_q*x_n+b_q,
	\label{eq:text_embed2}
\end{equation}
where the parameters of the linear layer are denoted as $\mathbf{W}_q \in \mathbb{R}^{H_s \times E_s}$ and $b_q$, where $H_s$ represents the output size of the linear layer. The subscript $q$ indicates ``query'' \cite{vaswani2017attention}.


The key vector $K$ \cite{vaswani2017attention} is computed using another linear layer, defined as:

\begin{equation}
    K = \mathbf{W}_k*x_n+b_k,
	\label{eq:text_embed3}
\end{equation}
where the parameters of the linear layer are represented as $\mathbf{W}_k \in \mathbb{R}^{H_s \times E_s}$ and $b_k$, where $H_s$ indicates the output size of the linear layer. The subscript $k$ signifies ``key'' \cite{vaswani2017attention}.


The value vector $V$ \cite{vaswani2017attention} is produced by another linear layer, defined as:

\begin{equation}
    V = \mathbf{W}_v*x_n+b_v,
	\label{eq:text_embed4}
\end{equation}
where the parameters of the linear layer are denoted as $\mathbf{W}_v \in \mathbb{R}^{H_s \times E_s}$ and $b_v$, where $H_s$ is the output size of the linear layer. The subscript $v$ indicates ``value'' \cite{vaswani2017attention}.


Once $Q$, $K$, and $V$ are computed, the masked self-attention $Z$ is produced as per Equation (\ref{eq:attention}).

\begin{equation}
    Z = \textup{MaskAtten}(Q,K,V) = \textup{softmax}(m(\frac{QK^\intercal}{\sqrt{d_k}}))V,
	\label{eq:attention}
\end{equation}
where $m(\cdot)$ represents a masked self-attention function, and $d_k$ indicates a scaling factor \cite{vaswani2017attention}. $Z$ is defined as a function of $Q$, $K$, and $V$.

Then, the layer normalization is calculated using Equation (\ref{eq:layernorm}).

\begin{equation}
    Z_{\textup{Norm}} = \textup{LayerNorm}(Z),
	\label{eq:layernorm}
\end{equation}
where the function $\textup{LayerNorm}(\cdot)$ denotes layer normalization.


The introduced contextualized representation $F$ of the text-based query is obtained through Equations (1-6):

\begin{equation}
    F = \textup{FFN}(Z_{\textup{Norm}}) = \sigma(\mathbf{W}_1Z_{\textup{Norm}}+b_1)\mathbf{W}_2+b_2,
	\label{eq:ffn}
\end{equation}
where the function $\textup{FFN}(\cdot)$ denotes a position-wise feed-forward network (FFN), $\sigma$ represents an activation function, and $\mathbf{W}_{1}$, $\mathbf{W}_{2}$, $b_{1}$, and $b_{2}$ are learnable parameters of the FFN.

\subsection{Multi-modal Attentions}

To enhance both textual and visual representations, we propose a textual attention mechanism to strengthen the contextualized representation $F$ and a visual attention mechanism to enhance the visual features extracted by CNN.

\noindent\textbf{Textual Attention Mechanism}:
The proposed textual attention mechanism is characterized by a function denoted as $TextAtten(\cdot)$, as outlined in Equation (\ref{eq:text-attention}). This function takes the output of the FFN, i.e., $F$ from Equation (\ref{eq:ffn}), as input, computing attention and textual representation through an element-wise operation, specifically Hadamard textual attention.

\begin{equation}
    Z_{\textup{ta}} = \textup{TextAtten}(F),
	\label{eq:text-attention}
\end{equation}
where the subscript $\textup{ta}$ indicates ``textual attention''.

\noindent\textbf{Visual Attention Mechanism}:
The proposed visual attention mechanism is characterized by a function denoted as $VisualAtten(\cdot)$, as described in Equation (\ref{eq:visual-attention}). This function takes the visual representation $\phi(I)$, extracted by CNN, as input, computing attention and visual representation through an element-wise operation, i.e., Hadamard visual attention.

\begin{equation}
    Z_{\textup{va}} = \textup{VisualAtten}(\phi(I)),
	\label{eq:visual-attention}
\end{equation}
where the subscript $\textup{va}$ represents ``visual attention''.



\subsection{Interactive Attention Network}
In multi-modal video summarization, capturing interactive information between the query and the video is crucial. To achieve this, an interactive attention network is proposed. In Equation (\ref{eq:inter-attention}), $\textup{InterAtten}(\cdot)$ represents this interactive attention network, which utilizes one-by-one convolution, i.e., convolutional attention.

\begin{equation}
    Z_{\textup{ia}} = \textup{InterAtten}(Z_{\textup{ta}} \odot Z_{\textup{va}}),
	\label{eq:inter-attention}
\end{equation}
where $Z_{\textup{ta}}$ indicates textual attention, $Z_{\textup{va}}$ represents visual attention, and $\odot$ denotes the Hadamard product.


\subsection{Loss Function}

In \cite{huang2020query}, the authors treat multi-modal video summarization as a classification problem and employ the widely used cross-entropy loss as their loss function. Given that we similarly model the multi-modal video summarization problem as a classification task and conduct our experiments on the dataset from \cite{huang2020query}, we also utilize the cross-entropy loss function, as defined in Equation (\ref{eq:loss_chapter_3}), to formulate our proposed model.

\begin{equation}
    \textup{Loss}(x, \textup{class}) = -x \left[ \textup{class} \right] + \textup{ln}(\sum_{j}\textup{exp}(x \left[ j \right] )),
    \label{eq:loss_chapter_3}
\end{equation}
where $x$ denotes the prediction, $\textup{class}$ represents the ground truth class, and $j$ indicates the index for iteration \cite{NEURIPS2019_9015,huang2020query}.


\subsection{Video Summary Generator}
The objective of the video summary generator is to construct video summaries using the potent vector representation of the text-based query and video, derived from the outcome of Equation (\ref{eq:inter-attention}). The proposed summary generator utilizes a fully connected linear layer to produce a frame-based score vector for a given query-video pair. Subsequently, it generates the final video summary based on this vector. The complete video summary generation process is illustrated in Figure \ref{fig:flowchart_final}.

\section{Experiments and Analysis}
In this section, we provide a comprehensive description of the experimental setup, detailing the validation of our proposed multi-modal video summarization model using an established dataset \cite{huang2020query}. Subsequently, we delve into analyzing the effectiveness of the various attention-based modules and contextualized word representations proposed in this study. Finally, we present randomly selected qualitative results to further illustrate the performance of our model.

\subsection{Dataset Preparation and Evaluation Metrics}

\noindent\textbf{Dataset}:
In our experiments, we utilize the multi-modal video summarization dataset introduced by \cite{huang2020query} to validate our model. This dataset comprises $190$ videos, each spanning two to three minutes, all retrieved based on given text-based queries. The dataset is partitioned into training/validation/testing sets at a ratio of $60\%$/$20\%$/$20\%$, resulting in $114/38/38$ videos, respectively.

Human expert annotations are essential for the automatic evaluation of multi-modal video summarization methods. To this end, all $190$ videos are sampled at one frame per second (fps), and annotations are obtained through Amazon Mechanical Turk (AMT). Each frame is annotated with a relevance level to the given query. The distribution of relevance level annotations is as follows: ``Very Good'' ($18.65\%$), ``Good'' ($55.33\%$), ``Not Good'' ($13.03\%$), and ``Bad'' ($12.99\%$). These relevance level annotations are mapped to numerical values, with ``Very Good'' assigned $3$, ``Good'' assigned $2$, ``Not Good'' assigned $1$, and ``Bad'' assigned $0$.
For each query-video pair, a single ground truth relevance level label is derived by aggregating the corresponding annotations from AMT workers, as proposed by \cite{huang2020query}. Notably, the maximum allowed query length in this dataset is $8$ words.

To evaluate model performance, we adopt the majority vote rule, wherein a predicted relevance level is deemed correct if it matches the majority of human annotators' provided scores. Accuracy is thus employed as the primary evaluation metric \cite{huang2020query}.

\noindent\textbf{Evaluation Metrics}:
In \cite{huang2020query}, the authors employ accuracy, derived from the predicted and ground truth frame-based scores, as the metric for evaluating their model's performance. Given that the experiments in this chapter are conducted on the dataset introduced by \cite{huang2020query}, the same accuracy metric is adopted to assess the model's efficacy. Additionally, drawing inspiration from \cite{hripcsak2005agreement,gygli2014creating,song2015tvsum}, we utilize the $F_{\beta}$-score with the hyper-parameter $\beta=1$, as outlined in Equation (\ref{eq:f1-score}), to gauge the proposed model's performance. This score measures the agreement between the predicted scores and the gold standard scores provided by human annotators.

\begin{equation}
    F_{\beta}=\frac{1}{N}\sum_{i=1}^{N}\frac{(1+\beta ^{2})\times p_{i}\times r_{i}}{(\beta ^{2}\times p_{i})+r_{i}},
    \label{eq:f1-score}
\end{equation}
where $p_{i}$ indicates $i$-th precision, $r_{i}$ represents $i$-th recall, $N$ represents number of $(p_{i}, r_{i})$ pairs, ``$\times$'' indicates scalar product, and $\beta$ is used to balance the relative importance between recall and precision. 


\begin{table}[t!]
    \caption{Evaluation of method performance across varied dimensions of contextualized word representations. The optimal word embedding dimension (highlighted in bold) is determined empirically for maximizing each model's accuracy \cite{huang2020query}. The default output word embedding dimensions for each model: GPT-2$=768$, GPT-2-M$=1024$, GPT-2-L$=1280$, GPT-2-XL$=1600$.}
\begin{center}
\scalebox{0.89}{
    \begin{tabular}{c|c|c|c|c}
    \toprule
    \textbf{Word Embedding Dimension} & \textbf{GPT-2} & \textbf{GPT-2-M} & \textbf{GPT-2-L} & \textbf{GPT-2-XL} \\ 
    \midrule
    10 dimensions   & 0.7551  & 0.7399  & 0.7506  & 0.7547 \\ 
    \midrule
    50 dimensions   & 0.7541  & 0.7194  & 0.7509  & 0.7419 \\ 
    \midrule
    100 dimensions  & 0.7510  & 0.7527  & 0.7428  & 0.7342 \\ 
    \midrule
    \cellcolor{mygray} \textbf{150 dimensions}  & 0.7447  & 0.7484  & 0.7392  & \cellcolor{mygray} \textbf{0.7662} \\ 
    \midrule
    200 dimensions  & 0.7383  & 0.7447  & 0.7035  & 0.7468 \\ 
    \midrule
    250 dimensions  & 0.7355  & 0.7465  & 0.7158  & 0.7447 \\ 
    \midrule
    \cellcolor{mygray} \textbf{300 dimensions}  & \cellcolor{mygray} \textbf{0.7566}  & 0.7452  & 0.7552  & 0.7552 \\ 
    \midrule
    \cellcolor{mygray} \textbf{350 dimensions}  & 0.7543  & \cellcolor{mygray} \textbf{0.7558}  & 0.7473  & 0.7244 \\ 
    \midrule
    \cellcolor{mygray} \textbf{400 dimensions}  & 0.7318  & 0.7326  & \cellcolor{mygray} \textbf{0.7583}  & 0.6922 \\ 
    \midrule
    450 dimensions  & 0.7381  & 0.7470  & 0.7407  & 0.7435 \\ 
    \midrule
    500 dimensions  & 0.7436  & 0.7501  & 0.7451  & 0.7451 \\ 
    \midrule
    Default output dimensions  & 0.7375  & 0.7411  & 0.7460  & 0.7469 \\ 
    \bottomrule

    \end{tabular}}

\label{table:table1_chapter_3}
\end{center}
\end{table}


\begin{table}[t!]
\caption{Ablation study on different attention mechanisms using $F_{1}$-score \cite{hripcsak2005agreement,gygli2014creating,song2015tvsum} to assess model performance. In the notation, ``w/o'' signifies models without a specific type of attention, and ``w/'' denotes models with a specific type of attention. Based on the $F_{1}$-scores, it can be concluded that the proposed attentions are effective.}
\centering
\rotatebox{270}{
\scalebox{0.75}{
\begin{tabular}{c|c|c|c|c|c}
\toprule
\multicolumn{2}{c|}{\textbf{Attention Type}}   & \textbf{GPT-2} (300-dim) & \textbf{GPT-2-M} (350-dim) & \textbf{GPT-2-L} (400-dim) & \textbf{GPT-2-XL} (150-dim) \\ 
\midrule
\multirow{2}{*}{Visual Attention}           & w/o  & 0.4905  & 0.5199  & 0.5225  & 0.5140  \\ \cline{2-6}

& w/  & \cellcolor{mygray} \textbf{0.5200}   & \cellcolor{mygray} \textbf{0.5277}  & \cellcolor{mygray} \textbf{0.5247}  & \cellcolor{mygray} \textbf{0.5340}   \\ 
\midrule

\multirow{2}{*}{Textual Attention}          & w/o  & 0.4905  & 0.5199  & 0.5225  & 0.5140   \\ \cline{2-6} 
& w/  & \cellcolor{mygray} \textbf{0.5183}   & \cellcolor{mygray} \textbf{0.5334}   & \cellcolor{mygray} \textbf{0.5266}  & \cellcolor{mygray} \textbf{0.5260}   \\ 
\midrule

\multirow{2}{*}{Visual-Textual Attention}     & w/o  & 0.4905  & 0.5199  & 0.5225  & 0.5140    \\ \cline{2-6} 
& w/  & \cellcolor{mygray} \textbf{0.5247} & \cellcolor{mygray} \textbf{0.5357}  & \cellcolor{mygray} \textbf{0.5319}   & \cellcolor{mygray} \textbf{0.5363}    \\ 
\midrule

\multirow{2}{*}{Interactive Attention}        & w/o  & 0.4905  & 0.5199  & 0.5225  & 0.5140    \\ \cline{2-6} 
& w/  & \cellcolor{mygray} \textbf{0.5040} & \cellcolor{mygray} \textbf{0.5327}  & \cellcolor{mygray} \textbf{0.5275}  & \cellcolor{mygray} \textbf{0.5389}   \\ 
\midrule

\multirow{2}{*}{Interactive-Visual-Textual Attention}       & w/o  & 0.4905  & 0.5199  & 0.5225  & 0.5140    \\ \cline{2-6} 
& w/  & \cellcolor{mygray} \textbf{0.5410}  & \cellcolor{mygray} \textbf{0.5484}   & \cellcolor{mygray} \textbf{0.5473}   & \cellcolor{mygray} \textbf{0.5420}   \\ 
\bottomrule

\end{tabular}}
}
\label{table:table2_chapter_3}
\end{table}

\begin{table}[t!]
    \caption{Comparison with state-of-the-art \textbf{QueryVS} \cite{huang2020query} using accuracy \cite{huang2020query} and $F_{1}$-score \cite{hripcsak2005agreement}. The proposed approach surpasses the model in \cite{huang2020query} by $5.88\%$ in accuracy and $4.06\%$ in F1-score.}
\begin{center}
\rotatebox{270}{
\scalebox{0.75}{
    \begin{tabular}{c|c|c|c|c|c}
    \toprule
    \textbf{Evaluation Metric} & \textbf{GPT-2} (300-dim)& \textbf{GPT-2-M} (350-dim)& \textbf{GPT-2-L} (400-dim)& \textbf{GPT-2-XL} (150-dim) & \textbf{QueryVS }\cite{huang2020query}\\ 
    \midrule

    Accuracy \cite{huang2020query}  & 0.7424  & \cellcolor{mygray} \textbf{0.7625}  & 0.7493  & 0.7510 & 0.7037\\ 
    \midrule

    $F_{1}$-score \cite{hripcsak2005agreement}  & 0.5410  & \cellcolor{mygray} \textbf{0.5484}  & 0.5473  & 0.5420 & 0.5078\\ 
    \bottomrule

    \end{tabular}}
}
    \label{table:table3_chapter_3}
\end{center}
\end{table}

\vspace{+1cm}
\subsection{Experimental Setup}
The methodology of pre-training a CNN on ImageNet \cite{deng2009imagenet} and leveraging it for vision-related tasks, particularly as a visual feature extractor, has gained widespread acceptance due to its efficacy. 

In this chapter, we employ a 2D ResNet \cite{he2016deep} pre-trained on ImageNet to extract $199$ frame-based features for each video. These features are situated in the visual layer immediately preceding the classification layer. Notably, the video lengths in the dataset proposed by \cite{huang2020query} vary, resulting in different frame counts for each video due to their extraction at $1$ fps. The maximum number of frames for a video is capped at $199$ in their dataset. To maintain consistency, the authors utilize frame-repeating \cite{huang2020query} during video preprocessing, ensuring all videos have a uniform length of $199$ frames. For CNN input, the frame size is set to $224 \times 224$ with red, green, and blue channels. Each image channel is normalized using a mean of $=(0.4280, 0.4106, 0.3589)$ and a standard deviation of $=(0.2737, 0.2631, 0.2601)$.

Given that our implementation is based on the transformer-decoder architecture \cite{vaswani2017attention}, specifically GPT-2 \cite{radford2019language}, for the development of the contextualized video summary controller for text-based query embedding, initializing the summary controller with pre-trained GPT-2 weights is beneficial. According to \cite{radford2019language}, GPT-2 has been pre-trained on a large corpus, featuring a vocabulary size of $=50,257$.

For the implementation in this chapter, PyTorch is the chosen framework. The training process involves $10$ epochs, a learning rate of $1e-4$, and the Adam optimizer \cite{kingma2014adam}. The optimizer parameters include $\beta_{1}=0.9$ and $\beta_{2}=0.999$ as coefficients for computing moving averages of gradient and its square. Additionally, $\epsilon=1e-8$ is introduced to the denominator to enhance numerical stability.

\subsection{Effectiveness Analysis of Various Attentions}
As the size or dimension of word embeddings plays a crucial role in both training efficiency and model performance, multiple experiments have been undertaken using different word embedding dimensions to scrutinize the proposed method. Please refer to Table \ref{table:table1_chapter_3}, Table \ref{table:table2_chapter_3}, and Table \ref{table:table3_chapter_3} for detailed results. Subsequently, the model demonstrating the optimal word embedding dimension and performance is chosen for an in-depth ablation study, focusing on the various attentions proposed in this work.

\noindent\textbf{Textual Attention}:
Table \ref{table:table2_chapter_3} displays the results of the ablation study focusing on the textual attention mechanism. The findings underscore the effectiveness of textual attention in enhancing model performance. Specifically, it yields improvements of $2.78$\% for GPT-2, $1.35$\% for GPT-2-M, $0.41$\% for GPT-2-L, and $1.2$\% for GPT-2-XL, respectively.

\noindent\textbf{Visual Attention}:
The results of the ablation study presented in Table \ref{table:table2_chapter_3} demonstrate the effectiveness of the visual attention mechanism. It contributes to improvements in model performance, yielding enhancements of $2.95$\% for GPT-2, $0.78$\% for GPT-2-M, $0.22$\% for GPT-2-L, and $2$\% for GPT-2-XL, respectively.

\noindent\textbf{Interactive Attention}:
As indicated by the results in Table \ref{table:table2_chapter_3}, the proposed interactive attention network proves to be effective, contributing to improvements in model performances. Specifically, there are enhancements of $1.35$\% for GPT-2, $1.28$\% for GPT-2-M, $0.5$\% for GPT-2-L, and $2.49$\% for GPT-2-XL.

The aforementioned attention mechanisms play a crucial role in effectively leveraging the importance of features within a high-dimensional space, potentially aiding the model in converging to a superior local optimum. Through the conducted ablation study, it is concluded that all proposed attentions contribute significantly to the model's effectiveness.

\subsection{Effectiveness Analysis of Attentive Contextualized Word Representations}
\begin{figure*}[t!]
  \includegraphics[width=\textwidth]{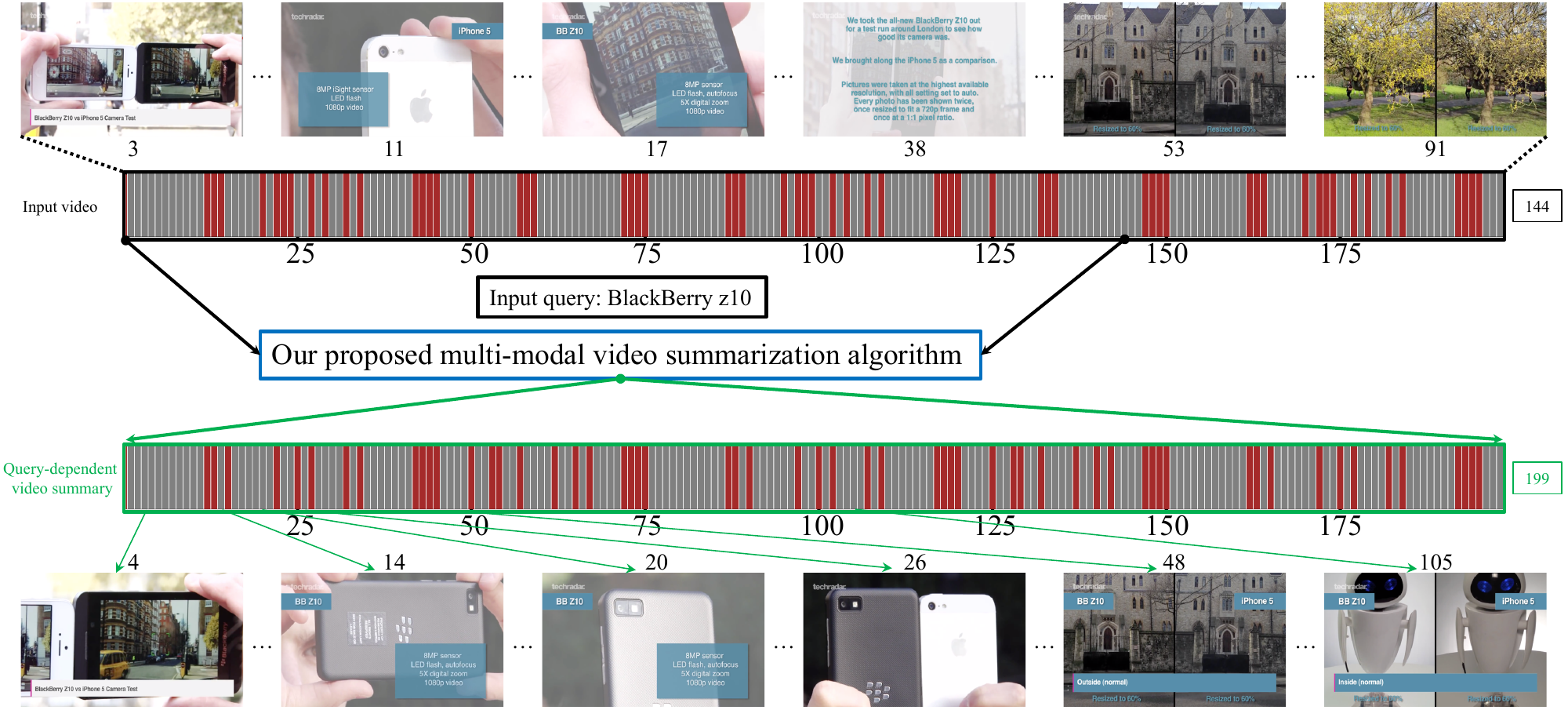}
  \caption{Input query: ``BlackBerry z10''. The accuracy of score predictions is calculated as the ratio of correct predictions to the total number of frames: $144/199$. The query-dependent video summary is highlighted in green. In the first two rows of this example, we present visualizations of the input video and the corresponding ground truth frame-based score annotations, respectively. The last two rows of the example exhibit visualizations of the predicted frame-based scores and a partial display of the query-dependent video summary. In each frame-based score pattern, gray indicates ``selected frames'', while red indicates ``not selected frames''. The number $144$ in this example denotes the video length before video preprocessing, while $199$ denotes the video length after preprocessing. Further details are provided in Subsection 4.2.}
  \label{fig:blackberry_z10}
\end{figure*}

\begin{figure*}[t!]
  \includegraphics[width=\textwidth]{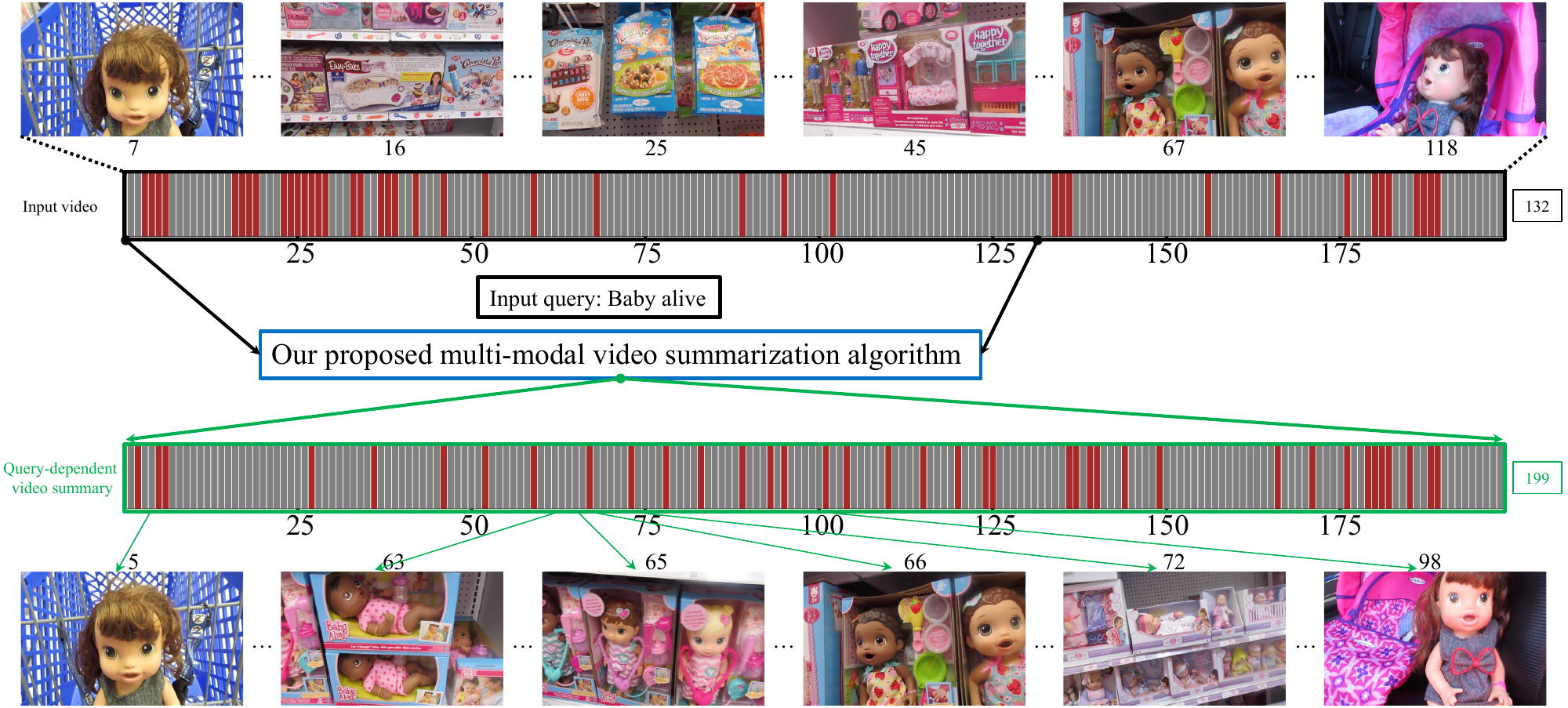}
  \caption{Input query: ``Baby alive''. ``Number of correct score predictions $/$ Total number of frames'' = $132/199$.
  The number $132$ represents the video length before video preprocessing, while $199$ represents the video length after preprocessing. The same explanatory framework as in Figure \ref{fig:blackberry_z10} applies to this example as well.}
  \label{fig:baby_alive}
\end{figure*}

The authors of \cite{huang2020query} propose a state-of-the-art model based on their newly introduced multi-modal video summarization benchmark \cite{huang2020query}. To demonstrate the effectiveness of our proposed attentive contextualized approach, the model's performance is compared against \cite{huang2020query}. As shown in Table \ref{table:table3_chapter_3}, the results indicate that the proposed method outperforms the state-of-the-art. This improvement can be attributed to the more effective embedding of multi-modal inputs, namely the text-based query and the video, compared to \cite{huang2020query}. It also highlights the superiority of contextualized word representations over the BoW approach utilized in \cite{huang2020query}. Additionally, qualitative results are illustrated in Figure \ref{fig:blackberry_z10} and Figure \ref{fig:baby_alive}.

\section{Conclusion}
In summary, this chapter has introduced a novel approach to enhance multi-modal video summarization, effectively addressing significant challenges in condensing lengthy videos into concise and contextually relevant clips. Through the utilization of a textual attention mechanism and contextualized word representations, particularly GPT-2, our proposed method markedly enhances the encoding of text-based queries, resulting in more accurate and nuanced video summaries. Moreover, the incorporation of a visual attention mechanism and a CNN-based interactive attention module enhances the modeling of interactions between textual and visual modalities, leading to improved performance in generating effective video summaries. Extensive experiments using a frame-based multi-modal video summarization dataset have highlighted the superior performance of our approach compared to state-of-the-art methods, demonstrating notable increases in accuracy and $F_{1}$-score. Overall, this research significantly contributes to advancing the field of multi-modal video summarization, providing a robust framework for generating high-quality video summaries tailored to user preferences and specific content requirements.

\chapter{Conditional Modeling Based Video Summarization}
\label{ch:focus}

\section{Abstract}
Video summarization is essential for making video content more accessible and understandable by condensing lengthy videos into concise clips. However, current automatic video summarization methods often neglect non-visual factors like interestingness and storyline consistency identified by human experts. This chapter introduces a novel approach that incorporates both visual and non-visual factors into the summarization process, drawing insights from human assessment of video summary quality. By leveraging a conditional modeling perspective and data intervention, the proposed method enhances the model's ability to learn relationships among diverse factors, resulting in higher-quality video summaries aligned with human perception standards. Innovative design choices, such as helper distributions and a conditional attention module, further improve model training and address potential performance degradation in multi-modal input scenarios. Comprehensive experiments demonstrate the superiority of the proposed approach in terms of accuracy and $F_{1}$-score, contributing to the advancement of video summarization by providing a robust framework that captures various aspects of video content.

\section{Introduction}
\label{sec:introduction}
Video summarization serves as a pivotal tool for enhancing the accessibility, searchability, and comprehension of video content, enabling users to navigate large video collections efficiently and extract pertinent information. A successful video summary is a concise clip that effectively communicates the central message and narrative of the original video. To achieve this objective, there has been a recent surge in interest and development of automatic video summarization algorithms \cite{plummer2017enhancing,chu2015video,panda2017collaborative,potapov2014category,rochan2019video,li2018local,zhou2018deep,sharghi2018improving,zhang2018retrospective,wu2022intentvizor}.

\begin{figure*}[t!]
\begin{center}
\includegraphics[width=1.0\textwidth]{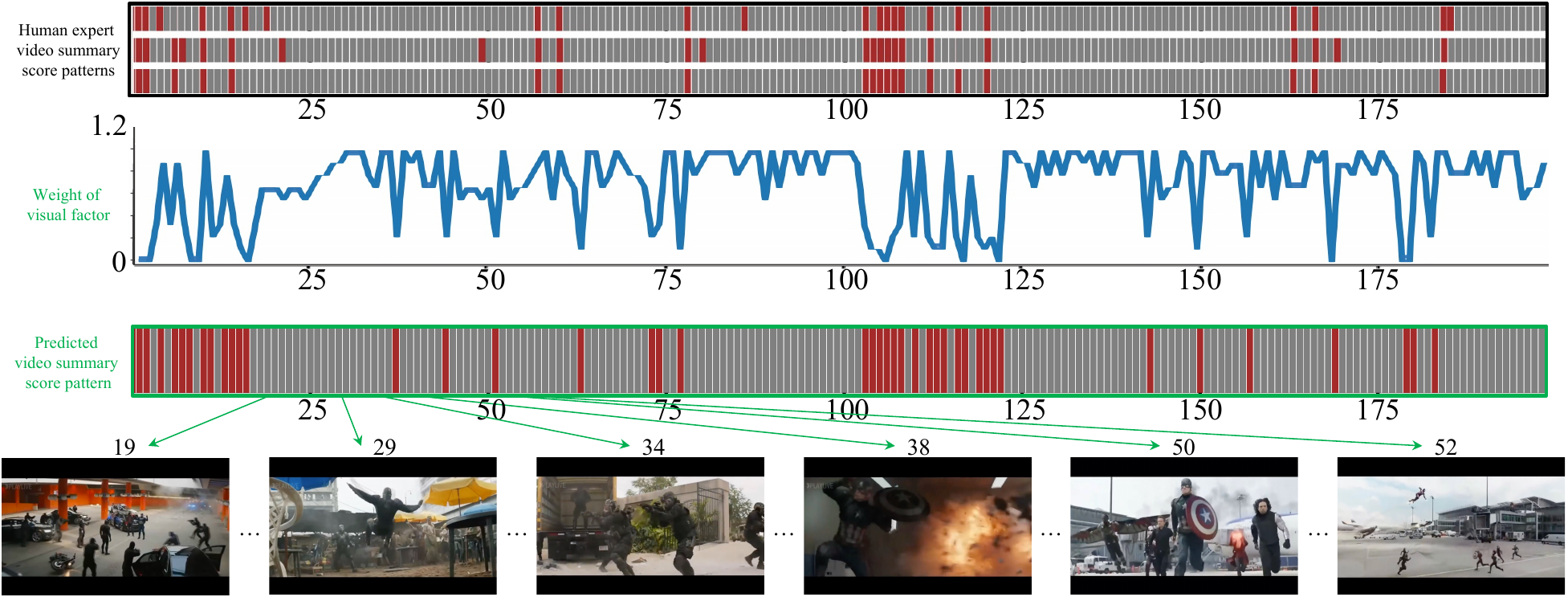}
\end{center}
    \captionof{figure}{Visualization of machine-predicted and human-annotated frame-level scores used for creating video summaries. In comparing the scoring patterns of human-annotated video summaries with those generated by contemporary state-of-the-art video summarization methods, such as \cite{vasudevan2017query,huang2020query,huang2021gpt2mvs}, it becomes evident that these methods proficiently capture the significance of visual consecutiveness and diversity. These factors, crucial for crafting effective video summaries, align with the criteria valued by human annotators. The primary focus of these advanced methods is on effectively incorporating visual elements to meet these objectives. An interesting observation arises when scrutinizing the three score patterns of human-annotated video summaries for the same video. It's apparent that these patterns differ, suggesting that in the process of video summarization, humans not only consider visual factors but also account for non-visual elements. Discarded frames are highlighted in red bars, while grey bars denote the selected frames used in constructing the video summary. The video comprises $199$ frames, with numbers indicating frame indices. The term ``Weight of visual factor'' is defined as the percentage or proportion of common visual objects present in a frame.
    }
\label{fig:discover_new_new_new_new_final}
\end{figure*}

When comparing these automatic algorithms with human experts, as emphasized by previous studies \cite{gygli2015video,song2015tvsum,vasudevan2017query,gong2014diverse,gygli2014creating,huang2020query}, it becomes apparent that human experts consider a diverse set of factors when crafting video summaries. These factors include both concrete visual elements, such as visual consecutiveness and diversity, and abstract non-visual elements, such as interestingness, representativeness, and storyline consistency. The interplay of these factors significantly influences the quality of the final video summary and underscores the need to account for them. However, existing video summarization methods predominantly focus on visual factors, as depicted in Figure \ref{fig:discover_new_new_new_new_final}, and often neglect or only marginally consider non-visual factors. Overlooking these non-visual aspects can lead to suboptimal performance in automatic video summarization \cite{huang2020query,jiang2022joint}.

\begin{figure}[t!]
\begin{center}
\includegraphics[width=1.0\linewidth]{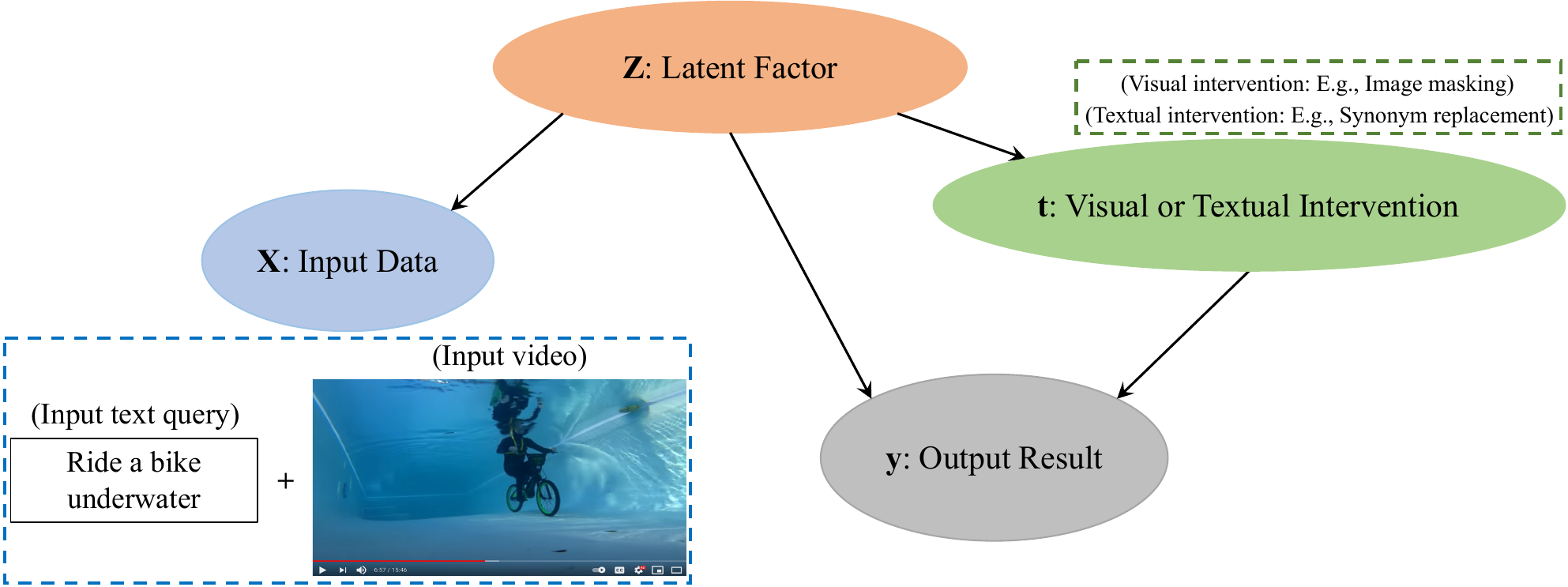}
\end{center}
  \caption{Example of a conditional graph in video summarization. $\textbf{t}$ is an intervention, e.g., visual or textual perturbation. $\textbf{y}$ is an outcome, e.g., an importance score of a video frame or a relevance score between the input text query and video. $\textbf{Z}$ is an unobserved factor, e.g., representativeness, interestingness, or storyline smoothness. $\textbf{X}$ is containing noisy views on the hidden factor $\textbf{Z}$, say the input text query and video. The conditional graph of video summarization leads to more explainable modeling.
  }
\label{fig:concept_explanation_causal_graphical_model}
\end{figure}

In this chapter, we embark on the understanding that the correlation between human-driven non-visual factors and the resultant summary is intricate and non-deterministic. Our approach delves into the complexity from a conditional modeling perspective. Drawing insights from studies such as \cite{song2015tvsum,gygli2014creating}, which delve into how humans assess the quality of video summaries (depicted in Figure \ref{fig:concept_explanation_causal_graphical_model}), we lay the groundwork for our proposed model. Our method is rooted in the concept of data intervention \cite{louizos2017causal}. Building upon intervention modeling techniques \cite{greenland1983correcting,selen1986adjusting}, we incorporate proxy variables to explore the conditions for conditional learning on large-scale datasets \cite{cai2012identifying,greenland2011bias,louizos2017causal,shalit2017estimating}. These methods augment the model's capacity to learn relationships by introducing additional information to the learning process.
Our proposed video summarization method involves four pivotal and meaningful random variables. These variables, characterizing the behavior of data intervention \cite{louizos2017causal}, the model's prediction, observed factors, and unobserved factors, collectively contribute to the effectiveness of our method.

The proposed method for modeling the insights on human summary assessment involves constructing both a prior joint distribution and its posterior approximation, based on the aforementioned four random variables. During training, the method is optimized by minimizing the distance between the prior distribution and the posterior approximation. However, accurately predicting the behaviors of the data intervention and the model outcome can prove challenging in practice, given factors such as video noise or motion/lens blur. To tackle this challenge, helper distributions are introduced, and a novel loss term is formulated to provide additional guidance for model learning.

\begin{figure*}[t!]
\begin{center}
\includegraphics[width=1.0\linewidth]{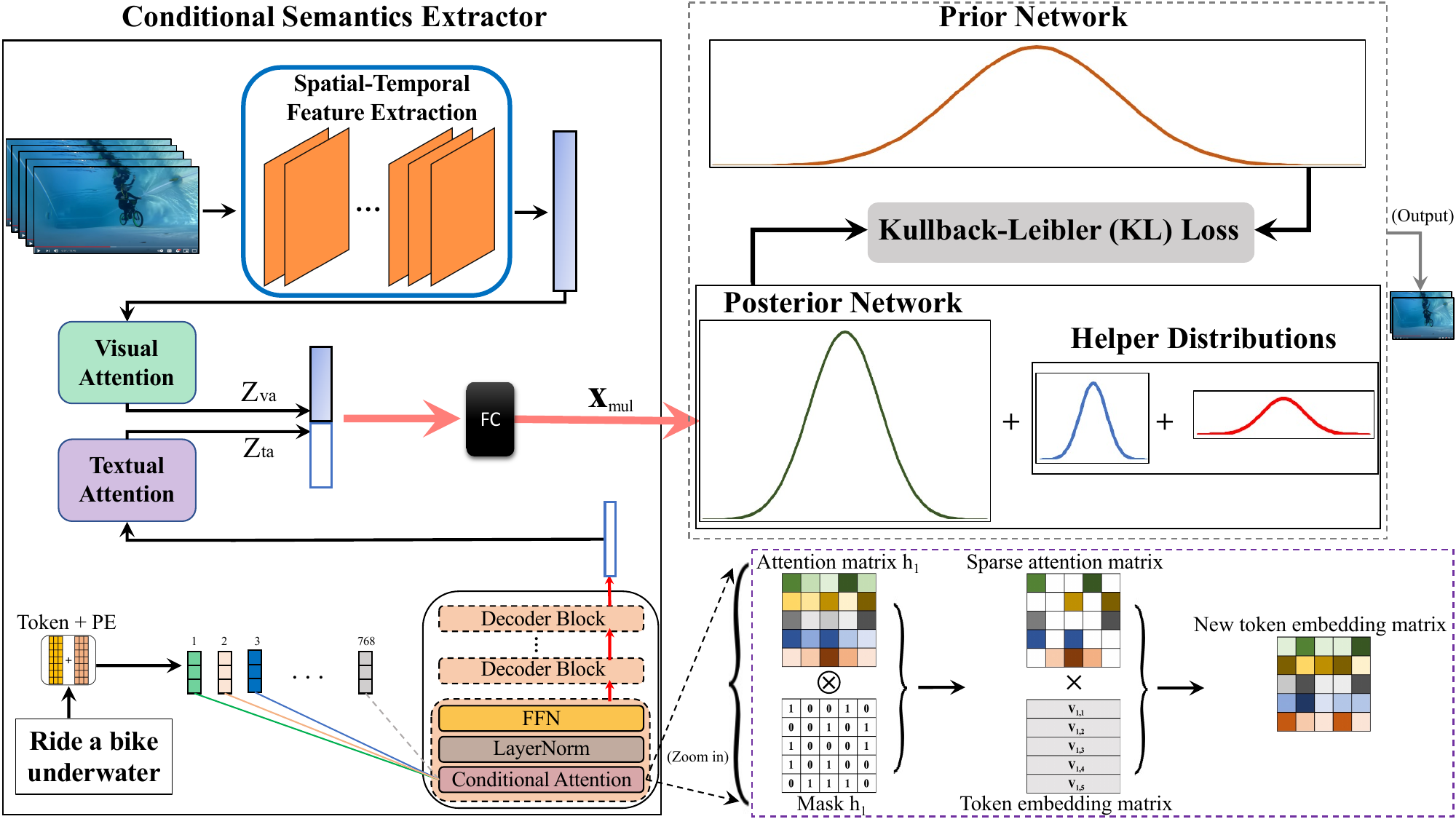}
\end{center}
   \caption{Flowchart of the proposed method for video summarization. The proposed method is mainly composed of a prior network, a posterior network, helper distributions, and a conditional attention module. $\otimes$ denotes element-wise multiplication and $\times$ indicates matrix multiplication. ``Token + PE'' denotes the operations of token embedding and positional encoding.
   }
\label{fig:causalainer_new}
\end{figure*}

Integrating textual information with visual data can occasionally detrimentally affect the model's performance due to ineffective interactions between different modalities. To address this challenge, we introduce a conditional attention module aimed at efficiently distilling mutual information from the multi-modal inputs. Figure \ref{fig:causalainer_new} provides a visual representation of the proposed approach's flowchart. Our video summarization method encompasses innovative design choices, including conditional modeling, helper distributions, and the conditional attention module, all directed at minimizing the disparity between human-generated and machine-generated video summaries.

Comprehensive experiments conducted on well-established video summarization datasets affirm that the proposed approach surpasses existing methods, achieving state-of-the-art performance.

\vspace{+3pt}
\noindent\textbf{Contributions.}

\begin{itemize}
    \item \textbf{Novel Conditional Modeling Perspective}: This chapter presents an innovative video summarization approach that concurrently integrates human-driven visual and non-visual factors. The proposed method adopts a novel conditional modeling perspective, drawing inspiration from human assessment of video summary quality. Additionally, it leverages the concept of data intervention to augment the model's capability in learning relationships among diverse factors impacting summarization. By considering both visual and non-visual elements, the approach seeks to elevate the quality and relevance of generated video summaries, aligning more closely with human perception standards.

    \item \textbf{Innovative Design Choices}: This chapter introduces innovative design choices, such as the utilization of helper distributions to improve model training and the design of a conditional attention module to address potential performance degradation in the presence of multi-modal input. These design choices aim to narrow the gap between human-generated and machine-generated video summaries, ultimately improving the quality and relevance of the generated summaries.

    \item \textbf{State-of-the-Art Performance}: Comprehensive experiments conducted on commonly used video summarization datasets demonstrate that the proposed approach outperforms existing methods and achieves state-of-the-art performance. By validating the effectiveness of the proposed method through rigorous experimentation, this chapter contributes to advancing the field of video summarization and offers a robust framework for generating high-quality video summaries.
\end{itemize}

The subsequent sections of this chapter are organized as follows: In Section 4.2, we offer a comprehensive review of related works. Section 4.3 details our proposed method, elucidating its design choices and implementation. In Section 4.4, we conduct a thorough evaluation of our approach, comparing it to existing state-of-the-art methods. Finally, Section 4.5 and Section 4.6 delve into the discussion of our findings and underscore potential future research directions.



\section{Related Work}
In the upcoming section, we offer a succinct overview of pertinent literature aligned with the proposed methodology. Video summarization poses a machine learning challenge that can be tackled through various supervision schemes, including fully supervised, weakly supervised, or unsupervised methods. Subsequent sections will present a compact summary of relevant methodologies within each category.

\subsection{Fully Supervised Methods with Visual Input Only}
Video summarization predominantly falls within the realm of fully supervised learning \cite{gong2014diverse,gygli2014creating,zhang2016video,zhao2017hierarchical,zhao2018hsa,zhang2019dtr,ji2019video,ji2020deep}, where human-defined labels guide the training process. These methods can be systematically categorized into three classes:

\noindent\textbf{Recurrent Neural Network (RNN)/Long Short-Term Memory (LSTM) Approaches}: This class encompasses methods that leverage RNN and LSTM architectures \cite{hochreiter1997long,zhao2017hierarchical,zhao2018hsa,zhang2019dtr,ji2019video}. In \cite{zhao2017hierarchical,zhao2018hsa}, hierarchical RNN architectures model temporal structures in video data, selecting shots or segments for the summary. \cite{zhang2019dtr} introduces a dilated temporal relational generative adversarial network that combines LSTM and dilated temporal relational units for frame-based video summarization. \cite{ji2019video} treats video summarization as a sequence-to-sequence learning problem, proposing an LSTM-based encoder-decoder model with an intermediate attention layer, later extended with a semantics-preserving embedding network \cite{ji2020deep}.

\noindent\textbf{Determinantal Point Process (DPP) Methods}: This class involves the application of DPP \cite{kulesza2012determinantal,zhang2016video}. In \cite{zhang2016video}, video summarization is framed as a structured prediction problem, with a deep-learning-based approach estimating the importance of video frames. LSTM captures temporal dependencies, and DPP enhances content diversity.

\noindent\textbf{Methods without RNN/LSTM and DPP}: This class encompasses methods that do not rely on RNN/LSTM and DPP \cite{gong2014diverse,gygli2014creating}. \cite{gong2014diverse} introduces a probabilistic model called the sequential DPP to account for the sequential structures in video data, selecting diverse subsets. This approach acknowledges intrinsic sequential structures, addressing the limitation of the standard DPP, which treats video frames as randomly permutable entities. \cite{gygli2014creating} proposes an automatic summarization method for user videos with interesting events, predicting visual interestingness features and selecting an optimal set of superframes for the summary.

The fully supervised approaches highlighted earlier heavily rely on comprehensive human expert annotations during model training, resulting in excellent performance. However, acquiring such annotations can be prohibitively expensive. Hence, there is a pressing need for a more cost-effective solution in video summarization. To address this challenge, we advocate for a more cost-effective alternative method grounded in conditional modeling. This approach not only offers enhanced interpretability but also improves generalization, provides increased flexibility, and enhances decision-making compared to existing methods.

\subsection{Fully Supervised Methods with Multi-modal Input}
Numerous methods have been devised to improve video summarization by leveraging supplementary modalities beyond visual inputs. These modalities encompass viewers' comments, video captions, and other contextual data sources \cite{li2017extracting, vasudevan2017query, sanabria2019deep, song2016category, zhou2018video, lei2018action, otani2016video, yuan2017video, wei2018video, huang2020query, huang2021gpt2mvs}.

For example, \cite{li2017extracting} presents a multi-modal video summarization technique for extracting key frames from first-person videos. In the work by \cite{sanabria2019deep}, a deep-learning-based method is proposed to summarize soccer game videos, leveraging multiple modalities. The authors of \cite{song2016category} introduce a model for category-driven video summarization aimed at preserving core elements found in summaries of the same category \cite{zhou2018video}. \cite{lei2018action} proposes a reinforcement learning-based approach that utilizes action classifiers trained with video-level labels for action-driven video fragmentation and labeling, followed by category-driven video summarization. Additionally, the studies \cite{otani2016video,yuan2017video} focus on generating video summaries by maximizing their relevance with the available metadata of the video, projecting visual and textual information onto a common latent feature space. Finally, The authors of \cite{wei2018video} employ a visual-to-text mapping and a semantic-based selection of video fragments based on the match between automatically generated and original video descriptions, utilizing semantic attended networks.

Multi-modal approaches, as demonstrated in the examples above, leverage additional modalities to enhance the performance of video summarization models. The effective fusion of these modalities is essential for the success of such approaches. However, inadequate fusion methods, as highlighted in previous studies \cite{huang2020query,vasudevan2017query,wei2018video}, can restrict the model's ability to fully exploit the complementary information offered by different modalities.

\subsection{Weakly Supervised Methods}
Several video summarization methodologies, as evidenced in \cite{panda2017weakly,ho2018summarizing,cai2018weakly,chen2019weakly,jiang2019comprehensive,yan2020self}, employ weakly supervised learning techniques to circumvent the requirement for vast datasets annotated by human experts. These approaches make use of more cost-effective weak labels, like video-level annotations from human experts, for model training. Despite the lower precision of weak labels compared to full human expert annotations, these approaches can proficiently train video summarization models, achieving satisfactory performance.

For instance, in \cite{panda2017weakly}, an intermediate approach between fully supervised and unsupervised learning is introduced, leveraging video-level metadata for video categorization. This method utilizes three-dimensional convolutional neural network (3D-CNN) features and a video summarization model to categorize new videos. Similarly, in \cite{cai2018weakly}, a hybrid model combining encoder-decoder architecture with soft attention and Variational AutoEncoder (VAE) is proposed. This model learns latent semantics from web videos using weakly supervised semantic matching loss during training to generate video summaries. In \cite{ho2018summarizing}, a model is presented that trains on fully annotated highlight scores from third-person videos and a limited number of annotated first-person videos, with only a fraction of the first-person videos having ground-truth annotations. Additionally, in \cite{chen2019weakly}, reinforcement learning is employed to train a video summarization model with a restricted set of human annotations and handcrafted rewards. This method utilizes a hierarchical key-fragment selection process and generates a final video summary based on rewards for diversity and representativeness.

The aforementioned weakly supervised video summarization methods offer a more cost-effective alternative to fully supervised approaches. However, it's worth noting that they may not always achieve optimal performance when compared to the latter.

\subsection{Unsupervised Methods}
Unsupervised approaches for video summarization operate under the premise that a good summary should encapsulate the essence of the video, a notion supported by evidence within the data itself. Thus, the focus lies on developing techniques to capture this intrinsic information \cite{zhao2014quasi,chu2015video,panda2017collaborative,mahasseni2017unsupervised,rochan2019video,herranz2012scalable,apostolidis2019stepwise,jung2019discriminative,yuan2019cycle,apostolidis2020unsupervised,sheinfeld2016video}.

In \cite{zhao2014quasi}, a dictionary is acquired from video data using group sparse coding. The resulting summary combines segments that resist sparse reconstruction based on the learned dictionary. \cite{chu2015video} introduces a maximal biclique finding algorithm to identify frequently co-occurring visual patterns, generating a summary by selecting shots with the highest co-occurrence frequency. In \cite{panda2017collaborative}, a video summarization model captures both specific and general features simultaneously. \cite{mahasseni2017unsupervised} proposes a model that integrates an LSTM-based key-frame selector, a trainable discriminator, and a VAE for summary creation through adversarial learning. \cite{apostolidis2019stepwise} advocates for a stepwise label-based strategy to enhance the adversarial training proposed in \cite{mahasseni2017unsupervised}. \cite{jung2019discriminative} presents a VAE-GAN architecture, and \cite{yuan2019cycle} employs a cycle-consistent adversarial learning objective to maximize mutual information between the summary and the video. In \cite{apostolidis2020unsupervised}, a modification of \cite{apostolidis2019stepwise} substitutes the VAE with a deterministic attention auto-encoder, resulting in a more effective key fragment selection. Finally, \cite{rochan2019video} introduces a novel formulation for video summarization, enforcing a diversity constraint on summary mapping through adversarial training.

Current unsupervised methods for video summarization do not utilize human expert annotations or pseudo labels for supervision during training, often resulting in performance that lags behind fully supervised methods. In contrast, our proposed approach harnesses both human expert annotations and pseudo-label supervision to gain an edge. It focuses on discerning cause-effect relationships between input data and desired summaries, a facet often overlooked in existing machine learning-based video summarization methods. Our method is adaptable and can be employed within both fully supervised and weakly supervised frameworks, depending on the experimental setup.

\section{Methodology}
\label{methodology:method}
Video summarization involves automatically creating a condensed representation of a video that captures its most crucial and informative content. Given an input video $\textbf{x}=(x_{0},x_{1},..., x_{n})$ consisting of $n$ frames, video summarization is treated as a classification problem. Here, the objective is to determine, through a classification function $f: y_{i}=f(x_{i})$, whether a frame $x_{i}$ is important ($y_{i} = 1$) or unimportant ($y_{i}=0$) to the summary. The classification problem is constrained by a user-defined summary budget $N_{Y}$, ensuring that $\sum_{i=0}^{n} y_{i} \leq N_{Y}$. The summary budget serves as a user-defined hyper-parameter \cite{huang2020query}.

This chapter presents an approach that models video summarization as a conditional learning problem. The proposed method utilizes conditional modeling to enhance the conditional inference ability of a machine learning-based video summarization model. The components involved include proxy variables $\textbf{X}$ and the hidden factor variable $\textbf{Z}$. The model aims to improve its inferential capabilities regarding the joint distribution $p(\textbf{X}, \textbf{Z})$, encompassing these proxy and hidden variables, by assimilating information from noisy interventions \cite{cai2012identifying,greenland2011bias,louizos2017causal,miao2018identifying,pearl2012measurement,kuroki2014measurement,wooldridge2009estimating,edwards2015all}. Subsequently, armed with this enhanced understanding, the model adeptly employs this knowledge to fine-tune and adjust the hidden factor variable.

We begin by presenting the assumptions of conditional modeling. Four random variables, $\textbf{y}$, $\textbf{t}$, $\textbf{X}$, and $\textbf{Z}$, are introduced, characterizing the model’s prediction behavior, data intervention, and observed and unobserved factors. $\textbf{X}$ represents an input video without or with a text query, $\textbf{t}$ denotes a visual or textual intervention assignment, and $\textbf{y}$ represents the importance score of a video frame for a video summary or the relevance score between an input text-based query and a video frame. The video summary is created based on $\textbf{y}$. The section further details the derivation of the training objective with helper distributions and introduces the proposed conditional attention module. The proposed approach comprises two probabilistic networks: the prior and posterior networks, as depicted in Figure \ref{fig:causalainer_new}.

\subsection{Assumptions}
In the realm of real-world observational studies, conditional modeling presents inherent complexities \cite{louizos2017causal,abbasnejad2020counterfactual,yang2021causal,agarwal2020towards}. Leveraging established methodologies \cite{pearl2018theoretical,louizos2017causal,zhang2018advances} tailored for conditional modeling under noisy interventions, we adopt two fundamental assumptions in the context of video summarization.

Firstly, we treat the presence or absence of visual/textual intervention, denoted as $\textbf{t}$, as binary information. Secondly, we assume that the observations $(\textbf{X}, \textbf{t}, \textbf{y})$ obtained from a deep neural network are sufficient to approximately reconstruct the joint distribution $p(\textbf{Z}, \textbf{X}, \textbf{t}, \textbf{y})$ encompassing the unobserved factor variable $\textbf{Z}$, the observed factor variable $\textbf{X}$, the intervention $\textbf{t}$, and the outcome $\textbf{y}$. Our proposed method builds upon these foundational assumptions and employs a range of probability distributions, which are elaborated upon in the subsequent subsections.

\subsection{Conditional Modeling for Video Summarization}
In our training process, we presume access to a collection of videos. Let $\textbf{x}_i$ represent an input video, possibly accompanied by a text-based query, indexed by $i$ in this collection. The latent factor is denoted by $\textbf{z}_i$, $t_i \in \{0,1\}$ signifies the intervention assignment, and $y_i$ denotes the outcome.

\noindent\textbf{Prior Probability Distributions}:
The prior network conditions on the latent variable $\textbf{z}_i$ and primarily comprises the following components:

\noindent(i) The latent factor distribution: 

\begin{align}
    p(\textbf{z}_i) = \prod_{z\in \textbf{z}_i} \mathcal{N}(z | \mu=0, \sigma^2=1),
\label{eq:eq1}
\end{align}
where we utilize $\mathcal{N}(\cdot | \mu, \sigma^2)$ to represent a Gaussian distribution with a random variable $z$ drawn from $\textbf{z}_i$, characterized by the mean $\mu$ and variance $\sigma^2$, following the conventions established in \cite{kingma2013auto}, specifically $\mu=0$ and $\sigma^2=1$.

\noindent(ii) The conditional data distribution:

\begin{align}
    p(\textbf{x}_i | \textbf{z}_i) = \prod_{x \in \ \textbf{x}_i} p(x | \textbf{z}_i),
\label{eq:eq2}
\end{align}
where the probability distribution $p(x |\textbf{z}_i)$ is suitable for modeling the distribution of a random variable $x$, conditioned on $\textbf{z}_i$, where $x$ is an element of $\textbf{x}_i$.

\noindent(iii) The conditional intervention distribution:

\begin{align}
    p(t_i | \textbf{z}_i) = \textup{Bernoulli} (\sigma(f_{\theta_1}(\textbf{z}_i))),
\label{eq:eq3}
\end{align}
where $\sigma(\cdot)$ represents a logistic function, $\textup{Bernoulli}(\cdot)$ denotes a Bernoulli distribution for modeling a discrete outcome, and $f_{\theta_1}(\cdot)$ signifies a neural network parameterized by $\theta_1$.

\noindent(iv) The conditional outcome distribution:

\begin{align}
    p(y_i | \textbf{z}_i, t_i) = \sigma(t_{i} f_{\theta_2}(\textbf{z}_i) + (1 - t_{i}) f_{\theta_3}(\textbf{z}_i)),
\label{eq:eq4}
\end{align}
where $f_{\theta_2}(\cdot)$ and $f_{\theta_3}(\cdot)$ represent neural networks parameterized by vectors $\theta_2$ and $\theta_3$, respectively.

\noindent In this chapter, $y_i$ is tailored for a categorical classification problem.

\noindent\textbf{Posterior Probability Distribution}:
Because we lack prior knowledge about the latent factor, we need to marginalize over it to learn the model parameters, $\theta_1$, $\theta_2$, and $\theta_3$ in Equations (\ref{eq:eq3}) and (\ref{eq:eq4}). The involvement of non-linear neural network functions makes direct inference intractable. Therefore, variational inference \cite{kingma2013auto} is employed in conjunction with the posterior network.

These neural networks produce the parameters of a fixed-form posterior approximation over the latent variable $\textbf{z}$, given the observed variables. Similar to \cite{louizos2017causal,rezende2014stochastic}, the proposed posterior network conditions on observations in this chapter. Moreover, the true posterior over $\textbf{Z}$ depends on $\textbf{X}$, $\textbf{t}$, and $\textbf{y}$. Thus, the below-defined posterior approximation is employed to construct the posterior network.

\begin{align*}
    q(\textbf{z}_i | \textbf{x}_i, y_i, t_i) = \prod_{z\in \textbf{z}_i} \mathcal{N}(z | \bm{\mu}_i, \bm{\sigma^2}_i), \nonumber
\label{eq:eq5}
\end{align*}

\begin{equation}
    \bm{\mu}_i = t_i \bm{\mu}_{t=1, i} + (1 - t_i) \bm{\mu}_{t=0,i}, \nonumber
    \label{equ_6}
\end{equation}

\begin{equation}
    \bm{\sigma^2}_i = t_i \bm{\sigma^2}_{t=1, i} + (1 - t_i) \bm{\sigma^2}_{t=0, i}, \nonumber
    \label{equ_7}
\end{equation}

\begin{equation}
    \bm{\mu}_{t=0, i} = g_{\phi_1} \circ g_{\phi_0}(\textbf{x}_i, y_i), \nonumber
    \label{equ_8}
\end{equation}

\begin{equation}
    \bm{\sigma^2}_{t=0, i} = \sigma(g_{\phi_2} \circ g_{\phi_0}(\textbf{x}_i, y_i)), \nonumber
    \label{equ_9}
\end{equation}

\begin{equation}
    \bm{\mu}_{t=1, i} = g_{\phi_3} \circ g_{\phi_0}(\textbf{x}_i, y_i), \nonumber
    \label{equ_10}
\end{equation}

\begin{equation}
    \bm{\sigma^2}_{t=1, i} = \sigma(g_{\phi_4} \circ g_{\phi_0}(\textbf{x}_i, y_i)),
    \label{equ_11}
\end{equation}
where $g_{\phi_k}(\cdot)$ represents a neural network with learnable parameters $\phi_k$, where $k$ takes integers from $0$ to $4$. The function $g_{\phi_0}(\textbf{x}_i, y_i)$ corresponds to a shared representation.

\subsection{Training Objective with Helper Distributions}
In real-world scenarios, accurately predicting the effects of data intervention and the model's outcomes can be challenging, given the presence of uncontrollable factors such as video noise, motion blur, or lens blur. To mitigate these challenges, two helper distributions are introduced.

\begin{figure}[t!]
\begin{center}
\includegraphics[width=1.0\linewidth]{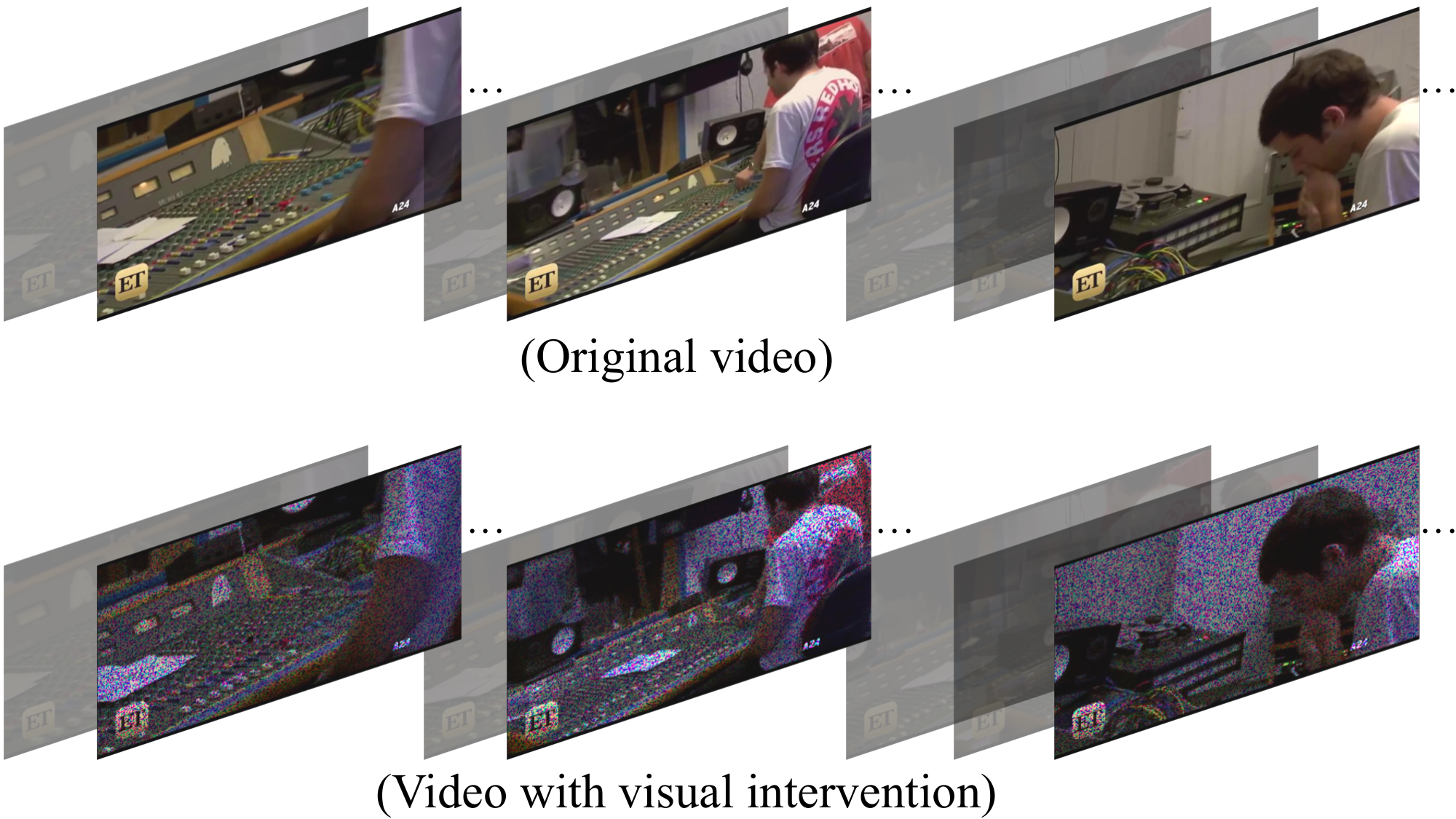}
\end{center}
  \caption{A dataset example demonstrating visual intervention with ``salt and pepper'' disturbance. 
  }
\label{fig:dataset_example_s-p_1}
\end{figure}

\begin{figure}[t!]
\begin{center}
\includegraphics[width=0.9\linewidth]{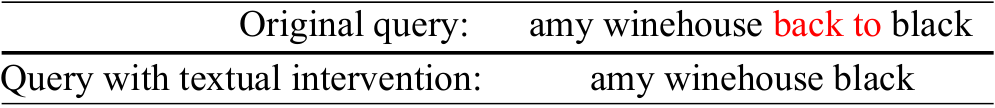}
\end{center}
  \caption{A dataset example demonstrating textual intervention with ``randomly missing some words in a sentence'' disturbance.}
\label{fig:dataset_example_remove_2_words}
\end{figure}

To infer the distribution over $\textbf{Z}$ effectively, knowledge of both the intervention $\textbf{t}$ and its outcome $\textbf{y}$ is essential. Therefore, a helper distribution introduced below is defined specifically for the intervention $\textbf{t}$. For visual illustrations of the role of $\textbf{t}$, please refer to Figures \ref{fig:dataset_example_s-p_1} and \ref{fig:dataset_example_remove_2_words}.

\begin{align}
    q(t_i | \textbf{x}_i) = \textup{Bernoulli}(\sigma(g_{\phi_5}(\textbf{x}_i))).
\end{align}

The other helper distribution introduced below is specifically designed for the outcome $y_{i}$.

\begin{align}
    q(y_i | \textbf{x}_i, t_i) = \sigma(t_{i} g_{\phi_6}(\textbf{x}_i) + (1 - t_{i}) g_{\phi_7}(\textbf{x}_i)),
\end{align}
where $g_{\phi_k}(\cdot)$ represents a neural network with variational parameters $\phi_k$ for indices $k=5, 6, 7$.

\noindent The introduced helper distributions contribute to the prediction of $t_i$ and $y_i$ for new samples. To estimate the variational parameters of the distributions $q(t_i | \textbf{x}_i)$ and $q(y_i | \textbf{x}_i, t_i)$, we introduce a helper objective function, as shown in Equation (\ref{equ_helper}), to the final training objective over $N$ data samples, where $\textbf{x}_i^*$, $t_i^*$, and $y_i^*$ are the observed values in the training set.

\begin{equation}
    \mathcal{L}_{\textup{helper}} = \sum^{N}_{i=1}[ \log q(t_i=t_i^{*} | \textbf{x}_i^{*}) + \log q(y_i=y_i^{*} | \textbf{x}_i^{*}, t_i^{*})].
    \label{equ_helper}
\end{equation}

Finally, the overall training objective $\mathcal{L}_{\textup{conditional}}$ for our proposed approach is defined below.

\begin{equation}
    \mathcal{L}_{\textup{conditional}} = \mathcal{L}_{\textup{helper}} ~ + \nonumber
\end{equation}

\begin{equation}
    \sum^{N}_{i=1}\mathbb{E}_{q(\textbf{z}_{i}|\textbf{x}_{i}, t_i, y_i)}[\log p(\textbf{x}_i, t_i | \textbf{z}_i) ~ + \nonumber
\end{equation}

\begin{equation}
    \log p(y_i|t_i, \textbf{z}_i) + \log p(\textbf{z}_i) - \log q(\textbf{z}_i|\textbf{x}_i, t_i, y_i)].
    \label{equ_15}
\end{equation}

\subsection{Conditional Attention Module}
Given that the textual input may not consistently enhance model performance due to challenges in effectively extracting mutual information from visual and textual inputs \cite{huang2020query,song2015tvsum}, we introduce a self-attention-based conditional attention module. This extractor is constructed using transformer blocks \cite{vaswani2017attention}. While vanilla transformers utilize all tokens in each layer for attention computation, the proposed conditional attention module is designed to effectively employ fewer but more informative tokens for attention map computation. The computation of the vanilla attention matrix $\mathcal{A} \in \mathbb{R}^{n\times n}$ is based on the dot-product \cite{vaswani2017attention}:

\begin{align}
    \mathcal{A} &= \textup{softmax}\left ( \frac{\mathbf{Q}\mathbf{K}^\top}{\sqrt{d}} \right );
    \mathbf{Q} = \mathbf{T}\mathbf{W}_{q},
    \mathbf{K} = \mathbf{T}\mathbf{W}_{k},
    \label{eq:eq11}
\end{align}
where the query matrix $\mathbf{Q} \in \mathbb{R}^{n\times d}$ and key matrix $\mathbf{K} \in \mathbb{R}^{n\times d}$ are generated by linearly projecting the input token matrix $\mathbf{T} \in \mathbb{R}^{n\times d_{m}}$ using learnable weight matrices $\mathbf{W}_{q} \in \mathbb{R}^{d_{m}\times d}$ and $\mathbf{W}_{k} \in \mathbb{R}^{d_{m}\times d}$. Here, $n$ represents the total number of input tokens, $d$ denotes the embedding dimension, and $d_{m}$ is the dimension of an input token.

The new value matrix $\mathbf{V}_{\textup{new}} \in \mathbb{R}^{n\times d}$ can be obtained as follows:

\begin{align}
    \mathbf{V}_{\textup{new}} = \mathcal{A}\mathbf{V};
    \mathbf{V} = \mathbf{T}\mathbf{W}_{v},
    \label{eq:eq22}
\end{align}
where the value matrix $\mathbf{V} \in \mathbb{R}^{n\times d}$ and the learnable weight matrix $\mathbf{W}_{v} \in \mathbb{R}^{d_{m}\times d}$.

In the proposed conditional attention module, to enhance computational efficiency, only the top $\kappa$ most similar keys and values for each query are used to compute the conditional attention matrix, as opposed to calculating all the query-key pairs in the vanilla attention matrix, as described in \cite{vaswani2017attention}. 

Similar to \cite{vaswani2017attention}, the calculation involves computing all the queries and keys using the dot-product. Then, the row-wise top $\kappa$ elements are selected for the \textup{softmax} calculation. In this module, the value matrix $\mathbf{V}_{\kappa} \in \mathbb{R}^{n\times d}$ is defined as:

\begin{align}
    \mathbf{V}_{\kappa} =\textup{softmax}\left (\tau _{\kappa} (\mathbf{\mathcal{A}}) \right ) \mathbf{V}_{\textup{new}}  \nonumber \\
    = \textup{softmax}\left (\tau _{\kappa} \left ( \frac{\mathbf{Q}\mathbf{K}^\top}{\sqrt{d}} \right )\right )\mathbf{V}_{\textup{new}},
    \label{eq:eq33}
\end{align}
where $\tau _{\kappa}(\cdot)$ represents an operator for the row-wise top $\kappa$ elements selection, where $\tau _{\kappa}(\cdot)$ is defined as:
\begin{align}
[\tau _{\kappa}(\mathbf{\mathcal{A}})]_{ij}=\begin{cases}
\mathcal{A}_{ij} &, \mathcal{A}_{ij}\in \text{top $\kappa$ elements at row~$i$} \\ 
-\infty  &, \text{ otherwise}.
\end{cases}
\label{eq:eq44}
\end{align}

Then, $\mathbf{V}_{\kappa}$ can be further employed to generate $\mathbf{X}_{\textup{mul}}$, representing an output of our proposed conditional attention module. The procedure for calculating $\mathbf{X}_{\textup{mul}}$ is defined below.

\begin{equation}
    Z_{\textup{ta}} = \textup{\textup{TextAtten}}(\textup{FFN}(\textup{LayerNorm}(\mathbf{V}_{\kappa})),
    \label{eq:zta}
\end{equation}
where $\textup{LayerNorm}(\cdot)$ represents layer normalization, $\textup{FFN}(\cdot)$ indicates a feed forward network, and $\textup{TextAtten}(\cdot)$ stands for a textual attention mechanism based on element-wise multiplication.

\begin{equation}
    Z_{\textup{va}} = \textup{VisualAtten}(\textup{C3D}(\textbf{x})),
    \label{eq:zva}
\end{equation}
where $\textbf{x}$ represents an input video, $\textup{C3D}(\cdot)$ denotes an operation for spatial-temporal feature extraction, such as the 3D version of ResNet-34 \cite{he2016deep,hara2018can}, applied to the input video, and $\textup{VisualAtten}(\cdot)$ signifies a visual attention mechanism employing element-wise multiplication.

\begin{equation}
    \mathbf{X}_{\textup{mul}} = \textup{FC}(Z_{\textup{ta}} \odot Z_{\textup{va}}),
    \label{eq:xmul}
\end{equation}
where the symbol $\odot$ denotes feature concatenation, and $\textup{FC}(\cdot)$ represents a fully connected layer. It's important to note that the output $\mathbf{X}_{\textup{mul}}$ of the conditional attention module serves as an input to the proposed posterior network, adhering to the strategy of utilizing multi-modal inputs.

Similar to the final step of video summary generation in \cite{huang2020query}, after completing the end-to-end training of the proposed conditional video summarization model, the trained model can be utilized for video summary generation. Finally, a set of video frames is selected from the original input video based on the generated score labels to form the final video summary.

\section{Experiments}
In this section, we begin by detailing the experimental setup and the datasets employed. Subsequently, we evaluate, analyze, and compare the effectiveness of the proposed video summarization method with existing state-of-the-art methods. Finally, we present a conditional graph, depicted in Figure \ref{fig:concept_explanation_causal_graphical_model}, to illustrate the enhancement in modeling explainability for video summarization.

\subsection{Experimental Setup and Datasets Preparation}
\label{section:section4-1}

\noindent\textbf{Experimental Setup}:
In this chapter, we explore three distinct scenarios. Firstly, in a fully supervised scheme, the proposed model is trained using a comprehensive dataset featuring human expert annotations, specifically frame-level labels. Secondly, within the framework of fully supervised learning and multi-modal input, we incorporate the text-based query as an additional input. Thirdly, informed by the empirical findings of \cite{song2015tvsum}, where a two-second segment length is identified as suitable for capturing local video context with strong visual coherence, we adopt a segment-level scoring approach. This involves generating a video segment-level score every two seconds based on provided frame-level scores. This segment-level label can be viewed as a form of weak label within a weakly supervised learning paradigm \cite{cai2018weakly,chen2019weakly,apostolidis2021video}.

\noindent\textbf{Video Summarization Datasets}:
To ensure a comprehensive and equitable evaluation of our proposed video summarization approach, we conducted experiments on established datasets widely used in the field, namely TVSum \cite{song2015tvsum}, QueryVS \cite{huang2020query}, and SumMe \cite{gygli2014creating}.

The TVSum dataset, presented by \cite{song2015tvsum}, comprises $50$ videos spanning diverse genres, including documentaries, how-to videos, news, egocentric videos, and vlogs. Each video is associated with a title that serves as a text-based query input. The dataset is annotated by $20$ crowd-workers per video. Video lengths in TVSum range from $2$ to $10$ minutes, and human expert frame-level importance scores range from $1$ to $5$.

QueryVS, introduced by \cite{huang2020query}, is a more extensive dataset featuring $190$ videos. In QueryVS, each video has frame-based annotations based on a frame rate of $1$ frame per second (fps). Video lengths range from $2$ to $3$ minutes, and human expert frame-level relevance scores range from $0$ to $3$. In QueryVS, each video is retrieved based on a provided text-based query.

SumMe, proposed by \cite{gygli2014creating}, is a benchmark dataset comprising $25$ videos. Each video is annotated with at least $15$ human summaries ($390$ in total), obtained through a controlled psychological experiment, offering an objective means to evaluate video summarization models and gain insights. Video durations in SumMe range from $1$ to $6$ minutes, and the importance scores annotated by human experts range from $0$ to $1$. Notably, SumMe is not employed for multi-modal video summarization, indicating that textual input is not accessible during model evaluations on this dataset.

Creating comprehensive video summarization datasets with human expert annotations that include query-based information on a large scale can be a costly endeavor. Consequently, commonly utilized datasets such as TVSum, QueryVS, and SumMe are often relatively compact in size.

To validate our proposed conditional modeling approach, we introduce three new video summarization datasets based on TVSum, QueryVS, and SumMe. For consistency, videos from these datasets are sampled at $1$ fps. The input image size is set to $224 \times 224$ with RGB channels, each normalized with a standard deviation of $(0.2737, 0.2631, 0.2601)$ and a mean of $(0.4280, 0.4106, 0.3589)$. Implementation and model training are carried out using PyTorch and an NVIDIA TITAN Xp GPU for $60$ epochs, employing a learning rate of $1e-6$. The Adam optimizer \cite{kingma2014adam} is utilized with hyper-parameters set to $\epsilon=1e-8$, $\beta_{1}=0.9$, and $\beta_{2}=0.999$. 

\noindent\textbf{Conditional Learning Dataset}:
\label{section:section3-5-1}
In observing people's writing behaviors, researchers have identified common patterns such as synonym replacement and inadvertent omission of words in sentences \cite{brand1985hot,shermis2014challenges}. Drawing inspiration from these observations, we randomly selected one behavior, specifically the ``accidental omission of words in a sentence'', and designed a textual intervention function to simulate it. Similarly, recognizing that visual disturbances can occur in daily life videos, such as salt and pepper noise, image masking, blurring, etc., we selected a couple of these, like ``blur'' and ``salt and pepper'' noise, and created a visual intervention function for simulation. 

Leveraging these controllable visual and textual intervention simulation functions, we generated a conditional video summarization dataset. The dataset creation involved two main steps. Firstly, $50\%$ of the \textit{(video, query)} pairs are randomly chosen from the original training, validation, and testing sets. Subsequently, for each selected video, $0$ or $1$ intervention labels are randomly assigned to $30\%$ of the video frames and their corresponding queries. Dataset examples with visual and textual interventions are illustrated in Figure \ref{fig:dataset_example_s-p_1} and Figure \ref{fig:dataset_example_remove_2_words}, respectively. It's worth noting that in the real world, there are various potential disturbances beyond those randomly selected for visual and textual interventions in this chapter. The other interventions can also be used in our proposed approach.

\subsection{Evaluation and Analysis}
\label{section:section4-2}

\noindent\textbf{Evaluation Protocol}:
Building upon prior research \cite{jiang2022joint,huang2021gpt2mvs,huang2020query,gygli2014creating,song2015tvsum}, we assess the proposed method within the same framework. The TVSum, QueryVS, and SumMe datasets are each randomly partitioned into five splits. For each split, $80\%$ of the data is allocated for training, while the remainder is reserved for evaluation.
To gauge the alignment between the generated video summaries $\mathbb{S}_{i}$ and the ground-truth counterparts $\hat{\mathbb{S}}_{i}$ for video $\mathbf{x}_{i}$, we employ the $F_{1}$-score metric \cite{hripcsak2005agreement,gygli2014creating,song2015tvsum,jiang2022joint}. This metric, which accounts for both precision and recall, is calculated based on the temporal overlap between $\hat{\mathbb{S}}_{i}$ and $\mathbb{S}_{i}$ as follows:

\begin{align}
    P = \frac{|\mathbb{S}_{i} \cap \hat{\mathbb{S}}_{i}|}{|\mathbb{S}_{i}|}, R = \frac{|\mathbb{S}_{i} \cap \hat{\mathbb{S}}_{i}|}{|\hat{\mathbb{S}}_{i}|}, F_{1}=\frac{2PR}{P+R}.
\end{align}

For videos featuring multiple human-annotated video summaries,, the calculation of metrics in \cite{jiang2022joint,huang2021gpt2mvs} is followed.

\begin{figure}[t!]
\begin{center}
\includegraphics[width=1.0\linewidth]{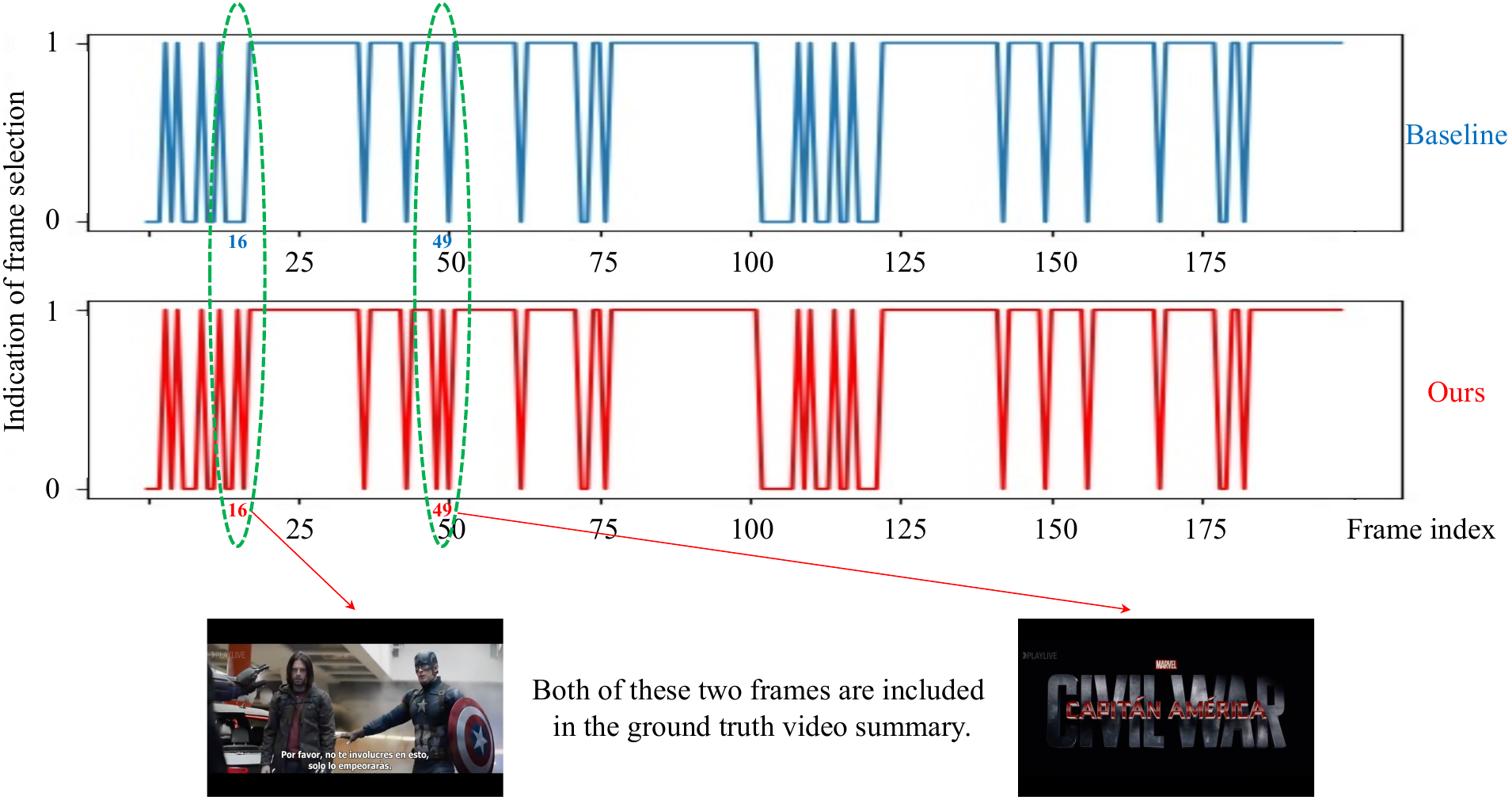}
\end{center}
  \caption{Qualitative analysis. This demonstration aims to highlight the efficacy of the proposed method in jointly considering visual and non-visual factors. The comparison is performed using the identical baseline method and a sample video, as illustrated in Figure \ref{fig:discover_new_new_new_new_final}.
  }
\label{fig:concept_explanation_causal_graphical_model9}
\end{figure}

\noindent\textbf{State-of-the-art Comparisons}:
In Table \ref{table:table2_chapter_4}, Table \ref{table:table3_chapter_4}, Table \ref{table:table4_chapter_4}, and Figure \ref{fig:concept_explanation_causal_graphical_model9}, we compare the proposed method with existing state-of-the-art models across different supervision schemes. The results consistently demonstrate the superiority of the proposed model over existing state-of-the-art methods. This superiority is attributed to the reinforced conditional modeling, which enhances the video summarization model's ability to infer causal relationships governing the process and influencing the outcomes.

\noindent\textbf{Ablation Studies}:
To assess the effectiveness of the proposed approach, we present the results of the ablation study in Table \ref{table:table1_chapter_4}. The findings indicate that the introduced components significantly enhance the model's performance. The key reasons behind this improvement are as follows: 
(i) The proposed conditional attention module adeptly captures interactions among various components of the method. 
(ii) There are explicit/implicit factors influencing the conditional inference of a video summarization model, and conditional attention effectively utilizes the causal contributions of these factors. 
(iii) The contextualized query representation, facilitated by conditional attention, proves more effective than Bag of Words (BoW) \cite{scott1998text,soumya2014text} for text-based query embedding. 
(iv) Incorporating a text-based query positively impacts model performance. 
(v) The use of a 3D CNN for video encoding surpasses the effectiveness of a 2D CNN. 
(vi) The introduction of helper distributions improves the prediction of data intervention behavior and the model's outcomes.

\begin{table}[t!]
    \caption{Comparative analysis with fully Supervised state-of-the-art approaches. The proposed method achieves the best performance on both datasets, even without the use of textual query input in this experiment.
    }
\centering
\scalebox{0.89}{
\begin{tabular}{c|ccc}
\toprule
\multicolumn{2}{c|}{\textbf{Fully Supervised Method}} & \textbf{TVSum}   &    \textbf{SumMe}    \\ 
\midrule
\multicolumn{2}{c|}{vsLSTM \cite{zhang2016video}}   & 54.2         & 37.6                                \\ 
\midrule
\multicolumn{2}{c|}{dppLSTM \cite{zhang2016video}}  & 54.7        & 38.6                       \\ 
\midrule
\multicolumn{2}{c|}{SASUM \cite{wei2018video}}    & 53.9       & 40.6                                           \\ 
\midrule
\multicolumn{2}{c|}{ActionRanking \cite{elfeki2019video}}   & 56.3       & 40.1                             \\ 
\midrule
\multicolumn{2}{c|}{H-RNN \cite{zhao2017hierarchical}}   & 57.7        & 41.1                                          \\ 
\midrule
\multicolumn{2}{c|}{DR-DSN$_{sup}$ \cite{zhou2018deep}}  & 58.1          & 42.1                                         \\ 
\midrule
\multicolumn{2}{c|}{PCDL$_{sup}$ \cite{zhao2019property}}    & 59.2       & 43.7                                    \\ 
\midrule
\multicolumn{2}{c|}{UnpairedVSN$_{psup}$ \cite{rochan2019video}}   & 56.1         & 48.0                                   \\ 
\midrule
\multicolumn{2}{c|}{SUM-FCN \cite{rochan2018video}}  & 56.8        & 47.5                                              \\ 
\midrule
\multicolumn{2}{c|}{SF-CVS \cite{huang2019novel_1}}    & 58.0         & 46.0                                              \\ 
\midrule
\multicolumn{2}{c|}{SASUM$_{fullysup}$ \cite{wei2018video}}   & 58.2         & 45.3                                \\ 
\midrule
\multicolumn{2}{c|}{A-AVS \cite{ji2019video}}    & 59.4       & 43.9                                             \\  
\midrule
\multicolumn{2}{c|}{CRSum \cite{yuan2019spatiotemporal}}   & 58.0       & 47.3                                       \\
\midrule
\multicolumn{2}{c|}{HSA-RNN \cite{zhao2018hsa}}  & 59.8           & 44.1                                                \\ 
\midrule
\multicolumn{2}{c|}{M-AVS \cite{ji2019video}}   & 61.0       & 44.4                                           \\ 
\midrule
\multicolumn{2}{c|}{ACGAN$_{sup}$ \cite{he2019unsupervised}}  & 59.4       & 47.2                                \\ 
\midrule
\multicolumn{2}{c|}{SUM-DeepLab \cite{rochan2018video}}    & 58.4         & 48.8                                  \\ 
\midrule
\multicolumn{2}{c|}{CSNet$_{sup}$ \cite{jung2019discriminative}}   & 58.5          & 48.6                               \\ 
\midrule
\multicolumn{2}{c|}{DASP \cite{ji2020deep}}  & 63.6        & 45.5                                    \\ 
\midrule
\multicolumn{2}{c|}{SMLD \cite{chu2019spatiotemporal}}    & 61.0          & 47.6       \\ 
\midrule
\multicolumn{2}{c|}{H-MAN \cite{liu2019learning}}   & 60.4          & 51.8        \\ 
\midrule
\multicolumn{2}{c|}{VASNet \cite{fajtl2019summarizing}}  & 61.4       & 49.7                                    \\ 
\midrule
\multicolumn{2}{c|}{iPTNet \cite{jiang2022joint}}    & 63.4    & \textbf{54.5}                       \\ 
\midrule
\rowcolor{mygray} \multicolumn{2}{c|}{\textbf{Ours}}    & \textbf{67.5}        & 52.4                           \\
\bottomrule
\end{tabular}}
\label{table:table2_chapter_4}
\end{table}

\noindent\textbf{Effectiveness Analysis of the Proposed Conditional Modeling}:
Given that the primary distinction between the proposed approach and existing methods lies in the introduced conditional modeling, the results in Table \ref{table:table2_chapter_4}, Table \ref{table:table3_chapter_4}, and Table \ref{table:table4_chapter_4} can be viewed as an ablation study of conditional learning across different supervision schemes. The outcomes underscore the effectiveness of the introduced conditional modeling. A pivotal element in our method is the auxiliary task/distribution. The role of this auxiliary task is to train the model to diagnose inputs accurately, enabling correct inferences for the main task, i.e., video summary inference, even in the presence of irrelevant interventions.

During training, the ground truth binary causation label is supplied to instruct the model on whether an intervention has occurred. If the model performs well regardless of the intervention's presence or absence, it indicates the model's ability to analyze inputs for optimal performance in the main task. Essentially, this implies the model possesses conditional inference ability and can be considered more robust. However, it's important to note that, unlike traditional robustness analysis, which involves applying small interventions to analyze a system \cite{huang2019novel}, the interventions in our proposed conditional modeling are designed to aid the model in learning the relationships within a system. Consequently, the strength of an intervention in our approach is not necessarily small; rather, the goal is to enhance the model's understanding of causal relationships.

\begin{table}[t!]
    \caption{Comparative analysis with the multi-modal state-of-the-art approaches. Our proposed approach outperforms the existing multi-modal methods. `-' indicates unavailability from previous work. 
    }
\centering
\scalebox{0.89}{
\begin{tabular}{c|ccc}
\toprule
\multicolumn{2}{c|}{\textbf{Multi-modal Method}} & \textbf{TVSum}   &    \textbf{QueryVS}    \\ 
\midrule
\multicolumn{2}{c|}{DSSE \cite{yuan2017video}}    & 57.0     & -    \\
\midrule
\multicolumn{2}{c|}{QueryVS \cite{huang2020query}}    & -      & 41.4  \\
\midrule
\multicolumn{2}{c|}{DQSN \cite{zhou2018video}}    & 58.6     & - \\
\midrule
\multicolumn{2}{c|}{GPT2MVS \cite{huang2021gpt2mvs}}    & -       & 54.8  \\
\midrule
\rowcolor{mygray} \multicolumn{2}{c|}{\textbf{Ours}}   & \textbf{68.2}     & \textbf{55.5} \\
\bottomrule
\end{tabular}}
\label{table:table3_chapter_4}
\end{table}

\begin{table}[t!]
    \caption{Comparative analysis with the weakly supervised state-of-the-art approaches. The performance of our proposed approach surpasses that of the existing weakly supervised method.
    }
\centering
\scalebox{0.89}{
\begin{tabular}{c|cc}
\toprule
\multicolumn{2}{c|}{\textbf{Weakly Supervised Method}} & \textbf{TVSum}      \\ 
\midrule
\multicolumn{2}{c|}{Random summary}  & 54.4         \\
\midrule
\multicolumn{2}{c|}{WS-HRL \cite{chen2019weakly}}  & 58.4    \\ 
\midrule
\rowcolor{mygray} \multicolumn{2}{c|}{\textbf{Ours}} & \textbf{66.9} \\ 
\bottomrule
\end{tabular}}
\label{table:table4_chapter_4}
\end{table}

\begin{table}[t!]
    \caption{Ablation study on our proposed video summarization approach. 
    The notation ``w/o'' denotes a model without using a specific component. ``w/o CM'' signifies the absence of conditional modeling. ``w/o C3D'' indicates the use of 2D CNN for video encoding. ``w/ BoW'' denotes the utilization of bag-of-words for query embedding instead of the Conditional Attention Module (CAM). ``w/o CAM'' implies the absence of the input textual query. ``w/o Helper Dist.'' indicates the exclusion of helper distributions. $\mathbf{\Delta_{1}}$ and $\mathbf{\Delta_{2}}$ represent the $F_{1}$-score differences compared to the full model.
    }
\centering
\scalebox{0.89}{
\begin{tabular}{c|ccccc}
\toprule
\multicolumn{2}{c|}{\textbf{Model}}  & \textbf{QueryVS}  & $\mathbf{\Delta_{1}}$ & \textbf{TVSum}  & $\mathbf{\Delta_{2}}$  \\ 
\midrule
\multicolumn{2}{c|}{w/o CM}  & 51.8 & -3.7  & 59.2  &  -9.0    \\
\midrule
\multicolumn{2}{c|}{w/o C3D}  & 53.8 & -1.7 & 63.7   &   -4.5  \\
\midrule
\multicolumn{2}{c|}{w/ BoW}  & 52.8 & -2.7 & 67.8   &  -0.4 \\
\midrule
\multicolumn{2}{c|}{w/o CAM}  & 52.3 & -3.2 & 67.5  &  -0.7 \\
\midrule
\multicolumn{2}{c|}{w/o Helper Dist.}  & 54.1 &  -1.4 & 66.7   & -1.5 \\
\midrule
\multicolumn{2}{c|}{w/o ConditionalAtten}  & 54.9 & -0.6 & 67.6   & -0.6  \\
\midrule
\multicolumn{2}{c|}{w/o VisualAtten}  & 55.1 & -0.4 & 67.9   &  -0.3 \\
\midrule
\multicolumn{2}{c|}{w/o TextAtten}  & 55.0 &  -0.5 & 67.7   & -0.5 \\
\midrule
\rowcolor{mygray} \multicolumn{2}{c|}{\textbf{Full model}} & \textbf{55.5} &  \textbf{0} & \textbf{68.2}  &  \textbf{0}  \\
\bottomrule
\end{tabular}}
\label{table:table1_chapter_4}
\end{table}

\section{Discussion}
\subsection{Why Video Summarization Can Benefit from Conditional Modeling?}

\noindent \textbf{Enhanced Interpretability}: Conditional modeling provides a structured framework for comprehending the interdependencies among various variables within a system, thereby facilitating the development of more interpretable video summarization models. By explicitly capturing the relationships between different components of a video, such as objects, events, or scenes, we can gain deeper insights into the factors influencing the overall content and structure of the video summary. Leveraging a conditional graph for video summarization, as depicted in Figure \ref{fig:concept_explanation_causal_graphical_model}, enhances the interpretability of the modeling process.

\noindent \textbf{Improved Generalization}: Conditional modeling aids in constructing video summarization models with enhanced generalizability across diverse video domains and contexts. By modeling the causal connections between different variables, we can grasp the underlying mechanisms governing the system's behavior and leverage this understanding to design more robust and adaptable video summarization models. 

\noindent \textbf{Enhanced Flexibility}: Conditional modeling offers a flexible framework for developing video summarization models that can easily adapt to varying scenarios and contexts. By explicitly modeling the relationships between different variables, we can adjust and refine the model based on new observations or changes in the underlying data, eliminating the need for retraining the entire model from scratch.

\noindent \textbf{Better Decision-making}: Conditional modeling empowers the development of video summarization models capable of making more informed and accurate decisions regarding the content to include in the summary. By capturing the causal relationships between different elements of the video, such as visual factors, semantic information, or temporal dynamics, we can better understand how these factors contribute to the overall content and structure of the summary, enabling more informed decision-making processes regarding content inclusion and exclusion.

\subsection{Main Limitation of the Proposed Approach}
Typically, the proposed conditional model requires annotated data for training, making it inappropriate to categorize it as an unsupervised video summarization method. While the proposed method exhibits strong performance, the associated cost of annotated data is a practical consideration that cannot be overlooked in real-world applications. In practice, a more balanced approach is to opt for a weakly supervised model, which offers a good compromise between performance and deployment cost, as illustrated in Table \ref{table:table4_chapter_4}.

\section{Conclusion}  
In conclusion, this chapter presents a significant advancement in video summarization by addressing the limitations of current automatic methods. The proposed approach, integrating both visual and non-visual factors based on human expert insights, excels in creating higher-quality video summaries aligned with human perception standards. Leveraging a conditional modeling perspective and data intervention, the method effectively learns intricate relationships among diverse factors. Innovative design choices, including helper distributions and a conditional attention module, contribute to improved model training and mitigate potential issues with multi-modal input. Comprehensive experiments showcasing superior accuracy and $F_{1}$-score underline its significance in the field. This research offers a robust framework for video summarization, enhancing accessibility and comprehension by capturing the richness of visual and non-visual aspects in video content. With the rapid growth of video content, this method has the potential to significantly improve video exploration efficiency, empowering users to efficiently process and comprehend vast amounts of video data for better decision-making and exploration across various applications, optimizing resource utilization and enhancing user experience.

\chapter{Pseudo-Label Supervision Enhances Video Summarization}
\label{ch:things}

\section{Abstract}
Query-dependent video summarization is a challenging task commonly addressed as a fully supervised machine learning problem. However, the construction of large-scale manually annotated datasets for such tasks is prohibitively expensive. To tackle this challenge, self-supervised learning has emerged as a promising strategy, offering a means to balance the costs associated with data labeling and the performance gains of fully supervised deep models. Nonetheless, prevalent self-supervision methods often assume a limited connection between human-defined labels and pseudo labels, which may not hold true for query-based video summarization. This chapter introduces a segment-based video summarization pretext task with specially designed pseudo labels, aiming to capture the implicit relations between pretext and target tasks. Additionally, a novel mutual attention mechanism combined with a semantics booster is proposed to effectively capture interactive information between textual and visual modalities, enhancing model performance. Extensive experiments conducted on widely used video summarization benchmarks validate the effectiveness of the proposed approach, highlighting its state-of-the-art performance in query-based video summarization. Contributions include a novel segment-based pretext task, a mutual attention mechanism with a semantics booster, and state-of-the-art performance demonstrated through comprehensive experiments.

\section{Introduction}
\label{sec:intro}
\begin{figure}[t!]
\begin{center}
\includegraphics[width=1.0\textwidth]{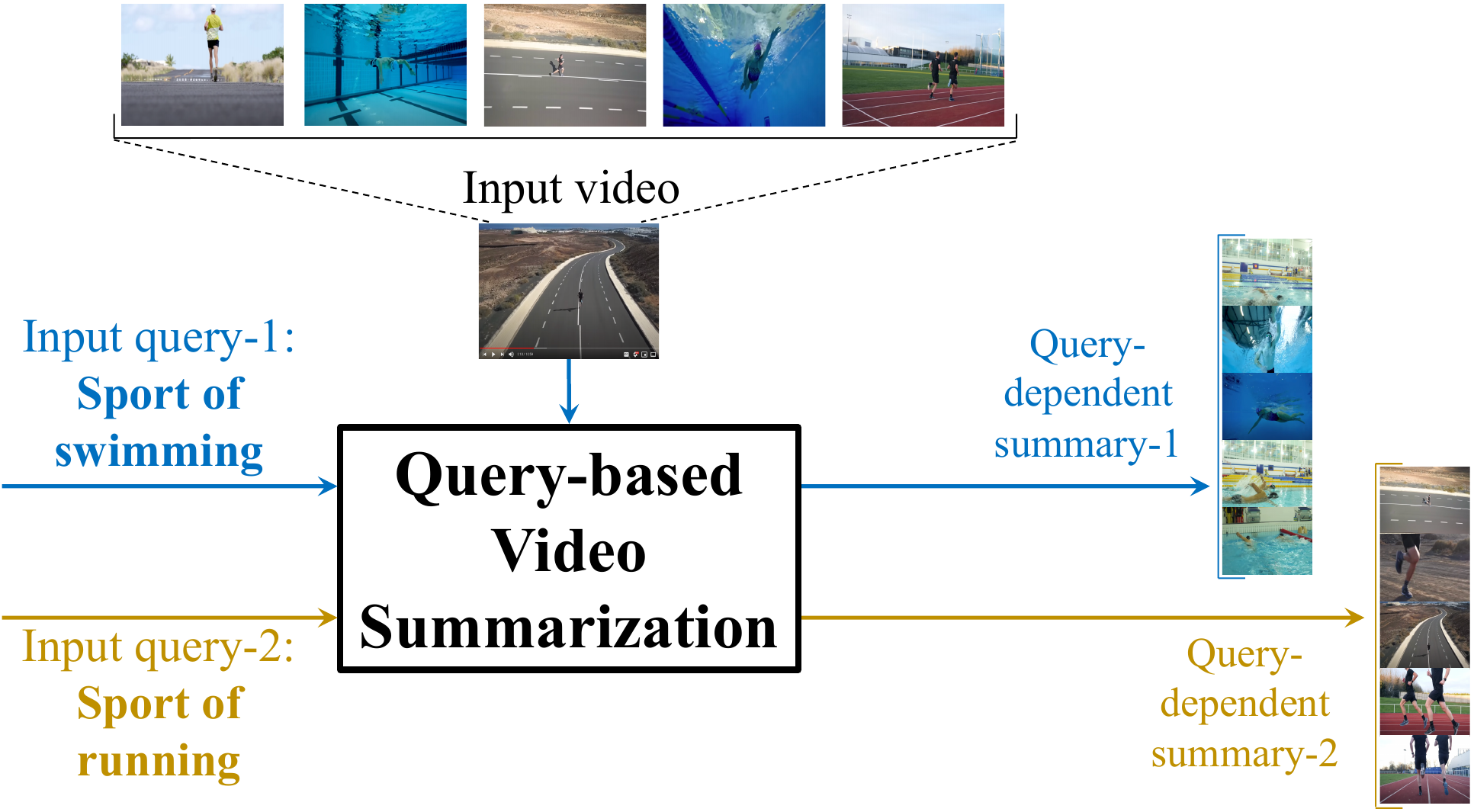}
\end{center}
   \caption{Textual query-driven video summarization. Query-based video summarization involves summarizing a video based on textual queries. The summarization algorithm operates independently for each query, generating a tailored summary corresponding to the specific input query.}
\label{fig:figure_1luka}
\end{figure}

The generation of query-dependent video summaries, as illustrated in Figure~\ref{fig:figure_1luka}, is commonly approached as a fully supervised machine learning task \cite{vasudevan2017query,huang2020query,huang2021gpt2mvs}. This modeling scheme is favored for its ability to achieve relatively superior performance compared to other supervision schemes. However, the construction of large-scale manually annotated video datasets for such fully supervised tasks proves to be prohibitively expensive in the context of video summarization. Consequently, datasets for such tasks tend to be relatively small, exemplified by TVSum \cite{song2015tvsum}, SumMe \cite{gygli2014creating}, and QueryVS \cite{huang2020query}.

The insufficiency of extensive human-annotated datasets poses a common challenge in fully supervised deep learning tasks. To tackle this limitation, self-supervised learning has emerged as a promising strategy \cite{doersch2015unsupervised,alwassel2020self,lai2019self,kim2019self}. Self-supervision, as emphasized in \cite{alwassel2020self,caron2018deep}, provides a means to strike a balance between the costs associated with data labeling and the performance gains of fully supervised deep models. In essence, self-supervised learning entails defining a pretext task and utilizing additional data with reliable pseudo labels to pre-train a fully supervised deep model for a target task \cite{doersch2015unsupervised,alwassel2020self}.

Nonetheless, prevalent self-supervision methods often operate under the assumption of a limited or non-existent connection between the target task, characterized by human-defined labels, and the pretext task, driven by pseudo labels \cite{alwassel2020self,caron2018deep}. This assumption may not hold for query-based video summarization, where human-defined labels for video frames act as supervision signals for the target task, while pseudo labels for video segments are utilized for the pretext task. Given that a video segment is composed of individual frames, an implicit relationship naturally exists between the entire segment and its constituent frames. Neglecting these implicit relations could hinder the improvement of model performance.

\begin{figure*}[t!]
{\includegraphics[width=1.0\textwidth]{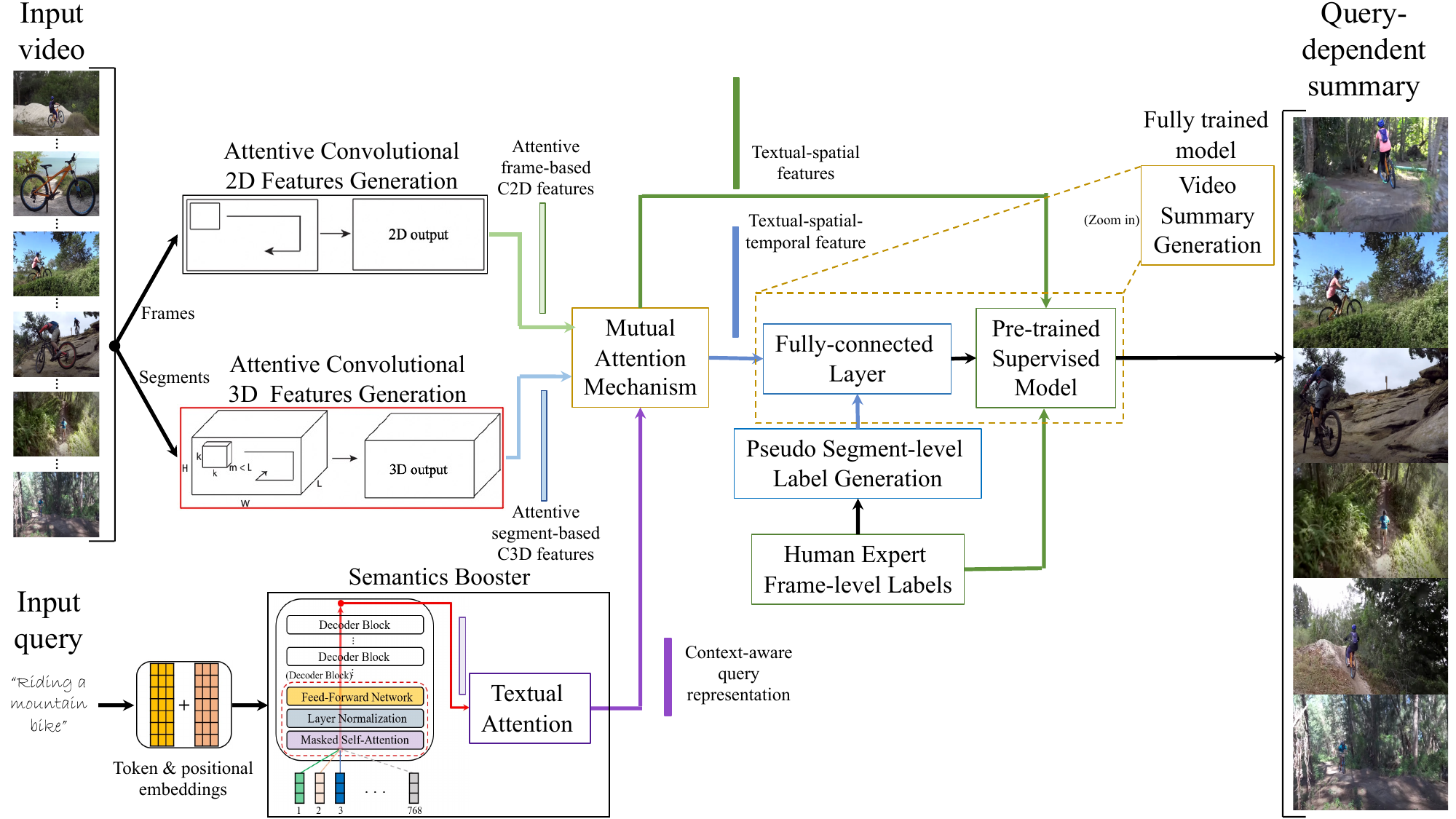}}
{
\caption{Flowchart illustrating our self-supervision method proposed for query-based video summarization. The model undergoes pre-training utilizing textual-spatial features obtained from the ``Mutual Attention Mechanism'' and pseudo segment-level labels. Once fully trained, the video summary generator utilizes a fully-connected layer to generate a frame-level score vector for the input video, culminating in the production of the final query-dependent video summary.}
\label{fig:figure_2luka}
}
\end{figure*}

To tackle the aforementioned challenge, this chapter introduces a segment-based video summarization pretext task with specially designed pseudo labels, as depicted in Figure~\ref{fig:figure_2luka}. These pseudo labels are derived from existing human-defined annotations, aiming to capture the implicit relations between the pretext task and the target task, which is frame-based video summarization \cite{huang2020query,song2015tvsum,gygli2014creating}. Generating accurate query-dependent video summaries in query-based video summarization proves challenging due to the ineffective embedding of semantics from textual queries. To counteract this, we propose a semantics booster that generates context-aware query representations capable of efficiently capturing semantics. Furthermore, we notice that the query input does not consistently enhance model performance, likely due to the inadequate modeling of interactions between textual and visual modalities. To overcome this challenge, we introduce mutual attention to capture interactive information between different modalities.

These innovative design choices significantly bolster the model performance of query-based video summarization. Comprehensive experiments confirm the effectiveness of the proposed approach, showcasing its state-of-the-art performance. Furthermore, considering the dichotomy between frame-level and segment-level labels, the proposed approach can also be interpreted as a form of weakly supervised video summarization. Therefore, we include existing weakly supervised methods as baselines in this chapter for comparison purposes.

\vspace{+3pt}
\noindent\textbf{Contributions.}

\begin{itemize}
    \item \textbf{Novel Segment-Based Pretext Task}: This chapter introduces a segment-based video summarization pretext task with specially designed pseudo labels, derived from existing human-defined annotations. This innovative approach aims to capture the implicit relations between the pretext task and the target task, i.e., frame-based video summarization, thereby addressing the challenge of limited human-annotated datasets in fully supervised deep learning tasks.

    \item \textbf{Mutual Attention Mechanism with Semantics Booster}: A novel mutual attention mechanism combined with a semantics booster is introduced to effectively capture interactive information between various modalities, particularly textual and visual modalities. This innovative design choice addresses the challenge of insufficiently modeling interactions between textual and visual modalities, thereby enhancing the overall performance of the video summarization model.

    \item \textbf{State-of-the-Art Performance}: Extensive experiments have been undertaken to confirm the efficacy of the proposed approach, highlighting its cutting-edge performance in query-based video summarization. Furthermore, an ablation study has been conducted to showcase the effectiveness of the proposed approach in overcoming the challenges associated with fully supervised deep learning tasks within the realm of query-based video summarization.
\end{itemize}

The subsequent sections of this chapter are organized as follows: In Section 5.2, we offer a comprehensive review of related work to contextualize our approach within the existing landscape. Following that, Section 5.3 explores the specifics of our method, delineating its key design choices and implementation details. To evaluate its effectiveness, Section 5.4 presents a rigorous evaluation comparing our approach to state-of-the-art methods. Finally, Section 5.5 summarizes our key findings and concludes the chapter.

\section{Related Work}

In this section, we provide a concise overview of existing video summarization methods categorized into fully supervised, weakly supervised, and self-supervised schemes. Additionally, we delve into the discussion of commonly employed word embedding methods.

\subsection{Fully Supervised Video Summarization}
Fully supervised learning stands as a prevalent approach in modeling video summarization \cite{gygli2014creating,zhang2016video,zhao2017hierarchical,zhao2018hsa,jiang2022joint}. This method entails leveraging labels provided by human experts to supervise the model during the training phase.

In the work by \cite{gygli2014creating}, an automated video summarization technique is introduced to summarize user-generated videos featuring noteworthy events. The methodology involves video segmentation based on superframes tailored to raw videos. Various levels of features are then employed to predict visual interestingness scores for each superframe. The final video summary is generated through an optimized selection of superframes.
In \cite{zhao2017hierarchical,zhao2018hsa}, a hierarchical Recurrent Neural Network (RNN) is deployed to model temporal structures within video data. Meanwhile, \cite{zhang2016video} treats video summarization as a structured prediction challenge, proposing a deep learning-based method to estimate the importance of video frames by modeling their temporal dependencies.
A novel approach is presented in \cite{jiang2022joint}, where an importance propagation-based collaborative teaching network (iPTNet) is proposed for video summarization. This method transfers samples from a video moment localization task, correlated with abundant training data.

Furthermore, \cite{huang2020query,huang2021gpt2mvs,huang2022causal,huang2023causalainer,yuan2017video,zhou2018video} broaden the scope beyond visual input, incorporating additional modalities such as viewers' comments, video captions, or contextual data for enhanced video summarization.

The mentioned fully supervised methods rely on a comprehensive set of human expert annotations to supervise the model during training, yielding strong performance. However, this approach comes with a significant cost. Consequently, there is a need for a more cost-effective solution in the realm of video summarization.

\subsection{Weakly Supervised Video Summarization}
In works such as \cite{panda2017weakly,ho2018summarizing,chen2019weakly}, the task of video summarization is approached as a weakly supervised learning problem. This paradigm offers a more economical alternative to extensive datasets with detailed human expert annotations. Instead of relying on a comprehensive dataset with meticulous frame-level labels, weakly supervised methods leverage more affordable annotations, such as video-level labels provided by human experts.

While these weak labels may not possess the precision of a complete set of human expert annotations, they prove to be effective in training video summarization models.

\subsection{Self-supervision in Video Summarization}
In works like \cite{yan2020self,jiang2019comprehensive}, the concept of image pretext tasks \cite{alwassel2020self} is expanded to the domain of videos, forming a foundation for self-supervised video summarization. In \cite{yan2020self}, the identification of keyframes involves singling out frames with distinct optical flow features and appearances compared to the rest. Meanwhile, \cite{jiang2019comprehensive} emphasizes the importance of a video sequence encoder that accurately captures the temporal order of video segments. This is achieved by selecting segments from a video in fixed proportions, shuffling them randomly, and then using them to train a neural network. The odd-position segments are particularly singled out, introducing a controlled difficulty to the auxiliary self-supervision task.

The realm of self-supervised learning in video summarization remains largely unexplored, with a dearth of emphasis on query-based video summarization in existing literature. Our proposed method stands as one of the pioneering endeavors in the realm of self-supervised techniques tailored specifically for query-based video summarization.

\subsection{Word Embedding Methods}
According to \cite{ethayarajh2019contextual}, when encoding textual data, both static word embeddings and contextualized word representations are commonly employed, with both methods proving more effective than the traditional Bag of Words (BoW) approach. Models like ``Skip-gram with negative sampling'' \cite{mikolov2013distributed} and ``global vectors for word representation'' \cite{pennington2014glove} are well-established for generating static word embeddings, with research \cite{levy2014linguistic,levy2014neural} confirming their iterative learning process. However, it's been established that these models implicitly factorize a word-context matrix containing co-occurrence statistics.

In static word embeddings, as noted by \cite{ethayarajh2019contextual}, all meanings of a polysemous word must share a single vector due to the creation of a singular representation for each word. Therefore, the contextualized word representation method surpasses static word embeddings due to its context-sensitive word representations. Notably, neural language models proposed by \cite{devlin2018bert,peters2018deep,radford2019language} are fine-tuned to develop deep learning-based models tailored for a wide array of downstream natural language processing tasks.

In this chapter, we leverage a contextualized word representation-based approach to encode the text-based input query, aligning with the insights provided by existing literature.

\section{Methodology}
In this section, we provide a comprehensive description of our proposed approach, as depicted in Figure~\ref{fig:figure_2luka}. The approach is grounded in contextualized query representations, attentive convolutional 2D and 3D features, an interactive attention mechanism, mean-based pseudo-shot label generation, and the subsequent process of video summary generation.

\subsection{Semantics Booster}
Creating an accurate query-dependent video summary poses challenges due to the ineffective embedding of input textual queries' semantics. To address this, a semantics booster is introduced in this chapter to effectively capture query semantics. The transformer-based model architecture has emerged as a leading approach in language modeling and machine translation \cite{vaswani2017attention}. Consequently, the proposed semantics booster is constructed atop the transformer architecture to generate context-aware query representations. This booster enhances the understanding of query semantics and facilitates more precise video summarization.

For an input token $k_{n}$, its embedding $x_{n}$ is obtained as  $x_n = \mathbf{W}_{e} * k_{n} + P_{k_{n}}, n \in \{1,..., N\}$, where $\mathbf{W}_{e} \in \mathbb{R}^{E_{s} \times V_{s}}$ represents the input text-based query token embedding matrix with a vocabulary size of $V_{s}$ and a word embedding size of $E_{s}$. The positional encoding of $k_{n}$ is denoted as $P_{k_{n}}$, and $N$ indicates the number of input tokens. Here, the subscripts $s$ and $e$ represent size and embedding, respectively.

The current word's representation, denoted as $Q$, is produced by a linear layer defined as $Q = \mathbf{W}_{q} * x_{n}+b_{q}$. Here, $b_{q}$ and $\mathbf{W}_{q} \in \mathbb{R}^{H_{s} \times E_{s}}$ represent the learnable parameters of the linear layer. The output size of the linear layer is $H_{s}$, and the subscript $q$ indicates ``query''.

The key vector $K$ is computed using another linear layer defined as $K = \mathbf{W}_{k} * x_{n}+b_{k}$, where $b_{k}$ and $\mathbf{W}_{k} \in \mathbb{R}^{H_{s} \times E_{s}}$ are learnable parameters of the linear layer. The subscript $k$ indicates ``key''.

Likewise, the value vector $V$ is produced by another linear layer defined as $V = \mathbf{W}_{v} * x_n+b_{v}$, where $b_{v}$ and $\mathbf{W}_{v} \in \mathbb{R}^{H_{s} \times E_{s}}$ are learnable parameters of the linear layer. The subscript $v$ indicates ``value''.

After computing $Q$, $K$, and $V$, the masked self-attention is derived as: 

\begin{equation}
    \textup{MaskAtten}(Q,K,V) = \textup{softmax}(m(\frac{QK^\intercal}{\sqrt{d_{k}}}))V,
\end{equation}
where $m(\cdot)$ represents a masked self-attention function and $d_{k}$ is a scaling factor.

Layer normalization is then performed as: $Z_{\textup{Norm}} = \textup{LayerNorm}(\textup{MaskAtten}(Q,K,V))$, utilizing the layer normalization function denoted by $\textup{LayerNorm}(\cdot)$.

Subsequently, the context-aware representation $\mathcal{R}_\textup{context}$ of the input text-based query is computed as:

\begin{equation}
    \mathcal{R}_\textup{context} = \sigma(\mathbf{W}_{1} Z_{\textup{Norm}}+b_{1}) \mathbf{W}_{2}+b_{2},
\end{equation}
where $\sigma$ denotes an activation function, and $\mathbf{W}_{1}$, $\mathbf{W}_{2}$, $b_{1}$, and $b_{2}$ are learnable parameters of a position-wise feed-forward network.

To further improve textual representations, a textual attention function $\textup{TextAtten}(\cdot)$ is introduced to enhance the context-aware representation. This function operates on $\mathcal{R}_\textup{context}$ as input and computes the attention and textual representation element-wise. The resulting attentive context-aware representation is obtained as:

\begin{equation}
    Z_{\textup{ta}} = \textup{TextAtten}(\mathcal{R}_\textup{context}),
\end{equation}
where $\textup{ta}$ signifies textual attention.

\subsection{Visual Attention}
The proposed method leverages both a 2D ConvNet and a 3D ConvNet to extract information from individual video frames and video segments, respectively. Additionally, a visual attention function, denoted as $\textup{AttenVisual}(\cdot)$, is introduced to enhance the quality of the generated 2D and 3D features.

Let $E$ represent a feature generator and $X$ a set of video clips. The feature generator $E$ maps an input $x \in X$ to a feature vector $f \in \mathbb{R}^{d}$. We denote the set of features produced by $E$ as $F=\{f=E(x) \in \mathbb{R}^{d} ~|~x \in X\}$. Additionally, let $F_{s}$ denote the features generated by the video spatial feature generator $E_{s}$, and $F_{\textup{st}}$ represent the features generated by the video spatiotemporal feature generator $E_{\textup{st}}$. The proposed query-based video summarization model utilizes both frame-level and segment-level data for training, which implies $F = F_{s} \cup F_{\textup{st}}$.

In the frame-level scenario, the attentive feature generator $\textup{AttenVisual}(\cdot)$ learns attention weights and generates attentive spatial features denoted as $Z_{\textup{as}}=\{f_{\textup{as}}=\textup{AttenVisual}(f) \in \mathbb{R}^{d} ~|~f \in F_{s}\}$. These represent attentive convolutional 2D features. On the other hand, in the segment-level case, the attentive feature generator also learns attention weights and produces attentive spatio-temporal features. These are referred to as $Z_{\textup{ast}}=\{f_{\textup{ast}}=\textup{AttenVisual}(f) \in \mathbb{R}^{d} ~|~f \in F_{\textup{st}}\}$ and represent attentive convolutional 3D features.

\subsection{Mutual Attention}
We've noticed that the performance of the model is not consistently improved by textual queries, largely due to the insufficient modeling of interactions between the video and query inputs.

To address this limitation, this chapter introduces a mutual attention mechanism denoted as $\textup{MutualAtten}(\cdot)$, designed to enhance the model's ability to capture the interactive information between the video and query inputs. The mutual attention $Z_{\textup{ma}}$ leverages convolutional attention, performing a $1 \times 1$ convolution operation.

\begin{equation}
    Z_{\textup{ma}} = \textup{MutualAtten}(Z_{\textup{ta}} \odot Z_{\textup{as}} \odot Z_{\textup{ast}}),
\end{equation}
where $Z_{\textup{ta}}$ represents textual attention and $\odot$ signifies the Hadamard product.

\subsection{Pseudo Segment-level Label Generation}
Consider $S_{f}$ as the set of frame-level score annotations provided by human experts, and $P$ as a pseudo score annotation generator tasked with mapping these frame-level scores to segment-level pseudo scores. 

In \cite{song2015tvsum}, the authors empirically determined that a two-second segment effectively captures the local context of a video while maintaining good visual coherence. Building on this observation, this chapter introduces a pseudo label generator $P$ designed to assign a segment-level score every two seconds. However, since these generated pseudo score annotations lack validation from human experts, they may contain noisy or biased information. Drawing from insights presented in \cite{zhi2004analysis}, the $\textup{Mean}$ function emerges as an effective method for reducing the noise inherent in segment-level pseudo labels.

Consequently, the proposed pseudo label generator $P$ utilizes the $\textup{Mean}$ function to generate the mean score $S_{\textup{mean}}=P(S_{f})=\textup{Mean}(S_{f})$, representing the segment-level pseudo score label for two-second segments. During the training phase, this mean-based pseudo segment label $S_{\textup{mean}}$ is leveraged not only for spatial supervision but also for temporal supervision. This integration of temporal supervision, facilitated by segment-level pseudo annotations, contributes to the improved performance of the query-based video summarization model.

\subsection{Loss Function}
In alignment with \cite{huang2020query}, query-based video summarization is conceptualized as a classification problem. Therefore, in this chapter, the proposed method employs the categorical cross-entropy loss function.

\begin{equation}
    \textup{Loss} = -\frac{1}{N}\sum_{i=1}^{N}\sum_{c=1}^{C}\mathbf{1}_{y_{i}\in C_{c}}\textup{log}(P_{\textup{model}}\left [y_{i}\in C_{c} \right ]),
    \label{eq:loss}
\end{equation}
where $N$ represents the number of observations, $C$ signifies the number of categories, $\mathbf{1}_{y_{i} \in C_{c}}$ serves as an indicator function denoting whether the $i$-th observation belongs to the $c$-th category, and $P_{\textup{model}}[y_i \in C_{c}]$ denotes the probability predicted by the model that the $i$-th observation falls within the $c$-th category.

\section{Experiments and Analysis}
In this section, we provide a comprehensive overview of the datasets utilized, the evaluation metric employed, and the experimental setup. Following this, we proceed to assess, analyze, and compare the efficacy of our proposed video summarization method against existing state-of-the-art methods.

\subsection{Datasets and Evaluation Metrics}
\noindent\textbf{Datasets}: 
TVSum \cite{song2015tvsum} is a widely utilized dataset for traditional video summarization, comprising solely the video input. However, in works such as \cite{yuan2017video,zhou2018video}, TVSum metadata, including video titles, is treated as text-based query input to generate query-dependent video summaries. For our experiments, we randomly partition the TVSum dataset into training, validation, and testing sets with a ratio of $40/5/5$ videos, respectively. The video durations range from $2$ to $10$ minutes, and human expert score labels, annotated with $20$ frame-level responses per video \cite{zhou2018video}, span from $1$ to $5$.

The SumMe dataset \cite{gygli2014creating} is partitioned into training, validation, and testing sets with $19$, $3$, and $3$ videos, respectively. Video durations in SumMe range from $1$ to $6$ minutes, with human expert annotation scores varying from $0$ to $1$. Notably, SumMe is not utilized for query-based video summarization, and hence, no query input is provided when evaluating models on this dataset.

QueryVS \cite{huang2020query} is a dedicated dataset crafted for query-based video summarization. For our experiments, the QueryVS dataset is split into training ($114$ videos), validation ($38$ videos), and testing ($38$ videos) sets. Video durations in QueryVS vary from $2$ to $3$ minutes, with each video retrieved based on a specified text-based query.

To assess the effectiveness of our proposed video summarization approach, we create three segment-level datasets derived from the aforementioned frame-level datasets. Both the segment-level dataset (used for pre-training) and the frame-level dataset (the target dataset) are employed in our experiments.

\noindent\textbf{Evaluation Metric}: 
Based on \cite{song2015tvsum,gygli2014creating,yuan2017video,zhou2018video,hripcsak2005agreement}, the $F_{\beta}$-score with the hyper-parameter $\beta=1$ serves as a widely adopted metric for evaluating the performance of supervised video summarization methods. This metric is rooted in quantifying the agreement between predicted scores and the ground truth scores provided by human experts. The $F_{\beta}$-score is mathematically expressed as: 

\begin{equation}
    F_{\beta}=\frac{1}{N}\sum_{i=1}^{N}\frac{(1+\beta ^{2})\times p_{i}\times r_{i}}{(\beta ^{2}\times p_{i})+r_{i}},
\end{equation}
where $r_{i}$ represents the $i$-th recall, $p_{i}$ represents the $i$-th precision, $N$ denotes the number of $(r_{i}, p_{i})$ pairs, ``$\times$'' indicates the scalar product, and $\beta$ is employed to adjust the relative importance between recall and precision.

\subsection{Experimental Settings}
In the experiments, we employ a 2D ResNet-34 network pre-trained on the ImageNet dataset \cite{deng2009imagenet} to extract frame-level features for each input video. Specifically, we extract $512$ features from the visual layer, one layer below the classification layer. For segment-level features, we utilize a 3D ResNet-34 pre-trained on the Kinetics benchmark \cite{carreira2017quo}, extracting $512$-dimensional features from the visual layer immediately after the global average pooling layer. 

The video lengths in the SumMe, TVSum, and QueryVS datasets vary, with the maximum number of frames in a video being $388$ for SumMe, $199$ for QueryVS, and $647$ for TVSum. To ensure uniformity, we employ a frame-repeating preprocessing technique \cite{huang2020query} to make all videos in each dataset the same length.

For the CNN input, we resize frames to $224 \times 224$ pixels and use RGB channels. Each channel is normalized using a standard deviation of $(0.2737, 0.2631, 0.2601)$ and a mean of $(0.4280, 0.4106, 0.3589)$. We implement and train models using PyTorch for $100$ epochs with a learning rate of $1e-7$. We utilize the Adam optimizer with hyper-parameters set to $\epsilon=1e-8$, $\beta_{1}=0.9$, and $\beta_{2}=0.999$.

\subsection{Ablation Study}
The ablation study of the proposed approach is depicted in Table~\ref{table:table1_chapter_5}. It is evident that the baseline model, lacking the mutual attention mechanism, pseudo segment-level label pre-training, and semantics booster, underperforms significantly compared to models incorporating one or more of these enhancements. Notably, in the absence of the semantics booster, the BoW embedding method is employed.

The mutual attention mechanism enhances the model's ability to capture the interaction between the input query and video. Pseudo segment-level label pre-training contributes to better initialization of the proposed model. The semantics booster plays a pivotal role in capturing the semantic meaning of the text-based query.

\begin{table}[t!]
    \caption{Ablation study on pseudo segment-level label pre-training, semantics booster, and mutual attention mechanism based on $F_{1}$-Score. The symbol ``\checkmark'' indicates that a component is available.}
\centering
\scalebox{0.89}{
\renewcommand{\arraystretch}{1.3}
\begin{tabular}{c|c|c|c|c}
\toprule
\makecell{Pseudo label\\pre-training} & \makecell{Mutual\\attention} & \makecell{Semantics\\booster} &
\textbf{TVSum} & \textbf{QueryVS} \\ 
\midrule
 &  &  & 47.5 & 50.8 \\ 
\midrule
\checkmark &  &  & 61.3 & 52.9 \\ 
\midrule
 & \checkmark &  & 58.9 & 52.0 \\ 
\midrule
 &  & \checkmark & 56.4 & 52.3 \\ 
\midrule
\checkmark & \checkmark & \checkmark & \cellcolor{mygray} \textbf{68.4} & \cellcolor{mygray} \textbf{55.3} \\ 
\bottomrule
\end{tabular}}
\label{table:table1_chapter_5}
\end{table}

\begin{table}[t!]
    \caption{Comparison with state-of-the-art video summarization approaches using $F_{1}$-score, highlighting the best performances in bold. ``-'' indicates unavailability from previous work.}
\centering
\scalebox{0.89}{
\renewcommand{\arraystretch}{1.3}
\begin{tabular}{c|c|c|c|c}
\toprule
\textbf{Model} & \textbf{Method} & \textbf{TVSum} & \textbf{SumMe}  &  \textbf{QueryVS} \\
\midrule
vsLSTM \cite{zhang2016video} & \multirow{5}{*}{\shortstack{Fully\\supervised}} & 54.2 & 37.6 & - \\ 
\cline{1-1}\cline{3-5}
H-RNN \cite{zhao2017hierarchical} & & 57.7 & 41.1 & - \\
\cline{1-1}\cline{3-5}
HSA-RNN \cite{zhao2018hsa} & & 59.8 & 44.1 & - \\
\cline{1-1}\cline{3-5}
iPTNet \cite{jiang2022joint} & & 63.4 & 54.5 & - \\
\cline{1-1}\cline{3-5}
SMLD \cite{chu2019spatiotemporal} & & 61.0 & 47.6 & - \\
\cline{1-1}\cline{3-5}
SMN \cite{wang2019stacked} & & 64.5 & \textbf{58.3} & - \\
\midrule
FPVSF \cite{ho2018summarizing} & \multirow{2}{*}{\shortstack{Weakly\\supervised}} & - & 41.9 & - \\ 
\cline{1-1}\cline{3-5}
WS-HRL \cite{chen2019weakly} & & 58.4 & 43.6 & - \\
\midrule
DSSE \cite{yuan2017video} & \multirow{5}{*}{\shortstack{Query\\based}} & 57.0 & - & - \\
\cline{1-1}\cline{3-5}
DQSN \cite{zhou2018video} & & 58.6 & - & - \\
\cline{1-1}\cline{3-5}
QueryVS \cite{huang2020query} & & - & - & 41.4 \\
\cline{1-1}\cline{3-5}
GPT2MVS \cite{huang2021gpt2mvs} & & - & - & 54.8 \\
\cline{1-1}\cline{3-5}
\cellcolor{mygray} Ours & & \cellcolor{mygray} \textbf{68.4} & \cellcolor{mygray} 52.4 & \cellcolor{mygray} \textbf{55.3} \\
\bottomrule
\end{tabular}}
\label{table:table2_chapter_5}
\end{table}

\begin{figure}[t!]
\begin{center}
\includegraphics[width=1.0\textwidth]{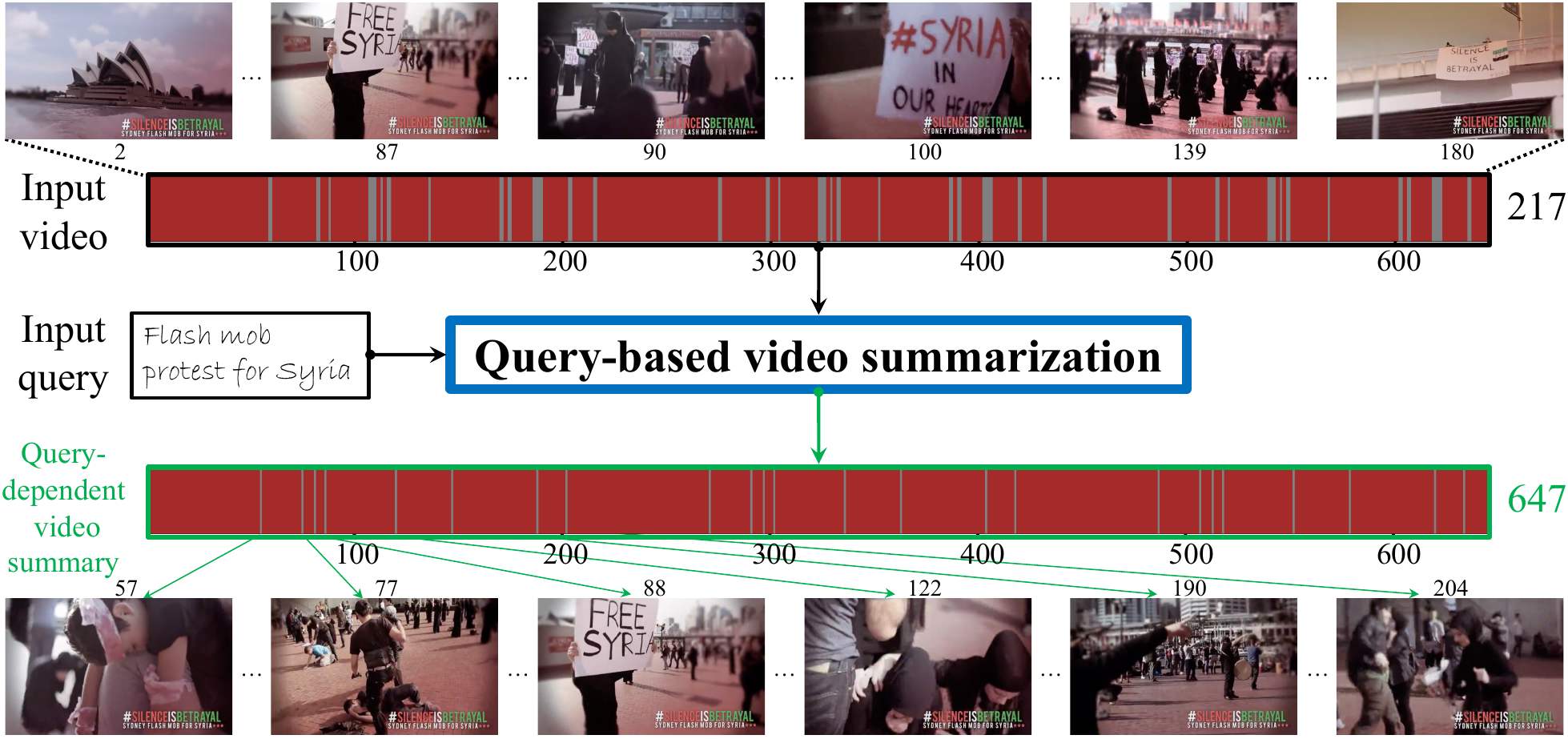}
\end{center}
   \caption{Qualitative analysis. 
   The first two rows illustrate the input video visualization along with the corresponding ground truth frame-based score annotations. The last two rows showcase the visualization of predicted frame-based scores and a partial visualization of the query-dependent video summary. In each frame-based score pattern, gray indicates ``not selected frames'', and red indicates ``selected frames''. The numbers ``$217$'' and ``$647$'' denote the video length before and after video preprocessing, respectively. Additionally, the indices of the visualized selected frames are also displayed.
   }
\label{fig:qualitative_mm}
\end{figure}

\subsection{Comparison with State-of-the-art Models}
The comparison with existing query-based, weakly supervised, and fully supervised approaches is detailed in Table~\ref{table:table2_chapter_5}. The results underscore the superior performance of our proposed method on the TVSum and QueryVS datasets, coupled with competitive results on the SumMe dataset.

Although the correctness of the generated segment-level pseudo labels is not guaranteed by human experts, they still contain valuable information, such as improved temporal information, to supervise the proposed model during pre-training. In weakly supervised methods, although the correctness of coarse labels, such as video-level labels, is assured by human experts, it is insufficient to outperform our proposed method. In query-based summarization methods, although the utilization of another modality contributes to enhancing model performance, the inefficiency of multi-modal feature fusion could constrain the potential performance improvement.

\subsection{Qualitative Analysis}
Qualitative results, demonstrating the effectiveness of the proposed method, are illustrated in Figure~\ref{fig:qualitative_mm}.

\section{Conclusion}
In summary, this chapter tackles the challenge of query-dependent video summarization by introducing a segment-based video summarization pretext task with meticulously crafted pseudo labels. These labels are specifically designed to capture the underlying relations between the pretext and target tasks, aiming to overcome the constraints posed by small human-annotated datasets in fully supervised deep learning tasks. Moreover, a novel mutual attention mechanism coupled with a semantics booster is proposed to efficiently capture interactive information between textual and visual modalities, thereby bolstering model performance.
Extensive experiments are conducted on widely recognized video summarization benchmarks to validate the effectiveness of the proposed approach, showcasing its state-of-the-art performance in query-based video summarization. Additionally, an ablation study is conducted to illustrate how the proposed approach effectively addresses the challenges associated with fully supervised deep learning tasks within the context of query-based video summarization.

\chapter{Summary and Conclusions}

\section{Summary}

This thesis is focused on improving automatic video summarization, and aims to answer the research question: \textit{How can we enhance the effectiveness of automatic video summarization?} In \textbf{Chapter~\ref{ch:spectral}}, we investigate the use of a text-based query to generate a query-dependent video summary in a multi-modal input scenario. In \textbf{Chapter~\ref{ch:filtering}}, we propose a method to effectively embed the query, which is considered the main driving force behind query-dependent video summary generation. In \textbf{Chapter~\ref{ch:focus}}, we introduce a new perspective on conditional modeling inspired by how humans create video summaries. Finally, in \textbf{Chapter~\ref{ch:things}}, we propose a method to address the issue of data scarcity in supervised video summarization. A brief overview of each chapter is provided below:


\textbf{Chapter~\ref{ch:spectral}: Query-dependent Video Summarization.} The aim of this chapter is to propose a video summarization model that can generate a query-dependent video summary. 
The chapter discusses the challenge of exploring video collections efficiently, particularly when dealing with large collections, and investigates how video summarization techniques can be used to address this challenge. Conventional video summarization methods generate fixed summaries, which may not always meet the user's information needs. To overcome this limitation, a multi-modal deep learning method is proposed that uses both text-based queries and visual information to generate summaries tailored to a specific user's information needs. The approach involves fusing visual and textual information at different levels of the model architecture to generate accurate and relevant summaries. The model's performance is evaluated using metrics such as accuracy and F1-score, and the results indicate that incorporating text-based queries improves the model's performance.

\textbf{Chapter~\ref{ch:filtering}: GPT-2 for Multi-modal Video Summarization.} The purpose of this chapter is to present an effective method for embedding queries and videos. 
It focuses on multi-modal video summarization which aims to generate comprehensive and informative video summaries by incorporating multiple sources of information. Traditional video summarization methods generate fixed summaries that may not provide a complete understanding of the content, while multi-modal video summarization can utilize various sources, such as text annotations or emotional cues, to create more comprehensive summaries. Embedding a text-based query is critical to generate query-dependent video summaries. This chapter proposes a new method that employs a specialized attention network and contextualized word representations to effectively encode both the text-based query and the video. The experimental results demonstrate that the proposed model outperforms the state-of-the-art methods based on the evaluation on an existing multi-modal video summarization benchmark.

\textbf{Chapter~\ref{ch:focus}: Conditional Modeling Based Video Summarization.} In this chapter, a novel conditional modeling perspective for video summarization is proposed. 
The chapter highlights the limitations of current video summarization techniques that mainly rely on visual factors and tend to not, or in a very limited way, consider non-visual factors such as interestingness, representativeness, and storyline consistency. To address these limitations, the proposed approach integrates insights gained from how humans create high-quality video summaries. It employs a conditional modeling perspective by introducing multiple meaningful random variables and joint distributions to capture the key components of video summarization. The proposed method shows superior performance compared to existing methods and achieves state-of-the-art performance on commonly used video summarization datasets. Additionally, the method is associated with a conditional/causal graph for video summarization, which improves model explainability. The experimental results demonstrate that the proposed approach achieves state-of-the-art performance in terms of the F1-score.

\textbf{Chapter~\ref{ch:things}: Pseudo-Label Supervision Enhances Video Summarization.} The goal of this chapter is to propose a solution to the challenge of data scarcity in supervised video summarization. 
Specifically, we propose a segment-level pseudo-labeling approach that uses a pretext task to generate additional data with pseudo labels, which can be used to pre-train a supervised deep model. The approach generates pseudo-labels from existing frame-level labels provided by humans. To further enhance the accuracy of query-dependent video summaries, we introduce a semantics booster that produces context-aware query representations, and a mutual attention mechanism that captures the interaction between visual and textual modalities. We validate the proposed approach on three commonly used video summarization benchmarks, and our experimental results demonstrate that our algorithm achieves state-of-the-art performance.

\section{Conclusions}
This thesis delivers noteworthy contributions toward advancing the effectiveness of automatic video summarization. By integrating text-based queries with efficient embeddings, it has improved the performance of video summarization algorithms. Furthermore, the introduction of a novel conditional modeling perspective imparts a more human-like quality to the video summarization process. Addressing the challenge of data scarcity inherent in modeling video summarization as a supervised learning task, this thesis proposes a pseudo-label supervision method to bolster the training process. These innovative approaches not only yield promising results but also establish state-of-the-art performance on various widely used video summarization benchmarks. 

Beyond the realm of technical innovation, the impact of these advancements extends far beyond academic boundaries, revolutionizing video summarization practices across diverse applications and industries.
For instance, in the context of a pandemic like COVID-19, where the urgent need to identify potentially infected individuals for swift quarantine is paramount, the conventional method of manual surveillance video analysis proves inefficient. The novel video summarization technique proposed in this thesis offers a promising solution by enabling personalized or customized text-based queries tailored to known individual characteristics. This approach holds the potential to significantly enhance the efficiency and effectiveness of current workflows.
Similarly, in scenarios involving the detection of individuals with malicious intent, such as potential attackers in public spaces, the proposed video summarization technology offers a valuable tool. By utilizing personalized text-based queries that focus on specific behavioral or characteristic indicators associated with the individual under scrutiny, this method provides a more efficient and effective means of analysis.

Building upon the integration of text-based queries as an additional input modality explored in this thesis, a logical and promising direction for future research involves exploring the integration of audio-based queries as well. 
Harnessing the valuable mutual information from these distinct input modalities has the potential to further enhance model performance.
Since audio-based queries are not usually the initial focus, this extension presents a significant opportunity to advance video summarization practices and explore novel applications in real-world scenarios.
Furthermore, recent advancements in image/video generation and ChatGPT-like models offer an opportunity to integrate these models with current video summarization approaches. 
Leveraging generative modeling to enhance current video summarization technology has the potential to broaden and innovate the content of machine-generated video summaries.
This integration could expand the scope of real-world applications and drive innovation in video summarization techniques.

Improving the performance of video summarization not only streamlines time and resource utilization but also enhances the accessibility of video content for various purposes. Thus, in essence, the methodology presented in this thesis not only represents a technical breakthrough but also holds immense promise for improving practical applications in various real-world contexts.


\newpage

\bibliographystyle{abbrv}
\bibliography{thesis}


\appendix
\chapter{Complete List of Publications}
\label{ch:pub}

\begin{itemize}
  \item \textbf{Jia-Hong Huang}, Yixian Shen, Evangelos Kanoulas. ``GradNormLoRP: Gradient Weight-normalized Low-rank Projection 
        for Efficient LLM Training'', in submission to \textbf{Neural Information Processing Systems (NeurIPS)}, 2024.
  \item \textbf{Jia-Hong Huang}, Weijuan Xi, Eunyoung Kim, Wonsik Kim, Amelio Vázquez-Reina. ``CSR-ASD: Cross-modal Signal 
        Reprogramming Enhancing Audio-Visual Active Speaker Detection'', in submission to \textbf{European Conference on Computer Vision (ECCV)}, 2024.
  \item Hongyi Zhu*, \textbf{Jia-Hong Huang*}, Yixian Shen, Shuai Wang, Stevan Rudinac, Evangelos Kanoulas. ``Enhancing Text-to-Video Retrieval Through Multi-Modal Multi-Turn Conversation'', in submission to \textbf{ACM Multimedia (ACMMM)}, 2024.
  \item Shuai Wang, David W Zhang, \textbf{Jia-Hong Huang}, Stevan Rudinac, Monika Kackovic, Nachoem Wijnberg, Marcel Worring. 
        ``Ada-HGNN: Adaptive Sampling for Scalable Hypergraph Neural Networks'', in submission to \textbf{ACM International Conference on Information and Knowledge Management (CIKM)}, 2024. \cite{wang2024ada}
  \item \textbf{Jia-Hong Huang}, Chao-Han Huck Yang, Pin-Yu Chen, Min-Hung Chen, Marcel Worring. ``Conditional Modeling 
        Based Automatic Video Summarization'', in submission to \textbf{ACM Transactions on Multimedia Computing, Communications, and Applications (TOMM)}, 2024. \cite{huang2023conditional}
  \item \textbf{Jia-Hong Huang}, Modar Alfadly, Bernard Ghanem, Marcel Worring. ``Improving Visual Question Answering Models 
        through Robustness Analysis and In-Context Learning with a Chain of Basic Questions'', in submission to \textbf{IEEE 
        Transactions on Multimedia (TMM)}, 2024. \cite{huang2019assessing,huang2023improving}
  \item \textbf{Jia-Hong Huang}. ``Automated Retinal Image Analysis and Medical Report Generation through Deep Learning'', 
        \textbf{University of Amsterdam, Doctoral Thesis}, 2024. \cite{huang2024automated}
  \item \textbf{Jia-Hong Huang*}, Hongyi Zhu*, Yixian Shen, Stevan Rudinac, Alessio M. Pacces, Evangelos Kanoulas. ``A Novel 
        Evaluation Framework for Image2Text Generation'', \textbf{International ACM SIGIR Conference on Research and Development in Information Retrieval (SIGIR)} LLM4Eval Workshop, 2024. \cite{huang2024novel}
  \item \textbf{Jia-Hong Huang*}, Chao-Chun Yang*, Yixian Shen, Alessio M. Pacces, Evangelos Kanoulas. ``Optimizing Numerical 
        Estimation and Operational Efficiency in the Legal Domain through Large Language Models'', \textbf{ACM International Conference on Information and Knowledge Management (CIKM)}, 2024. \cite{huang2024optimizing}
  \item \textbf{Jia-Hong Huang}. ``Multi-modal Video Summarization'', \textbf{ACM International Conference on Multimedia 
        Retrieval (ICMR)}, 2024. \cite{huang2024multi}
  \item Hongyi Zhu*, \textbf{Jia-Hong Huang*}, Stevan Rudinac, Evangelos Kanoulas. ``Enhancing Interactive Image Retrieval With 
        Query Rewriting Using Large Language Models and Vision Language Models'', \textbf{ACM International Conference on Multimedia Retrieval (ICMR)}, 2024. \cite{zhu2024enhancing}
  \item Weijia Zhang*, \textbf{Jia-Hong Huang*}, Svitlana Vakulenko, Yumo Xu, Thilina Rajapakse, Evangelos Kanoulas. ``Beyond 
        Relevant Documents: A Knowledge-Intensive Approach for Query-Focused Summarization using Large Language Models'', \textbf{International Conference on Pattern Recognition (ICPR)}, 2024. \cite{zhang2024beyond}
  \item Weijia Zhang, Mohammad Aliannejadi, Jiahuan Pei, Yifei Yuan, \textbf{Jia-Hong Huang}, Evangelos Kanoulas. ``A Comparative 
        Analysis of Faithfulness Metrics and Humans in Citation Evaluation'', \textbf{International ACM SIGIR Conference on Research and Development in Information Retrieval (SIGIR)} LLM4Eval Workshop, 2024. \cite{zhang2024comparative}
  \item Weijia Zhang, Mohammad Aliannejadi, Yifei Yuan, Jiahuan Pei, \textbf{Jia-Hong Huang}, Evangelos Kanoulas. ``Towards Fine- 
        Grained Citation Evaluation in Generated Text: A Comparative Analysis of Faithfulness Metrics'', \textbf{International Natural Language Generation Conference (INLG)}, 2024. [\textbf{Oral}] \cite{zhang2024towards}
  \item Weijia Zhang, Vaishali Pal, \textbf{Jia-Hong Huang}, Evangelos Kanoulas, Maarten de Rijke. ``Beyond Relevant Documents: A 
        Knowledge-Intensive Approach for Query-Focused Summarization using Large Language Models'', \textbf{European Conference on Artificial Intelligence (ECAI)}, 2024. \cite{zhang2024qfmts}
  \item \textbf{Jia-Hong Huang}, Luka Murn, Marta Mrak, Marcel Worring. ``Query-based Video Summarization with Pseudo Label 
        Supervision'', \textbf{IEEE International Conference on Image Processing (ICIP)}, 2023. \cite{huang2023query}
  \item \textbf{Jia-Hong Huang}, Chao-Han Huck Yang, Pin-Yu Chen, Min-Hung Chen, Marcel Worring. ``Causalainer: Causal    
        Explainer for Automatic Video Summarization'', \textbf{IEEE/CVF Conference on Computer Vision and Pattern Recognition (CVPR)} New Frontiers in Visual Language Reasoning Workshop, 2023. \cite{huang2023causalainer}
  \item Ting-Wei Wu*, \textbf{Jia-Hong Huang*}, Joseph Lin, Marcel Worring. ``Expert-defined Keywords Improve 
        Interpretability of Retinal Image Captioning'', \textbf{IEEE/CVF Winter Conference on Applications of Computer Vision (WACV)}, 2023. [\textbf{Oral}]. \cite{wu2023expert}
  \item \textbf{Jia-Hong Huang}, Chao-Han Huck Yang, Pin-Yu Chen, Andrew Brown, Marcel Worring. ``Causal Video Summarizer 
        for Video Exploration'', \textbf{IEEE International Conference on Multimedia and Expo (ICME)}, 2022. \cite{huang2022causal}
  \item Riccardo Di Sipio, \textbf{Jia-Hong Huang}, Samuel Yen-Chi Chen, Stefano Mangini, Marcel Worring. ``The Dawn of 
        Quantum Natural Language Processing'', \textbf{IEEE International Conference on Acoustics, Speech and Signal Processing (ICASSP)}, 2022. \cite{di2022dawn}
  \item \textbf{Jia-Hong Huang}, Ting-Wei Wu, Chao-Han Huck Yang, Zenglin Shi, I-Hung Lin, Jesper Tegner, Marcel Worring. 
        ``Non-local Attention Improves Description Generation for Retinal Images'', \textbf{IEEE/CVF Winter Conference on Applications of Computer Vision (WACV)}, 2022. [\textbf{Oral}]. \cite{huang2022non}
  \item \textbf{Jia-Hong Huang}, Marta Mrak, Luka Murn, Marcel Worring. ``GPT2MVS: Generative Pre-trained Transformer-2 for 
        Multi-modal Video Summarization'', \textbf{ACM International Conference on Multimedia Retrieval (ICMR)}, 2021. [\textbf{Oral}]. \cite{huang2021gpt2mvs}
  \item \textbf{Jia-Hong Huang}, Ting-Wei Wu, Marcel Worring. ``Contextualized Keyword Representations for Multi-modal 
        Retinal Image Captioning'', \textbf{ACM International Conference on Multimedia Retrieval (ICMR)}, 2021. \cite{huang2021contextualized}
  \item \textbf{Jia-Hong Huang}, Ting-Wei Wu, Chao-Han Huck Yang, Marcel Worring. ``Deep Context-Encoding Network for 
        Retinal Image Captioning'', \textbf{IEEE International Conference on Image Processing (ICIP)}, 2021. \cite{huang2021deep,huang2021longer}
  \item \textbf{Jia-Hong Huang}, Chao-Han Huck Yang, Fangyu Liu, Meng Tian, Yi-Chieh Liu, Ting-Wei Wu, I-Hung Lin, Kang 
        Wang, Hiromasa Morikawa, Herng-Hua Chang, Jesper Tegner, Marcel Worring. ``DeepOpht: Medical Report Generation for Retinal Images via Deep Models and Visual Explanation'', \textbf{IEEE/CVF Winter Conference on Applications of Computer Vision (WACV)}, 2021. \cite{huang2021deepopht}
  \item \textbf{Jia-Hong Huang}, Marcel Worring. ``Query-controllable Video Summarization'', \textbf{ACM International 
        Conference on Multimedia Retrieval (ICMR)}, 2020. \cite{huang2020query}
  \item \textbf{Jia-Hong Huang}, Cuong Duc Dao, Modar Alfadly, Bernard Ghanem. ``A Novel Framework for Robustness Analysis 
        of Visual QA Models'', \textbf{AAAI Conference on Artificial Intelligence (AAAI)}, 2019. [\textbf{Oral}] \cite{huang2019novel}
  \item Tao Hu, Pascal Mettes, \textbf{Jia-Hong Huang}, Cees G. M. Snoek. ``SILCO: Show a Few Images, Localize the Common 
        Object'', \textbf{IEEE/CVF International Conference on Computer Vision (ICCV)}, 2019. \cite{hu2019silco}
  \item \textbf{Jia-Hong Huang}, Cuong Duc Dao, Modar Alfadly, Chao-Han Huck Yang, Bernard Ghanem. ``Robustness Analysis of  
        isual QA Models by Basic Questions'', \textbf{IEEE/CVF Conference on Computer Vision and Pattern Recognition (CVPR)} Visual Dialog Workshop, 2018. \cite{huang2018robustness}
  \item Chao-Han Huck Yang*, \textbf{Jia-Hong Huang*}, Fangyu Liu, Fang-Yi Chiu, Mengya Gao, Weifeng Lyu, I-Hung Lin, Jesper 
        Tegner. ``A Novel Hybrid Machine Learning Model for Auto-Classification of Retinal Diseases'', \textbf{International Conference on Machine Learning (ICML)} Computational Biology Workshop, 2018. \cite{yang2018novel}
  \item Yi-Chieh Liu, Hao-Hsiang Yang, Chao-Han Huck Yang, \textbf{Jia-Hong Huang}, Meng Tian, Hiromasa Morikawa, Yi-Chang 
        James Tsai, Jesper Tegner. ``Synthesizing New Retinal Symptom Images by Multiple Generative Models'', \textbf{Asian Conference on Computer Vision (ACCV)} Artificial Intelligence Applied to Retinal Image Analysis Workshop, 2018. [\textbf{Oral}] \cite{liu2019synthesizing}
  \item Chao-Han Huck Yang*, Fangyu Liu*, \textbf{Jia-Hong Huang*}, Meng Tian, I-Hung Lin, Yi-Chieh Liu, Hiromasa Morikawa, 
        Hao-Hsiang Yang, Jesper Tegner. ``Auto-Classification of Retinal Diseases in the Limit of Sparse Data Using a Two-Streams Machine Learning Model'', \textbf{Asian Conference on Computer Vision (ACCV)} Artificial Intelligence Applied to Retinal Image Analysis Workshop, 2018. \cite{huck2019auto}
  \item \textbf{Jia-Hong Huang}, Modar Alfadly, Bernard Ghanem. ``VQABQ: Visual Question Answering by Basic Questions'', 
        \textbf{IEEE/CVF Conference on Computer Vision and Pattern Recognition (CVPR)} Visual Question Answering Workshop, 2017. \cite{huang2017vqabq,huang2017robustness}
\end{itemize}

\chapter{Samenvatting}
Dit proefschrift richt zich op het verbeteren van automatische videosamenvatting en heeft als doel de onderzoeksvraag te beantwoorden: \textit{Hoe kunnen we de effectiviteit van automatische videosamenvatting verbeteren?} In \textbf{Hoofdstuk~\ref{ch:spectral}} onderzoeken we het gebruik van een op tekst gebaseerde zoekopdracht om een zoekafhankelijke videosamenvatting te genereren voor multimodale invoer. In \textbf{Hoofdstuk~\ref{ch:filtering}} stellen we een methode voor om de zoekopdracht effectief in te bedden, wat wordt beschouwd als de belangrijkste drijvende kracht achter de generatie van een zoekvraagafhankelijke videosamenvatting. In \textbf{Hoofdstuk~\ref{ch:focus}} introduceren we een nieuw perspectief op conditionele modellering, geïnspireerd op hoe mensen videosamenvattingen maken. Ten slotte stellen we in \textbf{Hoofdstuk~\ref{ch:things}} een methode voor om het probleem van data-schaarste bij handmatig geannoteerde videosamenvatting aan te pakken. Een beknopt overzicht van elk hoofdstuk wordt hieronder gegeven:

\textbf{Hoofdstuk~\ref{ch:spectral}: Zoekvraagafhankelijke Videosamenvatting.} Het doel van dit hoofdstuk is om een videosamenvattingsmodel voor te stellen dat een zoekvraagafhankelijke videosamenvatting kan genereren.
Het hoofdstuk bespreekt de uitdaging van het efficiënt verkennen van met name grote videocollecties, grote collecties, en onderzoekt hoe videosamenvattingstechnieken kunnen worden gebruikt om deze uitdaging aan te pakken. Conventionele videosamenvattingsmethoden genereren vaste beschrijvingen, die niet altijd voldoen aan de informatie behoeften van de gebruiker. Om deze beperking te overwinnen, wordt een multimodale diepe leermethode voorgesteld die zowel op tekst gebaseerde zoekopdrachten als visuele informatie gebruikt om beschrijvingen op maat te maken voor de specifieke informatie behoeften van een gebruiker. De aanpak omvat het samenvoegen van visuele en tekstuele informatie op verschillende niveaus van de modelarchitectuur om nauwkeurige en relevante beschrijvingen te genereren. De prestaties van het model worden geëvalueerd aan de hand van metrieken zoals nauwkeurigheid en F1-score, en de resultaten geven aan dat het opnemen van op tekst gebaseerde zoekopdrachten de prestaties van het model verbetert.

\textbf{Hoofdstuk~\ref{ch:filtering}: GPT-2 voor Multimodale Videosamenvatting.} Het doel van dit hoofdstuk is een effectieve methode voor het inbedden van zoekopdrachten en video's.
Het richt zich op multimodale videosamenvatting die tot doel heeft uitgebreide en informatieve videosamenvattingen te genereren door meerdere informatiebronnen op te nemen. Traditionele videosamenvattingsmethoden genereren vaste beschrijvingen die mogelijk geen volledig begrip van de inhoud bieden, terwijl multimodale videosamenvatting verschillende bronnen kan gebruiken, zoals tekstannotaties of emotionele aanwijzingen, om meer omvattende beschrijvingen te creëren. Het inbedden van een op tekst gebaseerde zoekopdracht is cruciaal om zoekvraagafhankelijke videosamenvattingen te genereren. Dit hoofdstuk stelt een nieuwe methode voor die een gespecialiseerd aandachtsnetwerk en gecontextualiseerde woordrepresentaties gebruikt om zowel de op tekst gebaseerde zoekopdracht als de video effectief te encoderen. De experimentele resultaten tonen aan dat het voorgestelde model beter presteert dan de state-of-the-art methoden op basis van de evaluatie van een bestaande benchmark voor multimodale videosamenvatting.

\textbf{Hoofdstuk~\ref{ch:focus}: Videosamenvatting op basis van Voorwaardelijke Modellering.} In dit hoofdstuk wordt een nieuwe perspectief op conditionele modellering voor videosamenvatting voorgesteld.
Het hoofdstuk belicht de beperkingen van de huidige videosamenvattingstechnieken die voornamelijk steunen op visuele factoren en neigen om niet, of in zeer beperkte mate, niet-visuele factoren zoals interessantheid, representativiteit en verhaalconsistentie te overwegen. Om deze beperkingen aan te pakken, integreert de voorgestelde aanpak inzichten uit hoe mensen hoogwaardige videosamenvattingen creëren. Het maakt gebruik van een perspectief op voorwaardelijke modellering door meerdere betekenisvolle stochastischevariabelen en gezamenlijke verdelingen te introduceren om de belangrijkste componenten van videosamenvatting vast te leggen. De voorgestelde methode vertoont een superieure prestatie in vergelijking met bestaande methoden en behaalt state-of-the-art prestaties op veelgebruikte datasets voor videosamenvatting. Bovendien wordt de methode geassocieerd met een conditionele/causale grafiek voor videosamenvatting, wat de modelverklaarbaarheid verbetert. De experimentele resultaten tonen aan dat de voorgestelde aanpak state-of-the-art prestaties behaalt in termen van de F1-score.

\textbf{Hoofdstuk~\ref{ch:things}: Pseudo-Label Supervisie Verbetert Videosamenvatting.} Het doel van dit hoofdstuk is om een oplossing voor het probleem van datagebrek in supervisie bij videosamenvatting voor te stellen.
Specifiek stellen we een segmentniveau pseudo-labelingbenadering voor die een pseudo-labeltaken gebruikt om extra data met pseudo-labels te genereren, die kunnen worden gebruikt om een supervisie diep model te trainen. De benadering genereert pseudo-labels op basis van bestaande frame-niveau labels die door mensen zijn verstrekt. Om de nauwkeurigheid van op zoek-afhankelijke videosamenvattingen verder te verbeteren, introduceren we een semantische booster die contextbewuste zoekvragen produceert, en een wederzijdse aandachtsmechanisme dat de interactie tussen visuele en tekstuele modaliteiten vastlegt. We valideren de voorgestelde aanpak op drie veelgebruikte benchmarks voor videosamenvatting, en onze experimentele resultaten tonen aan dat ons algoritme state-of-the-art prestaties behaalt.

\end{document}